\documentclass[12pt]{article}
\usepackage[natbibapa]{apacite}
\usepackage{rotating}
\usepackage{bbm}
\usepackage{latexsym}

\usepackage[hyperfootnotes=false]{hyperref} \usepackage{url}            \usepackage{booktabs}       \usepackage{amsfonts}       \usepackage{nicefrac}       \usepackage{microtype}      \usepackage[cmyk]{xcolor}
\usepackage{tikz,pgfplots,pgfplotstable}
\pgfplotsset{compat=newest}
\usetikzlibrary{positioning}
\usepgfplotslibrary{fillbetween}
\usetikzlibrary{shapes}
\usepgfplotslibrary{colormaps}

\definecolor{mp1color}{cmyk}{1.,1.,0,0}
\definecolor{sor20color}{cmyk}{1,0,0,0} \definecolor{myteal}{cmyk}{1,0,0,0.5} \definecolor{slateblue}{cmyk}{1,0.5,0,0}

\definecolor{nassarNatcolor}{cmyk}{0.5,1,0,0.02}

\definecolor{cmykblue}{cmyk}{1,0.65,0,0}

\definecolor{lightgreensor1}{cmyk}{1,0,1,0}
\definecolor{darkgreenpf20}{cmyk}{1,0,1,0.5}

\tikzset{
font = \small,
actualpcstyle/.style = {anchor = east, rotate = 90, isosceles triangle, fill = red, draw = none, inner sep = 1}, mustyle/.style = {anchor = east, rotate = 90, isosceles triangle, fill = blue, draw = none, inner sep = 1, minimum width=0.3cm}, smile/.style = {orange, very thick},
smileextended/.style = {red, very thick},
    pf1/.style = {lime, very thick},
pf20/.style = {darkgreenpf20, very thick}, leaky/.style = {gray, very thick},
    bayesfilter/.style = {black, very thick},
    gsor1/.style = {slateblue, very thick},
gsor20/.style = {blue, very thick},
    tsor1/.style = {slateblue, very thick},
tsor20/.style = {blue, very thick}, gnassarJN/.style = {magenta, very thick},
gnassarNatN/.style = {nassarNatcolor, very thick},
    gnassarJNOriginal/.style = {magenta, very thick, dashed},
gnassarNatNOriginal/.style = {nassarNatcolor, very thick, dashed},
gsorOriginal1/.style = {lightgreensor1, very thick},
    gsorOriginal20/.style = {sor20color!70!white, very thick},
tsorOriginal1/.style = {lightgreensor1, very thick},
    tsorOriginal20/.style = {sor20color!70!white, very thick},
smileOriginal/.style = {orange, very thick},
GNassarNatN_Sgmpositive/.style =  {orange, thick},
    GNassarNatN_Sgmnegative/.style =  {blue, thick},
    GNassarNatN_Sshpositive/.style =  {orange, thick},
    GNassarNatN_Sshnegative/.style =  {blue, thick},
    GParticleFilter_Sgmpositive/.style =  {orange, thick},
    GParticleFilter_Sgmnegative/.style =  {blue, thick},
    GParticleFilter_Sshpositive/.style =  {orange, thick},
    GParticleFilter_Sshnegative/.style =  {blue, thick},
GNassarNatN_Sgm/.style =  {green, thick},
    GNassarNatN_Ssh/.style =  {mp1color, thick},
GParticleFilter_Sgm/.style =  {green, thick},
    GParticleFilter_Ssh/.style =  {mp1color, thick},
}
\pgfplotsset{
    legend style = {draw = none},
    colorbar/width = 6pt,
}

\definecolor{RoyalBlue}{cmyk}{0.71,0.53,0,0.12}

\pgfplotscreatecolormap{blueColormap}{[2pt]color(0000pt)=(RoyalBlue!50!black);
       color(1000pt)=(RoyalBlue!50!black);
       color(1002pt)=(RoyalBlue!70!black);
       color(2000pt)=(RoyalBlue!70!black);
       color(2002pt)=(RoyalBlue!90!black);
       color(3000pt)=(RoyalBlue!90!black);
       color(3002pt)=(RoyalBlue!90!white);
       color(4000pt)=(RoyalBlue!90!white);
       color(4002pt)=(RoyalBlue!70!white);
       color(5000pt)=(RoyalBlue!70!white);
       color(5002pt)=(RoyalBlue!50!white);
       color(6000pt)=(RoyalBlue!50!white);
   }

\pgfplotscolorbarCMYKworkaroundfalse

\pgfplotsset{compat=newest,
  colormap={orangeyellow}{
  color=(green) color=(yellow)
  },
}

\pgfplotsset{compat=newest,
  colormap={myviridis}{
  color(0cm)=(blue); color(1cm)=(myteal); color(2cm)=(green); color(3cm)=(yellow)
  },
}

\usepackage{algorithm}
\usepackage[noend]{algpseudocode}
\usepackage{setspace}

\usepackage[reqno]{amsmath} \makeatletter
\newcommand{\leqnomode}{\tagsleft@true\let\veqno\@@leqno}
\makeatother

\usepackage{standalone}
\usepackage{xcolor}
\hypersetup{
    colorlinks,
    linkcolor={red!50!black},
    citecolor={blue!50!black},
urlcolor={blue!80!black}
}
\usepackage{amsthm}
\newtheoremstyle{mydefstyle}{}{}{\itshape}{}{\bf}{.}{.5em}{}

\theoremstyle{mydefstyle}
\newtheorem{mydef}{Definition}
\newtheorem*{mylemma}{Proposition}

\usepackage{caption}
\DeclareCaptionLabelFormat{adja-page}{\hrulefill\\#1 #2 \emph{(previous page)}}

\newcommand{\approxs}[1]{\hat{#1}}

\newcommand{\MyStatex}[1] {
  \Statex \textcolor{gray}{\# #1}}

\newcounter{Seq}

\newenvironment{Sequation}
{\stepcounter{Seq}\addtocounter{equation}{-1}\equation}
  {\endequation}

\newcounter{SAlgorithm}

\newenvironment{Salgorithm}
{\stepcounter{SAlgorithm}\addtocounter{algorithm}{-1}\algorithm}
  {\endalgorithm}

\newcounter{SFigure}

  \newenvironment{Sfigure}
  {\stepcounter{SFigure}\addtocounter{figure}{-1}\figure}
    {\endfigure}

\newcommand{\captionfonts}{\normalsize}

\makeatletter
\long\def\@makecaption#1#2{\vskip\abovecaptionskip
  \sbox\@tempboxa{{\captionfonts #1: #2}}\ifdim \wd\@tempboxa >\hsize
    {\captionfonts #1: #2\par}
  \else
    \hbox to\hsize{\hfil\box\@tempboxa\hfil}\fi
  \vskip\belowcaptionskip}
\makeatother

\begin{document}
\hspace{13.9cm}

\ \vspace{20mm}\\
\begin{center}
{\LARGE Learning in Volatile Environments with the \\Bayes Factor Surprise}\footnote{\normalsize To appear in \textit{Neural Computation}.}
\end{center}

\ \\
{\bf \large Vasiliki Liakoni$^{\displaystyle \dagger, \displaystyle \ddagger, \displaystyle *}$},
{\bf \large Alireza Modirshanechi$^{\displaystyle \dagger, \displaystyle \ddagger, \displaystyle *}$},
{\bf \large Wulfram Gerstner$^{\displaystyle \dagger}$}, and
{\bf \large Johanni Brea$^{\displaystyle \dagger}$}\\
{$^{\displaystyle \dagger}$\'{E}cole Polytechnique F\'{e}d\'{e}rale de Lausanne (EPFL), School of Computer and Communication Sciences and School of Life Sciences, Lausanne, Switzerland}\\
{$^{\displaystyle \ddagger}$Equal Contribution}\\
{$^{\displaystyle *}$Corresponding authors:\\ vasiliki.liakoni@epfl.ch (VL) and alireza.modirshanechi@epfl.ch (AM)}

\thispagestyle{empty}
\markboth{}{NC instructions}
\ \vspace{-0mm}\\

\begin{center} {\bf Abstract} \end{center}
  Surprise-based learning allows agents to rapidly adapt to non-stationary stochastic environments characterized
by sudden changes.
  We show that exact Bayesian inference in a hierarchical model gives rise to a surprise-modulated trade-off between forgetting old observations and integrating them with the new ones. The modulation depends on a probability ratio, which we call ``Bayes Factor Surprise'', that tests the prior belief against the current belief. We demonstrate that in several existing approximate algorithms the Bayes Factor Surprise modulates the rate of adaptation to new observations. We derive three novel surprised-based algorithms, one in the family of particle filters, one in the family of variational learning, and the other in the family of message passing, that have constant scaling in observation sequence length and particularly simple update dynamics for any distribution in the exponential family.
Empirical results show that these surprise-based algorithms estimate parameters better than alternative approximate approaches and reach levels of performance comparable to computationally more expensive algorithms. The Bayes Factor Surprise is related to but different from Shannon Surprise. In two hypothetical experiments, we make testable predictions for physiological indicators that dissociate the Bayes Factor Surprise from Shannon Surprise. The theoretical insight of casting various approaches as surprise-based learning, as well as the proposed online algorithms, may be applied to the analysis of animal and human behavior, and to reinforcement learning in non-stationary environments.

\section{Introduction}

\label{intro}

Animals, humans, and similarly reinforcement learning agents may safely assume that the world is stochastic and stationary during some intervals of time interrupted by change points.
The position of leafs on a tree, a stock market index, or the time it takes to travel from A to B in a crowded city is often well captured by stationary stochastic processes for extended periods of time.
Then sudden changes may happen, such that the distribution of leaf positions becomes different due to a storm, the stock market index is affected by the enforcement of a new law, or a blocked road causes additional traffic jams.
The violation of an agent's expectation caused by such sudden changes is perceived by the agent as surprise, which can be seen as a measure of how much the agent's current belief differs from reality.

Surprise, with its physiological manifestations in pupil dilation \citep{preuschoff2011pupil, nassar2012rational} and EEG signals \citep{modirshanechi2019trial, ostwald2012evidence, mars2008trial}, is believed to modulate learning, potentially through the release of specific neurotransmitters
\citep{angela2005uncertainty, gerstner2018eligibility}, so as to allow animals and humans to adapt quickly to sudden changes.
The quick adaptation to novel situations has been demonstrated in a variety of learning experiments \citep{nassar2012rational, nassar2010approximately, behrens2007learning, angela2005uncertainty, glaze2015normative, heilbron2019confidence}.
The bulk of computational work on surprise-based learning can be separated into two groups.
Studies in the field of computational neuroscience have focused on biological plausibility with little emphasis on the accuracy of learning \citep{nassar2012rational, angela2005uncertainty, nassar2010approximately, faraji2018balancing, friston2017active, schwartenbeck2013exploration, friston2010free, behrens2007learning, bogacz2017tutorial, ryali2018demystifying},
whereas exact and approximate Bayesian online methods \citep{adams2007bayesian,fearnhead2007line} for change point detection and parameter estimation have been developed without any focus on biological plausibility
\citep{aminikhanghahi2017survey, wilson2010bayesian, cummings2018differentially, lin2017sharp, masegosa2017bayesian}.

In this work, we take a top-down approach to surprise-based learning.
We start with a generative model of change points similar to the one that has been the starting point of multiple experiments \citep{nassar2012rational, nassar2010approximately, behrens2007learning, angela2005uncertainty, glaze2015normative, heilbron2019confidence, findling2019imprecise}.
We demonstrate that Bayesian inference on such a generative model can be interpreted as modulation of learning by surprise;
we show that this modulation leads to a natural definition of surprise which is different, but closely related to Shannon Surprise \citep{shannon1948mathematical}.
Moreover, we derive three novel approximate online algorithms with update rules that inherit the surprise-modulated adaptation rate of exact Bayesian inference.
The overall goal of the present study is to give a Bayesian interpretation for surprise-based learning in the brain, and to find approximate methods that are computationally efficient and biologically plausible while maintaining the learning accuracy at a high level.
As a by-product, our approach provides theoretical insights on commonalities and differences among existing surprise-based and approximate Bayesian approaches.
Importantly, our approach makes specific experimental predictions.

In the Results section, we first introduce the generative model, and then we present our surprise-based interpretation of Bayesian inference and our three approximate algorithms.
Next, we use simulations to compare our algorithms with existing ones on two different tasks inspired by and closely related to real experiments
\citep{nassar2010approximately,nassar2012rational, behrens2007learning, mars2008trial, ostwald2012evidence}.
At the end of the Results section, we formalize two experimentally testable predictions of our theory and illustrate them with simulations.
A brief review of related studies as well as a few directions for further work are supplied in the Discussion section.

\section{Results}
\label{Results}

In order to study learning in an environment that exhibits occasional and abrupt changes, we consider a hierarchical generative model (\autoref{fig:gen_model}A) in discrete time, similar to existing model environments \citep{nassar2012rational, nassar2010approximately, behrens2007learning, angela2005uncertainty}.
At each time point $t$, the observation $Y_t = y$ comes from a distribution with the time-invariant likelihood $P_Y(y|\theta)$ parameterized by $\Theta_t = \theta$,
where both $y$ and $\theta$ can be multi-dimensional.
In general, we indicate random variables by capital letters, and values by small letters.
Whenever there is no risk of ambiguity, we drop the explicit notation of random variables to simplify notation.
Abrupt changes of the environment correspond to sudden changes of the parameter $\theta_t$.
At every time $t$, there is a change probability $p_c \in (0,1)$ for the parameter $\theta_t$ to be drawn from its prior distribution $\pi^{(0)}$ independently of its previous value, and a probability $1 - p_c$ to stay the same as $\theta_{t-1}$.
A change at time $t$ is specified by the event $C_t = 1$; otherwise $C_t = 0$.
Therefore, the generative model can be formally defined, for any $T \geq 1$, as a joint probability distribution over $\Theta_{1:T} \equiv (\Theta_1, \ldots, \Theta_T)$, $C_{1:T}$, and $Y_{1:T}$ as
\begin{equation}
  \label{Eq:joint_GM}
  \textbf{P}(c_{1:T},\theta_{1:T},y_{1:T}) = \textbf{P}(c_1) \textbf{P}(\theta_1)  \textbf{P}(y_1 | \theta_1) \prod_{t=2}^T \textbf{P}(c_t) \textbf{P}(\theta_t|c_t,\theta_{t-1}) \textbf{P}(y_t|\theta_t) \, ,
\end{equation}
where $\textbf{P}(\theta_1) = \pi^{(0)}(\theta_1)$, $\textbf{P}(c_1) = \delta (c_1 - 1)$, and

\begin{align}
& \textbf{P}(c_t) = \text{Bernoulli}(c_t; p_c)\, , \label{Eq:GenModel} \\
& \textbf{P}(\theta_t|c_t, \theta_{t-1})= \left\{
  \begin{array}{lr}
    \delta (\theta_t - \theta_{t-1}) & \text{if} \quad c_t = 0\, ,\\
     \pi^{(0)}(\theta_t) & \text{if} \quad c_t = 1\, ,
  \end{array}
\right. \label{Eq:GenModel2}\\
& \textbf{P}(y_t|\theta_t) = P_{Y}(y_t | \theta_t)\, . \label{Eq:GenModel3}
\end{align}
$\textbf{P}$ stands for either probability density function (for the continuous variables) or probability mass function (for the discrete variables), and $\delta$ is the Dirac or Kronecker delta distribution, respectively.

Given a sequence of observations $y_{1:t}$, the \emph{agent's belief} $\pi^{(t)}(\theta)$ about the parameter $\theta$ at time $t$ is defined as the posterior probability distribution $\textbf{P}(\Theta_t=\theta|y_{1:t})$.
In the online learning setting studied here, the agent's goal is to update the belief $\pi^{(t)}(\theta)$ to the new belief $\pi^{(t+1)}(\theta)$, or an approximation thereof, upon observing $y_{t+1}$.

A simplified real-world example of such an environment is illustrated in \autoref{fig:gen_model}B.
Imagine that every day a friend of yours meets you at the coffee shop, starting after work from her office (\autoref{fig:gen_model}B left).
To do so, she needs to cross a river via a bridge.
The time of arrival of your friend (i.e. $y_t$) exhibits some variability, due to various sources of stochasticity (e.g. traffic and your friend's daily workload), but it has a stable average over time (i.e. $\theta_t$).
However, if a new bridge is opened, your friend arrives earlier, since she no longer has to take detour (\autoref{fig:gen_model}B right).
The moment of opening the new bridge is indicated by $c_{t+1} = 1$ in our framework, and the sudden change in the average arrival time of your friend by a sudden change from $\theta_t$ to $\theta_{t+1}$.
Even without any explicit discussion with your friend about this situation and only by observing her actual arrival time, you can notice the abrupt change and hence adapt your schedule to the new situation.

\subsection{Online Bayesian inference modulated by surprise}
\label{recbayesinf}
\begin{figure}
    \centering
\includegraphics{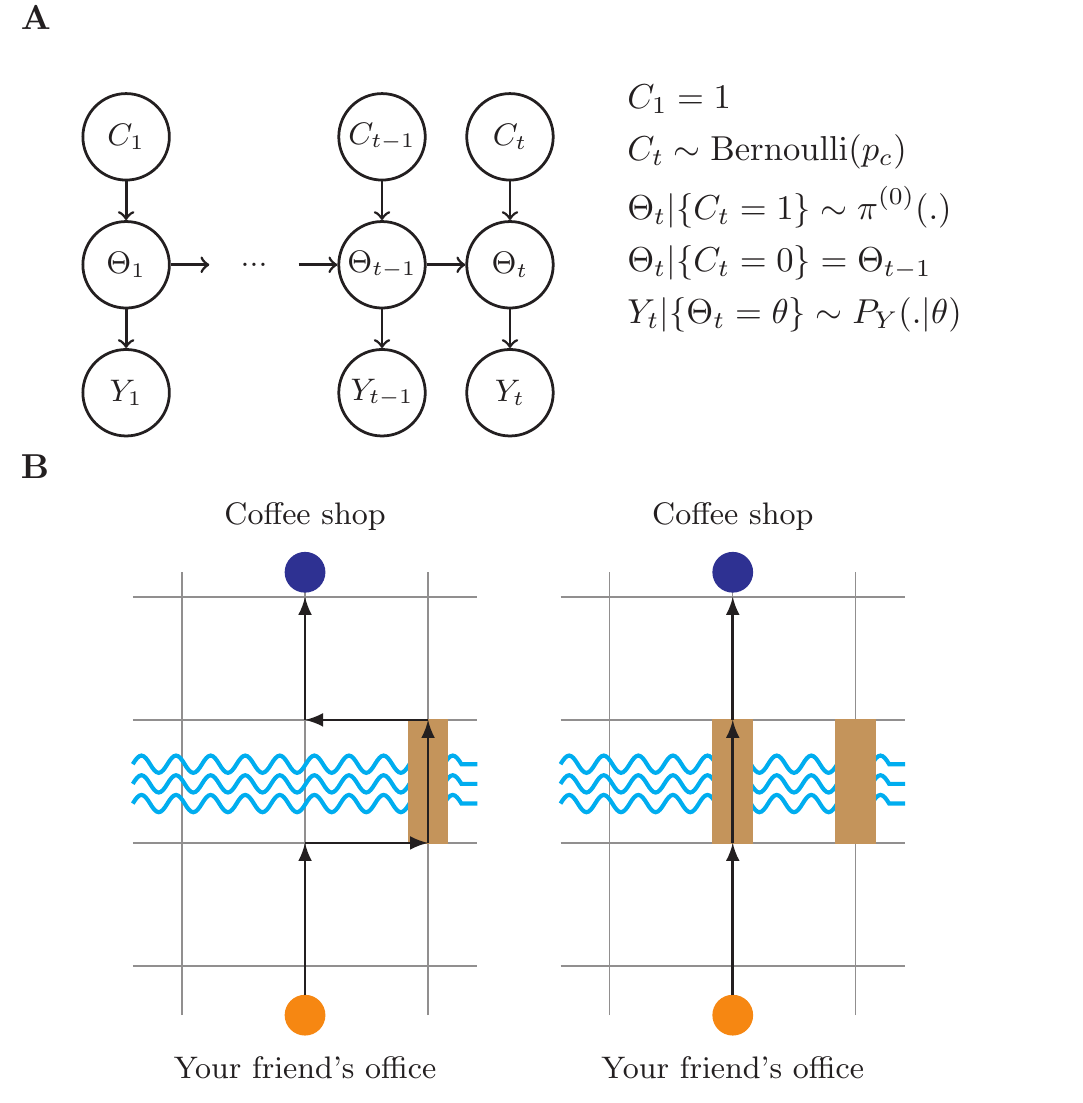}
    \caption{\textbf{Non-stationary environment.}
\textbf{A}. The generative model.
At each time point $t$ there is a probability $p_c \in (0,1)$ for a change in the environment.
When there is a change in the environment, i.e. $C_t = 1$, the parameter $\Theta_t$ is drawn from its prior distribution $\pi^{(0)}$, independently of its previous value.
Otherwise the value of $\Theta_t$ retains its value from the previous time step $t-1$.
Given a parameter value $\Theta_t = \theta$, the observation $Y_t=y_t$ is drawn from a probability distribution $P_Y(y_t|\theta)$.
We indicate random variables by capital letters, and values by small letters.
\textbf{B}. Example of a non-stationary environment.
Your friend meets you every day at the coffee shop (blue dot) starting after work from her office (orange dot) crossing a river.
The time of arrival of your friend is the observed variable $Y_t$, which due to the traffic or your friend's workload may exhibit some variability, but has a stable expectation (i.e. $\theta$).
If, however, a new bridge is opened (i.e. $C_{t} = 1$ where $t$ is the moment of change), your friend no longer needs to take a detour.
There is, then, a sudden change
in her observed daily arrival times.
    }\label{fig:gen_model}
\end{figure}

According to the definition of the hierarchical generative model (\autoref{fig:gen_model}A and \autoref{Eq:joint_GM} to \autoref{Eq:GenModel3}), the value $y_{t+1}$ of the observation at time $t+1$ depends only on the parameters $\theta_{t+1}$, and is (given $\theta_{t+1}$) independent of earlier observations and earlier parameter values.
We exploit this Markovian property and update, using Bayes' rule, the belief $\pi^{(t)}(\theta)\equiv \textbf{P}(\Theta_{t}=\theta|y_{1:t})$ at time $t$ to the new belief at time $t+1$
\begin{equation}
  \label{Eq:Belief_Def_2}
  \pi^{(t+1)}(\theta) = \frac{ P_Y(y_{t+1}|\theta) \textbf{P}(\Theta_{t+1}=\theta| y_{1:t})}{\textbf{P}(y_{t+1}|y_{1:t})} \, .
\end{equation}
So far, \autoref{Eq:Belief_Def_2} remains rather abstract.
The aim of this section is to rewrite it in the form of a surprise-modulated recursive update.
The first term in the numerator of \autoref{Eq:Belief_Def_2} is the likelihood of the current observation given the parameter $\Theta_{t+1}=\theta$,
and the second term is the agent's estimated probability distribution of $\Theta_{t+1}$ before observing $y_{t+1}$.
Because there is always the possibility of an abrupt change, the second term is not the agent's previous belief $\pi^{(t)}$, but $\textbf{P}(\Theta_{t+1}=\theta|y_{1:t}) = (1-p_c) \pi^{(t)}(\theta) + p_c \pi^{(0)}(\theta)$.
As a result, it is possible to find a recursive formula for updating the belief.
For the derivation of this recursive rule, we define the following terms.
\begin{mydef}
    The probability or density (for discrete and continuous variables respectively) of observing $y$ with a belief $\pi^{(t')}$ is denoted as
\begin{equation}
  \label{Eq:p_y_t}
    P(y; \pi^{(t')}) = \int P_Y(y|\theta) \pi^{(t')}(\theta) d\theta \, .
\end{equation}
\end{mydef}
Note that if $\pi$ is the exact Bayesian belief defined as above in \autoref{Eq:Belief_Def_2}, then $P(y; \pi^{(t')}) = \textbf{P}(Y_{t'+1} = y|y_{1:t'}, c_{t'+1} = 0)$.
In Section \ref{contrib} we will use also $P(y; \approxs{\pi}^{(t')})$ for an arbitrary $\approxs{\pi}^{(t')}$.
Two particularly interesting  cases of \autoref{Eq:p_y_t} are $P(y_{t+1}; \pi^{(t)})$, i.e. the probability of a new observation $y_{t+1}$ with the current belief $\pi^{(t)}$, and $P(y_{t+1}; \pi^{(0)})$, i.e. the probability of a new observation $y_{t+1}$ with the prior belief $\pi^{(0)}$.
\begin{mydef}
The ``Bayes Factor Surprise'' $\textbf{S}_{\mathrm{BF}}$ of the observation $y_{t+1}$ is defined as the ratio of the probability of observing $y_{t+1}$ given  $c_{t+1} = 1$ (i.e. when there is a change), to the probability of observing $y_{t+1}$ given  $c_{t+1} = 0$ (i.e. when there is no change), i.e.
\begin{equation}
    \label{Eq:S_GM}
    \textbf{S}_{\mathrm{BF}}(y_{t+1}; \pi^{(t)}) = \frac{P(y_{t+1}; \pi^{(0)})}{P(y_{t+1}; \pi^{(t)})}\, .
\end{equation}
\end{mydef}
This definition of surprise measures how much more probable the current observation is under the naive prior $\pi^{(0)}$ relative to the current belief $\pi^{(t)}$ (see
the Discussion section for further interpretation).
This probability ratio is the Bayes factor \citep{kass1995bayes, efron2016computer} that tests the prior belief $\pi^{(0)}$ against the current belief $\pi^{(t)}$.
We emphasize that our definition of surprise is not arbitrary, but essential in order to write the exact inference in \autoref{Eq:Belief_Def_2} on the generative model in the compact recursive form indicated in the Proposition that follows.
Moreover, as we show later, this term can be identified in multiple learning algorithms (among them \cite{nassar2010approximately, nassar2012rational}), but it has never been interpreted as a surprise measure.
In the following sections we establish the generality of this computational mechanism and identify it as a common feature of many learning algorithms.
\begin{mydef}
Under the assumption of no change $c_{t+1} = 0$, and using the most recent belief $\pi^{(t)}$ as prior, the exact Bayesian update for $\pi^{(t+1)}$ is denoted as
\begin{equation}
  \label{Eq:pi_B}
  \pi_B^{(t+1)}(\theta) = \frac{P_Y(y_{t+1}|\theta) \pi^{(t)}(\theta) } {P(y_{t+1}; \pi^{(t)})}\, .
\end{equation}
\end{mydef}
$\pi_B^{(t+1)}(\theta)$ describes the incorporation of the new information into the current belief via Bayesian updating.

\begin{mydef}
The ``Surprise-Modulated Adaptation Rate'' is a function $\gamma: \mathbb{R}^+ \times \mathbb{R}^+ \to [0,1]$ specified as
\begin{equation}
    \label{Eq:Gamma_def}
    \gamma ( \text{S}, m ) = \frac{m \text{S}}{1 + m \text{S}}\, ,
\end{equation}
where $\text{S} \geq 0$ is a surprise value, and $m \geq 0 $ is a parameter controlling the effect of surprise on learning.
\end{mydef}

Using the above definitions and \autoref{Eq:Belief_Def_2}, we have for the generative model of \autoref{fig:gen_model}A and \autoref{Eq:joint_GM} to \autoref{Eq:GenModel3} the following Proposition.

\begin{mylemma}
Exact Bayesian inference on the generative model is equivalent to the recursive update rule
\begin{equation}
  \label{Eq:Bayesian_Rec_Formula}
\pi^{(t+1)}(\theta) = \Big(1 - \gamma \Big( \textbf{S}_{\mathrm{BF}}^{(t+1)}, \frac{p_c}{1-p_c} \Big) \Big) \pi^{(t+1)}_B(\theta) + \gamma \Big( \textbf{S}_{\mathrm{BF}}^{(t+1)}, \frac{p_c}{1-p_c} \Big) P(\theta|y_{t+1})\, ,
\end{equation}
where $\textbf{S}_{\mathrm{BF}}^{(t+1)} = \textbf{S}_{\mathrm{BF}}(y_{t+1}; \pi^{(t)})$ is the Bayes Factor Surprise and
\begin{equation}
    \label{Eq:cond_prob_ThetaY}
  P(\theta|y_{t+1})
= \frac{P_Y(y_{t+1}|\theta) \pi^{(0)}(\theta)} {P(y_{t+1}; \pi^{(0)})}\,
\end{equation}
is the posterior if we take $y_{t+1}$ as the only observation.
\end{mylemma}
The proposition indicates that the exact Bayesian inference on the generative model discussed above (\autoref{fig:gen_model}) leads to an explicit trade-off between (i) integrating a new observation $y^{\text{new}}$ (corresponding to $y_{t+1}$) with the old belief $\pi^{\text{old}}$ (corresponding to $\pi^{(t)}$) into a distribution $\pi^{\text{integration}}$ (corresponding to $\pi_B^{(t+1)}$) and
(ii) forgetting the past observations, so as to restart with the belief $\pi^{\text{reset}}$ (corresponding to $P(\theta|y_{t+1})$) which relies only on the new observation and the prior $\pi^{(0)}$
\begin{equation}
    \label{Eq:tradeoff}
    \pi^{\text{new}}(\theta) =
    (1 - \gamma) \, \pi^{\text{integration}}(\theta| y^{\text{new}}, \pi^{\text{old}}) +
    \gamma \, \pi^{\text{reset}}(\theta | y^{\text{new}}, \pi^{(0)}) .
\end{equation}
This trade-off is governed by a \emph{surprise-modulated adaptation rate} $\gamma ( \text{S}, m ) \in [0,1]$, where $\text{S} = \textbf{S}_{\mathrm{BF}} \geq 0$ (corresponding to the Bayes Factor Surprise) can be interpreted as the surprise of the most recent observation, and $m = \frac{p_c}{1-p_c} \geq 0 $ is a parameter controlling the effect of surprise on learning.
Because the parameter of modulation $m$ is equal to $\frac{p_c}{1-p_c}$, for a fixed value of surprise $S$, the adaptation rate $\gamma$ is an increasing function of $p_c$.
Therefore, in more volatile environments, the same value of surprise $S$ leads to a higher adaptation rate than in a less volatile environment;
in the case of $p_c \to 1$, any surprise value leads to full forgetting, i.e. $\gamma=1$.

As a conclusion, our first main result is that a split as in \autoref{Eq:tradeoff} with a weighting factor (``adaptation rate'' $\gamma$) as in \autoref{Eq:Gamma_def} is exact and always possible for the class of environments defined by our hierarchical generative model.
This surprise-modulation gives rise to specific testable experimental predictions discussed later.

\subsection{Approximate algorithms modulated by surprise}

\label{contrib}

Despite the simplicity of the recursive formula in \autoref{Eq:Bayesian_Rec_Formula}, the updated belief $\pi^{(t+1)}$ is generally not in the same family of distributions as the previous belief $\pi^{(t)}$, e.g. the result of averaging two normal distributions is not a normal distribution.
Hence it is in general impossible to find a simple and exact update rule for e.g. some sufficient statistic.
As a consequence, the memory demands for $\pi^{(t+1)}$ scale linearly in time, and updating $\pi^{(t+1)}$ using $\pi^{(t)}$ needs $\mathcal{O}(t)$ operations.
In the following sections, we investigate three approximations (Algo. 1-3) that have simple update rules and finite memory demands, so that the updated belief remains tractable over a long sequence of observations.

As our second main result, we show that all three novel approximate algorithms inherit the surprise-modulated adaptation rate from the exact Bayesian approach, i.e. \autoref{Eq:Gamma_def} and \autoref{Eq:tradeoff}.
The first algorithm adapts an earlier algorithm of surprise minimization learning (SMiLe, \cite{faraji2018balancing}) to variational learning.
We refer to our novel algorithm as Variational SMiLe and abbreviate it by VarSMiLe (see Algo. 1).
The second algorithm is based on message passing \citep{adams2007bayesian} restricted to a finite number of messages $N$ .
We refer to this algorithm as MP$N$ (see Algo. 2).
The third algorithm uses the ideas of particle filtering \citep{gordon1993novel} for an efficient approximation for our hierarchical generative model.
We refer to our approximate algorithm as Particle Filtering with $N$ particles and abbreviate it by pf$N$ (see Algo. 3).
All algorithms are computationally efficient, have finite memory demands and are biologically plausible;
Particle Filtering has possible neuronal implementations \citep{kutschireiter2017nonlinear, shi2009neural, huang2014neurons, legenstein2014ensembles}, MP$N$ can be seen as a greedy version of pf$N$ without sampling, and Variational SMiLe may be implemented by neo-Hebbian \citep{lisman2011neohebbian, gerstner2018eligibility} update rules.
Simulation results show that the performance of the three approximate algorithms is comparable to and more robust across environments than other state-of-the-art approximations.

\subsubsection{Variational SMiLe Rule (Algo. 1)}
\label{var_smiles}
\newcommand{\varapprox}[1]{\hat{#1}}
A simple heuristic approximation to keep the updated belief in the same family as the previous beliefs consists in applying the weighted averaging of the exact Bayesian update rule (\autoref{Eq:Bayesian_Rec_Formula}) to the logarithm of the beliefs rather than the beliefs themselves, i.e.
\begin{equation}
\label{Eq:Var_SMiLe_Log_Rec_2nd_Ver}
\text{log} \big( \approxs{\pi}^{(t+1)}(\theta) \big) = (1-\gamma_{t+1}) \text{ log} \big( \approxs{\pi}_B^{(t+1)}(\theta) \big) + \gamma_{t+1} \text{ log} \big( P(\theta|y_{t+1}) \big) + \text{Const.}\, ,
\end{equation}
where $\gamma_{t+1} = \gamma \big(\textbf{S}_{\mathrm{BF}}(y_{t+1}; \approxs{\pi}^{(t)}) , m \big)$ is given by \autoref{Eq:Gamma_def} with a free parameter $m>0$ which can be tuned to each environment.
By doing so, we still have the explicit trade-off between two terms as in \autoref{Eq:Bayesian_Rec_Formula}, but in the logarithms;
yet an advantageous consequence of averaging over logarithms is that, if the likelihood function $P_Y$ is in the exponential family, and if the initial belief $\pi^{(0)}$ is its conjugate prior, then $\approxs{\pi}^{(t+1)}$ and $\pi^{(0)}$ are members of the same family.
In this particular case, we arrive at a simple update rule for the parameters of $\approxs{\pi}^{(t+1)}$ (see Algorithm \ref{Alg:Var_SMiLe} for pseudocode and Methods for details).
As it is common in variational approaches \citep{beal2003variational}, the price of this simplicity is that, except for the trivial cases of $p_c=0$ and $p_c=1$, there is no evidence other than simulations that the update rule of \autoref{Eq:Var_SMiLe_Log_Rec_2nd_Ver} will end up at an approximate belief close to the exact Bayesian belief.

One way to interpret the update rule of \autoref{Eq:Var_SMiLe_Log_Rec_2nd_Ver} is to rewrite it as the solution of a constraint optimization problem.
The new belief $\approxs{\pi}^{(t+1)}$ is a variational approximation of the Bayesian update $\approxs{\pi}_B^{(t+1)}$ (see Methods)
\begin{equation}
\begin{aligned}
\label{Eq:Var_SMiLe_Min_1_2nd_Ver}
\approxs{\pi}^{(t+1)}(\theta) &= \arg \min_q \textbf{D}_{KL} \big[ q(\theta) || \approxs{\pi}_B^{(t+1)}(\theta) \big],
\end{aligned}
\end{equation}
with a family of functions $q(\theta)$ constrained by the Kullback-Leibler divergence
\begin{equation}
\begin{aligned}
\label{Eq:Var_SMiLe_Min_1_2nd_Ver_Const}
\textbf{D}_{KL} \big[ q(\theta) || P(\theta|y_{t+1}) \big] \leq B_{t+1}\, ,
\end{aligned}
\end{equation}
where the bound $B_{t+1} \in \big[ 0, \textbf{D}_{KL}[ \approxs{\pi}_B^{(t+1)}(\theta) || P(\theta|y_{t+1}) ] \big]$ is a decreasing function of the Bayes Factor surprise $\textbf{S}_{\mathrm{BF}}(y_{t+1}; \approxs{\pi}^{(t)})$ (see Methods for proof), and $P(\theta|y_{t+1})$ is given by \autoref{Eq:cond_prob_ThetaY}.

Because of the similarity of the constraint optimization problem in \autoref{Eq:Var_SMiLe_Min_1_2nd_Ver} and \autoref{Eq:Var_SMiLe_Min_1_2nd_Ver_Const} to the Surprise Minimization Learning rule ``SMiLe'' \citep{faraji2018balancing}, we call this algorithm ``Variational Surprise Minimization Learning'' rule, or in short ``Variational SMiLe'' rule.
The differences between SMiLe and Variational SMiLe are discussed in the Methods section.

Our variational method, and particularly its surprise-modulated adaptation rate, is complementary to earlier studies \citep{ozkan2013marginalized, masegosa2017bayesian} in machine learning which assumed different generative models and used additional assumptions and different approaches for deriving the learning rule.

\begin{algorithm}
\caption{Pseudocode for Variational SMiLe (exponential family)}
\label{Alg:Var_SMiLe}
\begin{algorithmic}[1]
  \State Specify $P_Y(y|\theta)$, $\textbf{P}_{\pi}\big(\Theta = \theta; \chi, \nu \big)$, and $\phi(y)$ \newline
  where $P_Y \in \{\text{exponential family} \}$, $\textbf{P}_{\pi} \in \{\text{conjugate priors of } P_Y\}$ parametrized by $\chi$ and $\nu$, and $\phi(y)$ is the sufficient statistic.
  \State Specify $m$.
\State Initialize $\chi^{(0)}$, $\nu^{(0)}$, and $t \gets 0$.
\While {the sequence is not finished}
\State Observe $y_{t+1}$
\MyStatex{Surprise}
\State Compute $\textbf{S}_{\mathrm{BF}}(y_{t+1}; \approxs{\pi}^{(t)})$ using \autoref{Eq:S_GM_exp}
\MyStatex{Modulation factor}
\State Compute $\gamma_{t+1} = \gamma \big( \textbf{S}_{\mathrm{BF}}(y_{t+1}; \approxs{\pi}^{(t)}), m \big)$
\MyStatex{Updated belief}
\State \boxed{\chi^{(t+1)} \gets (1-\gamma_{t+1}) \chi^{(t)} + \gamma_{t+1} \chi^{(0)} + \phi(y_{t+1}) }
\State \boxed{\nu^{(t+1)} \gets (1-\gamma_{t+1}) \nu^{(t)} + \gamma_{t+1} \nu^{(0)} + 1 \quad \: \: \: \: \: \: \: \:}
\State $\approxs{\pi}^{(t+1)}(\theta) = \textbf{P}_{\pi}\big(\Theta = \theta; \chi^{(t+1)}, \nu^{(t+1)}\big)$
\MyStatex{Iterate}
\State $t \gets t+1$
\EndWhile
\end{algorithmic}
\end{algorithm}

\subsubsection{Message-Passing $N$ (Algo. 2)}
\label{sec:adam_mac_variants}

For a hierarchical generative model similar to ours, a message passing algorithm has been used to perform exact Bayesian inference \citep{adams2007bayesian}, where the algorithm's memory demands and computational complexity scale linearly in time $t$.
In this section, we first explain the idea of the message passing algorithm of \cite{adams2007bayesian} and its relation to our Proposition.
We then present our approximate version of this algorithm which has a constant (in time) computational complexity and memory demands.

The history of change points up to time $t$ is a binary sequence, e.g. $c_{1:t} = \{ 1,0,0,1,0,1,1 \}$, where the value $1$ indicates a change in the corresponding time step.
Following the idea of \cite{adams2007bayesian}, we define the random variable $R_t = \min \{ n \in \mathbb{N} : C_{t-n+1} = 1  \}$ in order to describe the time since the last change point, which takes values between 1 to $t$.
We can write the exact Bayesian expression for $\pi^{(t)}(\theta)$ by marginalizing $\textbf{P}(\Theta_{t} = \theta, r_t | y_{1:t})$ over the $t$ possible values of $r_t$ in the following way
\begin{equation}
\begin{aligned}
  \label{Eq:adam_mac_MPA_0}
  \pi^{(t)}(\theta) &= \sum_{k=0}^{t-1} \textbf{P}(R_t = t-k |y_{1:t}) \textbf{P}(\Theta_{t} = \theta | R_t = t-k, y_{1:t}).
\end{aligned}
\end{equation}
For consistency with Algorithm 3 (i.e. Particle Filtering), we call each term in the sum of \autoref{Eq:adam_mac_MPA_0} a ``particle'', and denote as $\pi^{(t)}_k(\theta) = \textbf{P}(\Theta_{t} = \theta | R_t = t-k, y_{1:t})$ the belief of the particle corresponding to $R_t = t-k$, and $w^{(k)}_t = \textbf{P}(R_t = t-k |y_{1:t})$ its corresponding weight at time $t$, i.e.
\begin{equation}
\begin{aligned}
  \label{Eq:adam_mac_MPA}
  \pi^{(t)}(\theta) &= \sum_{k=0}^{t-1} w^{(k)}_t \pi^{(t)}_k(\theta).
\end{aligned}
\end{equation}
For each particle, the term
$\pi^{(t)}_k(\theta)$
is simple to compute, because when $r_t$ is known, inference depends only on the observations after the last change point.
Therefore, the goal of online inference is to find an update rule for the evolution of the weights $w^{(k)}_t$ over time.

We can apply the exact update rule of our Proposition (\autoref{Eq:Bayesian_Rec_Formula}) to the belief expressed in the form of \autoref{Eq:adam_mac_MPA}.
Upon each observation of a new sample $y_{t+1}$, a new particle is generated and added to the set of particles, corresponding to $P(\theta|y_{t+1})$ (i.e. $\pi^{reset}$), modelling the possibility of a change point occurring at $t+1$.
According to the proposition, the weight of the new particle (i.e. $k=t$) is equal to
\begin{equation}
\begin{aligned}
  \label{Eq:MP_w_newparticle}
  & w^{(t)}_{t+1} = \gamma_{t+1},
\end{aligned}
\end{equation}
where $\gamma_{t+1} = \gamma \big(\textbf{S}_{\mathrm{BF}}(y_{t+1}; \pi^{(t)}), \frac{p_c}{1-p_c} \big)$ (cf. \autoref{Eq:Gamma_def}).
The other $t$ particles coming from $\pi^{(t)}$ corresponds to $\pi_B^{(t+1)}$ (i.e. $\pi^{integration}$) in the proposition.
The update rule (see Methods for derivation) for the weights of these particles (i.e. $ 0 \leq k \leq t-1$) is \begin{equation}
\begin{aligned}
  \label{Eq:MP_w_oldparticles}
  & w^{(k)}_{t+1} = (1-\gamma_{t+1}) w^{(k)}_{B,t+1}  = (1-\gamma_{t+1}) \frac{ P(y_{t+1}; \pi_k^{(t)}) } { P(y_{t+1}; \pi^{(t)}) } w^{(k)}_t.
\end{aligned}
\end{equation}
So far, we used the idea of \citep{adams2007bayesian} to write the belief as in \autoref{Eq:adam_mac_MPA} and used our proposition to arrive at the surprise-modulated update rules in \autoref{Eq:MP_w_newparticle} and \autoref{Eq:MP_w_oldparticles}.

The computational complexity and memory requirements of the complete message passing algorithm increase linearly with time $t$. To deal with this issue and to have a constant computation and memory demands over time, we implemented a message passing algorithm of the form of \autoref{Eq:adam_mac_MPA} to \autoref{Eq:MP_w_oldparticles}, but with a fixed number $N$ of particles, chosen as those with the highest weights $w^{(k)}_{t}$.
Therefore, our second algorithm adds a new approximation step to the full message passing algorithm of \cite{adams2007bayesian}:
Whenever $t>N$, after adding the new particle with the weight as in \autoref{Eq:MP_w_newparticle} and updating the previous weights as in \autoref{Eq:MP_w_oldparticles}, we discard the particle with the smallest weight (i.e. set its weight equal to 0), and renormalize the weights.
By doing so, we always keep the number of particles with non-zero weights equal to $N$.
Note that, for $t \leq N$, our algorithm is exact, and identical to the message passing algorithm of \citep{adams2007bayesian}.
We call our modification of the message passing algorithm of \cite{adams2007bayesian} ``Message Passing $N$'' and abbreviate it by ``MP$N$''.

To deal with the computational complexity and memory requirements, one may alternatively keep only the particles with weights greater than a cut-off threshold \citep{adams2007bayesian}.
However, such a constant cut-off leads to a varying number (smaller or equal to $t$) of particles in time.
Our approximation MP$N$ can therefore be seen as a variation of the thresholding algorithm in \cite{adams2007bayesian} with fixed number of particles $N$, and hence a variable cut-off threshold.
The work of \cite{fearnhead2007line} follows the same principle as \cite{adams2007bayesian}, but employs stratified resampling to eliminate particles with negligible weights, in order to reduce the total number of particles.
Their resampling algorithm involves solving a complicated non-linear equation at each time step, which makes it unsuitable for a biological implementation.
In addition, we experienced that in some cases, the small errors introduced in the resampling step of the algorithm of \cite{fearnhead2007line} accumulated and led to a worse performance than our MP$N$ algorithm which simply keeps the $N$ particles with the highest weight at each time step.

For the case where the likelihood function $P_Y(y|\theta)$ is in the exponential family and $\pi^{(0)}$ is its conjugate prior, the resulting algorithm of MP$N$ has a simple update rule for the belief parameters (see Algorithm \ref{Alg:MPN} and Methods for details).
For the sake of comparison, we also implemented in our simulations the full message passing algorithm of \cite{adams2007bayesian} with an almost zero cut-off (machine precision), which we consider as our benchmark ``Exact Bayes'', as well as the stratified optimal resampling algorithm of \cite{fearnhead2007line}, called ``SOR$N$''.

\begin{algorithm} \caption{Pseudocode for MP$N$ (exponential family)}
  \label{Alg:MPN} \begin{algorithmic}[1]
  \State Specify $P_Y(y|\theta)$, $\textbf{P}_{\pi}\big(\Theta = \theta; \chi, \nu \big)$, and $\phi(y)$ \newline
  where $P_Y \in \{\text{exponential family} \}$, $\textbf{P}_{\pi} \in \{\text{conjugate priors of } P_Y\}$ parametrized by $\chi$ and $\nu$, and $\phi(y)$ is the sufficient statistic.
  \State Specify $m=p_c/(1-p_c)$, and $N$.
  \State Initialize $\chi_1^{(0)}$, $\nu_1^{(0)}$, $w^{(1)}_{0} = 1$ and $t \gets 0$.
  \State Until $N=t$, do the exact message passing algorithm of \autoref{Eq:MP_w_oldparticles} and \autoref{Eq:MP_w_newparticle}
  \While {the sequence is not finished and $N < t$}
  \State Observe $y_{t+1}$
  \MyStatex{Surprise per particle $i$}
  \For {$i \in \{1, ..., N \}$}
  \State Compute $\textbf{S}_{\mathrm{BF}}(y_{t+1}, \approxs{\pi}_i^{(t)})$ using \autoref{Eq:S_GM_exp} with $\chi_i^{(t)}$, $\nu_i^{(t)}$ \EndFor
  \MyStatex{Global surprise}
  \State Compute $\textbf{S}_{\mathrm{BF}}(y_{t+1}, \approxs{\pi}^{(t)})$ as the weighted ($w^{(i)}_{t}$) harmonic mean of $\textbf{S}_{\mathrm{BF}}(y_{t+1}, \approxs{\pi}_i^{(t)})$
  \MyStatex{Modulation factor}
  \State Compute $\gamma_{t+1} = \gamma \big( \textbf{S}_{\mathrm{BF}}(y_{t+1}, \approxs{\pi}^{(t)}) , m \big)$
  \MyStatex{Weight per particle $i$}
  \For {$i \in \{1, ..., N \}$}
  \State Compute the Bayesian weight $w_{B, t+1}^{(i)}$ using \autoref{Eq:wB}
\State \boxed{w_{t+1}^{(i)} \gets (1-\gamma_{t+1})w_{B, t+1}^{(i)}}
\EndFor
  \MyStatex{Weight for the new particle}
\State \boxed{w_{t+1}^{(N+1)} \gets \gamma_{t+1}}
  \MyStatex{Updated belief per particle $i$}
  \For {$i \in \{1, ..., N \}$}
  \State $\chi_i^{(t+1)} \gets \chi_i^{(t)} + \phi(y_{t+1})$ and $\nu_i^{(t+1)} \gets \nu_i^{(t)} + 1$
\EndFor
  \State $\chi_{N+1}^{(t+1)} \gets \chi^{(0)} + \phi(y_{t+1})$ and $\nu_{N+1}^{(t+1)} \gets \nu^{(0)}+ 1$
  \MyStatex{Approximation}
  \State Keep the $N$ particles with highest weights among $w_{t+1}^{(1:N+1)}$, rename and normalize their weights
  \MyStatex{Updated belief}
  \State $\approxs{\pi}^{(t+1)}(\theta) = \sum_{i=1}^{N} w_{t+1}^{(i)} \textbf{P}_{\pi}\big(\Theta = \theta; \chi_i^{(t+1)}, \nu_i^{(t+1)}\big)$
  \MyStatex{Iterate}
  \State $t \gets t+1$
  \EndWhile
\end{algorithmic}
\end{algorithm}

\subsubsection{Particle Filtering (Algo. 3)}
\label{pf}

\autoref{Eq:adam_mac_MPA} demonstrates that the exact Bayesian belief $\pi^{(t)}$ can be expressed as a sum of two factors, i.e. as the marginalization of $\textbf{P}(\Theta_{t} = \theta, r_t | y_{1:t})$ over the time since the last change point $r_t$.
Equivalently, one can compute the exact Bayesian belief as the marginalization of $\textbf{P}(\Theta_{t} = \theta, c_{1:t} | y_{1:t})$ over the history of change points $c_{1:t}$, i.e.
\begin{equation}
\begin{aligned}
  \pi^{(t)}(\theta) &= \sum_{c_{1:t}} \textbf{P}(c_{1:t}| y_{1:t}) \textbf{P}(\Theta_{t} = \theta | c_{1:t}, y_{1:t}) \\
  &= \mathbb{E}_{\textbf{P}(C_{1:t} | y_{1:t})} \big[ \textbf{P}(\Theta_{t} = \theta | C_{1:t}, y_{1:t}) \big].
\end{aligned}
\end{equation}
The idea of our third algorithm is to approximate this expectation by particle filtering, i.e. sequential Monte Carlo sampling \citep{gordon1993novel, doucet2000sequential} from $\textbf{P}(C_{1:t} | y_{1:t})$.

We then approximate $\pi^{(t)}$ by
\begin{equation}
\label{Eq:pi_pf_approx}
\approxs{\pi}^{(t)}(\theta) = \sum_{i=1}^{N} w_{t}^{(i)} \approxs{\pi}_i^{(t)}(\theta) = \sum_{i=1}^{N} w_{t}^{(i)} \textbf{P}(\Theta_t = \theta | c_{1:t}^{(i)}, y_{1:t})\, ,
\end{equation}
where $\{ c_{1:t}^{(i)} \}_{i=1}^N$ is a set of $N$ realizations (or samples) of $c_{1:t}$ (i.e. $N$ particles) drawn from a proposal distribution $Q(c_{1:t} | y_{1:t})$, $\{ w_{t}^{(i)} \}_{i=1}^N$ are their corresponding weights at time $t$, and $\approxs{\pi}_i^{(t)}(\theta) = \textbf{P}(\Theta_t = \theta | c_{1:t}^{(i)}, y_{1:t})$ is the approximate belief corresponding to particle $i$.

Upon observing $y_{t+1}$, the update procedure for the approximate belief $\approxs{\pi}^{(t+1)}$ of \autoref{Eq:pi_pf_approx} includes two steps: (i) updating the weights, and (ii) sampling the new hidden state $c_{t+1}$ for each particle.
The two steps are coupled together through the choice of the proposal distribution $Q$, for which, we choose the optimal proposal function \citep{doucet2000sequential} (see Methods).
As a result, given this choice of proposal function, we show (see Methods) that
the first step amounts to
\begin{equation}
\begin{aligned}
  \label{Eq:wB}
& w_{t+1}^{(i)} = (1-\gamma_{t+1})w_{B, t+1}^{(i)} + \gamma_{t+1}w_{t}^{(i)}\, ,\\
& w_{B, t+1}^{(i)} = \frac{P (y_{t+1}; \approxs{\pi}_i^{(t)} ) }{P (y_{t+1}; \approxs{\pi}^{(t)} ) } w_{t}^{(i)}\, ,
\end{aligned}
\end{equation}
where $\gamma_{t+1} = \gamma \big(\textbf{S}_{\mathrm{BF}}(y_{t+1}; \approxs{\pi}^{(t)}) , m \big)$ with $ m = \frac{p_c}{1-p_c}$ (cf. \autoref{Eq:Gamma_def}), and $\{ w_{B, t+1}^{(i)} \}_{i=1}^N$ are the weights corresponding to the Bayesian update $\approxs{\pi}_{B}^{(t+1)}$
of \autoref{Eq:pi_B} (see Methods).
In the second step, we update each particle's history of change points by going from the sequence $\{ c_{1:t}^{(i)} \}_{i=1}^N$ to $\{ c_{1:t+1}^{(i)} \}_{i=1}^N$, for which we always keep the old sequence up to time $t$, and for each particle $i$, we add a new element $c_{t+1}^{(i)} \in \{ 0, 1 \}$ representing no change
$c_{1:t+1}^{(i)} = [ c_{1:t}^{(i)}, 0 ]$ or change $c_{1:t+1}^{(i)} = [ c_{1:t}^{(i)}, 1 ]$.
Note, however, that it is not needed to keep the whole sequences $c_{1:t+1}^{(i)}$ in memory, but instead one can use $c_{t+1}^{(i)}$ to update $\approxs{\pi}_{i}^{(t)}$ to $\approxs{\pi}_{i}^{(t+1)}$.
We sample the new element $c^{(i)}_{t+1}$ from the optimal proposal distribution $Q(c_{t+1}^{(i)}| c_{1:t}^{(i)}, y_{1:t+1})$ \citep{doucet2000sequential},
which
is given by (see Methods)
\begin{equation}
\begin{aligned}
  \label{Eq:proposal_final}
Q(c_{t+1}^{(i)} = 1| c_{1:t}^{(i)}, y_{1:t+1}) = \gamma \Big( \textbf{S}_{\mathrm{BF}}(y_{t+1}; \approxs{\pi}_i^{(t)}), \frac{p_c}{1-p_c} \Big)\, .
\end{aligned}
\end{equation}

Interestingly, the above formulas entail the same surprise modulation and the same trade-off as proposed by the Proposition \autoref{Eq:Bayesian_Rec_Formula}.
For the weight update, there is a trade-off between an exact Bayesian update and keeping the value of the previous time step, controlled by a adaptation rate modulated exactly in the same way as in \autoref{Eq:Bayesian_Rec_Formula}.
Note that in contrast to \autoref{Eq:Bayesian_Rec_Formula}, the trade-off for the particles' weights is not between forgetting and integrating, but between maintaining the previous knowledge and integrating.
However, the change probability (\autoref{Eq:proposal_final}) for sampling is equal to the adaptation rate and is an increasing function of surprise.
As a result, although the weights are updated less for surprising events, a higher surprise causes a higher probability for change, indicated by $c_{t+1}^{(i)} = 1$,
which implies forgetting, because for a particle $i$ with $c_{t+1}^{(i)} = 1$, the associated belief $\approxs{\pi}_i^{(t+1)} = \textbf{P}(\Theta_{t+1} = \theta | c_{t+1}^{(i)}=1 , c_{1:t}^{(i)}, y_{1:t+1})$
is equal to
$\textbf{P}(\Theta_{t+1} = \theta | c_{t+1}^{(i)}=1 , y_{t+1}) = P(\theta|y_{t+1})$ (see \autoref{fig:gen_model}A), which is equivalent to a reset of the belief as in \autoref{Eq:tradeoff}.
In other words, while in MP$N$ and the exact Bayesian inference in Proposition \autoref{Eq:Bayesian_Rec_Formula}, the trade-off between integration and reset is accomplished by adding at each time step a new particle with weight $\gamma_{t+1}$, in Particle Filtering, it is accomplished via sampling.
As a conclusion, the above formulas are essentially the same as the update rules of MP$N$ (c.f. \autoref{Eq:MP_w_oldparticles} and \autoref{Eq:MP_w_newparticle}) and have the same spirit as the recursive update of the Proposition \autoref{Eq:Bayesian_Rec_Formula}.

Equations \ref{Eq:pi_pf_approx} and \ref{Eq:wB} can be applied to the case where the likelihood function $P_Y(y|\theta)$ is in the exponential family and $\pi^{(0)}$ is its conjugate prior.
The resulting algorithm (Algorithm \ref{Alg:PF}) has a particularly simple update rule for the belief parameters (see Methods for details).

The theory of particle filter methods is well established \citep{gordon1993novel, doucet2000sequential, sarkka2013bayesian}.
Particle filters in simpler \citep{brown2009detecting} or more complex \citep{findling2019imprecise} forms have also been employed to explain human behaviour.
Here we derived a simple particle filter for the general case of generative models of \autoref{Eq:GenModel}, \autoref{Eq:GenModel2}, and \autoref{Eq:GenModel3}.
Our main contribution is to show that the use of the optimal proposal distribution in this particle filter leads to a surprise-based update scheme.

\begin{algorithm} \caption{Pseudocode for Particle Filtering (exponential family)}
  \label{Alg:PF}
  \begin{algorithmic}[1]
  \State Specify $P_Y(y|\theta)$, $\textbf{P}_{\pi}\big(\Theta = \theta; \chi, \nu \big)$, and $\phi(y)$ \newline
  where $P_Y \in \{\text{exponential family} \}$, $\textbf{P}_{\pi} \in \{\text{conjugate priors of } P_Y\}$ parametrized by $\chi$ and $\nu$, and $\phi(y)$ is the sufficient statistic.
  \State Specify $m=p_c/(1-p_c)$, $N$, and $N_{\text{thrs}}$
  \State Initialize $\chi^{(0)}$, $\nu^{(0)}$, $w^{(i)}_{0}$ $\forall i \in \{1 ... N \}$, and $t \gets 0$.
  \While {the sequence is not finished}
  \State Observe $y_{t+1}$
  \MyStatex{Surprise per particle $i$}
  \For {$i \in \{1, ..., N \}$}
  \State Compute $\textbf{S}_{\mathrm{BF}}(y_{t+1}; \approxs{\pi}_i^{(t)})$ using \autoref{Eq:S_GM_exp} with $\chi_i^{(t)}$, $\nu_i^{(t)}$
  \EndFor
\MyStatex{Global surprise}
  \State Compute $\textbf{S}_{\mathrm{BF}}(y_{t+1}; \approxs{\pi}^{(t)}) = \big[ \sum_{i=1}^{N}w^{(i)}_{t} [\textbf{S}_{\mathrm{BF}}(y_{t+1}; \approxs{\pi}_i^{(t)})]^{-1}\big]^{-1}$
  \MyStatex{Modulation factor}
  \State Compute $\gamma_{t+1} = \gamma \big( \textbf{S}_{\mathrm{BF}}(y_{t+1}; \approxs{\pi}^{(t)}) , m \big)$
  \MyStatex{Weight per particle $i$}
  \For {$i \in \{1, ..., N \}$}
\State Compute the Bayesian weight $w_{B, t+1}^{(i)}$ using \autoref{Eq:wB}
\State
  \boxed{w_{t+1}^{(i)} \gets (1-\gamma_{t+1})w_{B, t+1}^{(i)} + \gamma_{t+1}w_{t}^{(i)}}
\EndFor
  \MyStatex{Hidden state per particle $i$}
  \For {$i \in \{1, ..., N \}$}
\State Sample \boxed{c_{t+1}^{(i)} \sim \text{Bernoulli}\big(\gamma (\textbf{S}_{\mathrm{BF}}(y_{t+1}; \approxs{\pi}_i^{(t)}) , m )\big)}
  \EndFor
  \MyStatex{Resampling}
  \State $N_{\text{eff}} \gets (\sum_{i=1}^{N} w_{t+1}^{(i)^2})^{-1}$
  \State If $N_{\text{eff}} \leq N_{\text{thrs}}$: resample
  \MyStatex{Updated belief per particle $i$}
  \For {$i \in \{1, ..., N \}$}
  \If {$c_{t+1}^{(i)} = 0$}
  \State $\chi_i^{(t+1)} \gets \chi_i^{(t)} + \phi(y_{t+1})$ and $\nu_i^{(t+1)} \gets \nu_i^{(t)} + 1$
  \Else
  \State $\chi_i^{(t+1)} \gets \chi^{(0)} + \phi(y_{t+1})$ and $\nu_i^{(t+1)} \gets \nu^{(0)}+ 1$
  \EndIf
  \EndFor
  \MyStatex{Updated (output) belief}
  \State $\approxs{\pi}^{(t+1)}(\theta) = \sum_{i=1}^{N} w_{t+1}^{(i)} \textbf{P}_{\pi}\big(\Theta = \theta; \chi_i^{(t+1)}, \nu_i^{(t+1)}\big)$
  \MyStatex{Iterate}
  \State $t \gets t+1$
  \EndWhile
\end{algorithmic}
\end{algorithm}

\subsubsection{Surprise-modulation as a framework for other algorithms}

Other existing algorithms \citep{adams2007bayesian, fearnhead2007line, nassar2010approximately, nassar2012rational, faraji2018balancing} can also be formulated in the surprise-modulation framework of \autoref{Eq:Gamma_def} and \autoref{Eq:tradeoff} (see Methods).
Moreover, in order to allow for a transparent discussion and for fair comparisons in simulations, we extended the algorithms of \cite{nassar2010approximately, nassar2012rational} to a more general setting.
Here we give a brief summary of the algorithms we considered.
A detailed analysis is provided in subsection ``Surprise-modulation as a framework for other algorithms'' in the Methods.

The algorithms of \cite{nassar2010approximately, nassar2012rational} were originally designed for a Gaussian estimation task (see Simulations for details of the task) with a broad uniform prior.
We extended them to the more general case of Gaussian tasks with Gaussian priors, and we call our extended versions Nas10$^{*}$ and Nas12$^{*}$ for \cite{nassar2010approximately} and \cite{nassar2012rational} respectively (for a performance comparison between our extended algorithms and their original versions see Supplementary \autoref{fig:gauss_afterswitch_nassar} and Supplementary \autoref{fig:gauss_heatmaps_nassar}).
Both algorithms have the same surprise-modulation as in our Proposition (\autoref{Eq:Bayesian_Rec_Formula}).
There are multiple interpretations of the approaches of Nas10$^{*}$ and Nas12$^{*}$ and links to other algorithms.
One such link we identify is in relation to Particle Filtering with a single particle (pf$1$).
More specifically, one can show that pf$1$ behaves in expectation similar to Nas10$^{*}$ and Nas12$^{*}$ (see Methods and Supplementary Material).

To summarize, the algorithms Exact Bayes and SOR$N$ come from the field of change point detection, and whereas the former has high memory demands, the latter has the same memory demands as our algorithms pf$N$ and MP$N$.
The algorithms Nas10$^{*}$, Nas12$^{*}$, and SMiLe, on the other hand, come from the human learning literature and are more biologically oriented.

\subsection{Simulations}
\label{simulations}
With the goal of gaining a better understanding of different approximate algorithms, we evaluated the departure of their performance from the exact Bayesian algorithm in terms of mean squared error (MSE) of $\Theta_t$ (see Methods), on two tasks inspired by and closely related to real experiments \citep{nassar2010approximately,nassar2012rational, behrens2007learning, mars2008trial, ostwald2012evidence, maheu2019brain}: a Gaussian and a Categorical estimation task.

We compared our three novel algorithms VarSMiLe, Particle Filtering (pf$N$, where $N$ is the number of particles), and Message Passing with finite number of particles $N$ (MP$N$) to the online exact Bayesian Message Passing algorithm \citep{adams2007bayesian} (Exact Bayes), which yields the optimal solution with $\approxs{\Theta}_t = \approxs{\Theta}_t^{\text{Opt}}$.
Furthermore, we included in the comparison the stratified optimal resampling algorithm \citep{fearnhead2007line} (SOR$N$, where $N$ is the number of particles),
our variant of \cite{nassar2010approximately} (Nas10$^{*}$) and of \cite{nassar2012rational} (Nas12$^{*}$), the Surprise-Minimization Learning algorithm of \cite{faraji2018balancing} (SMiLe), as well as a simple Leaky Integrator (Leaky - see Methods).

\subsubsection{Gaussian estimation task}
\begin{figure}
    \centering
    \includegraphics[width=\textwidth]{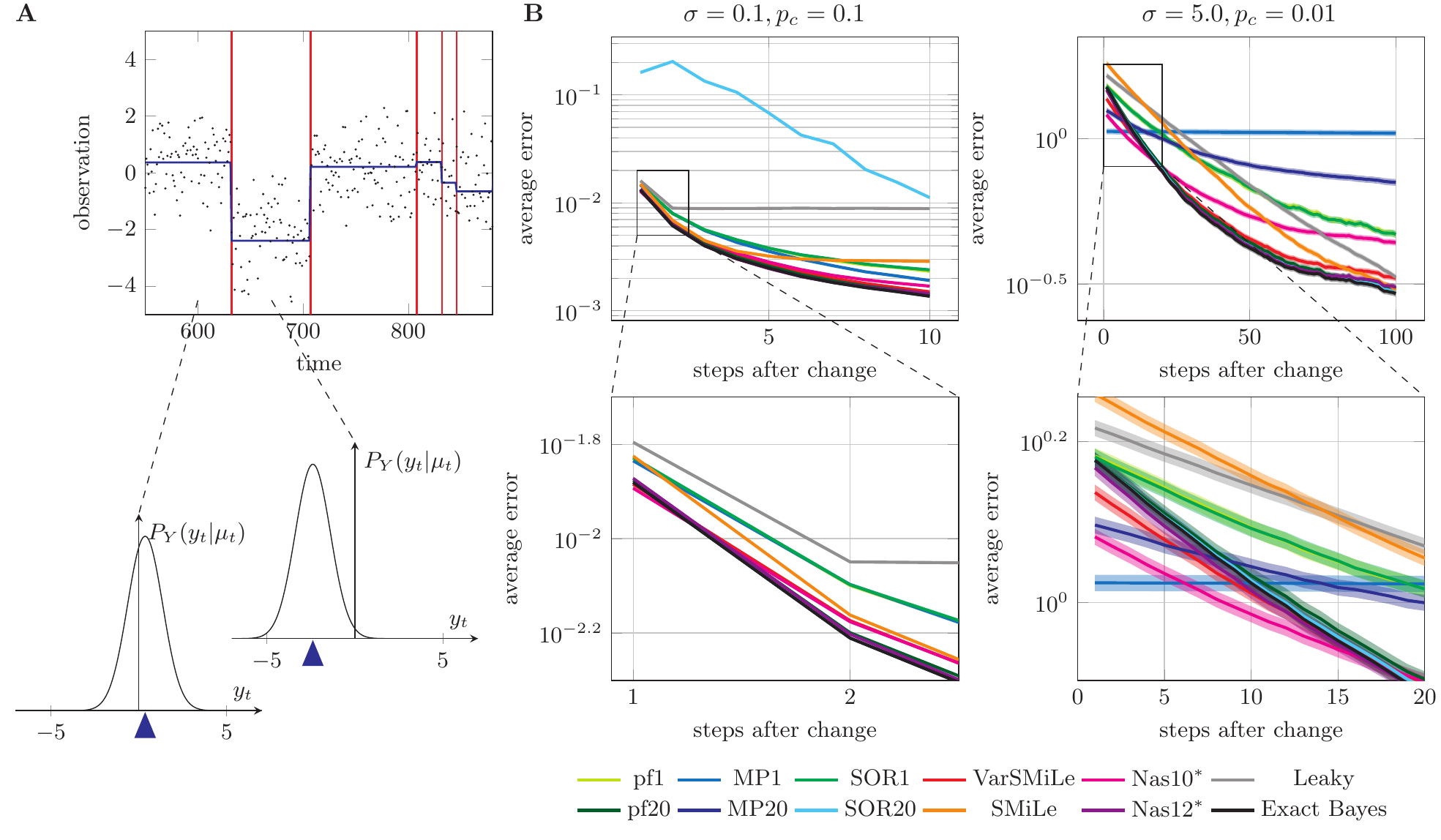}
    \caption{\textbf{Gaussian estimation task: Transient performance after changes.}
    \textbf{A}. At each time step an observation (depicted as black dot) is drawn from a Gaussian distribution $\sim \text{exp}(- (y_t - \mu_t)^2 / 2 \sigma^2)$ with changing mean $\mu_t$ (marked in blue) and known variance $\sigma^2$ (lower left panels).
    At every change of the environment (marked with red lines) a new mean $\mu_t$ is drawn from a standard Gaussian distribution $\sim \text{exp}(-\mu_t^2)$.
    In this example: $\sigma = 1$ and $p_c = 0.01$.
    \textbf{B}. Mean squared error for the estimation of $\mu_t$ at each time step $n$ after an environmental change, i.e.
    the average of $\textbf{MSE} [ \approxs{\Theta}_t | R_t = n]$ over time; $\sigma=0.1,\:p_c=0.1$ (left panel) and $\sigma=5,\:p_c=0.01$ (right panel).
    The shaded area corresponds to the standard error of the mean.
    \textit{Abbreviations:} pf$N$: Particle Filtering with $N$ particles,
    MP$N$: Message Passing with $N$ particles,
    VarSMiLe: Variational SMiLe,
    SOR$N$: Stratisfied Optimal Resampling with $N$ particles \citep{fearnhead2007line},
    SMiLe: \cite{faraji2018balancing},
    Nas10$^{*}$, Nas12$^{*}$: Variants of \cite{nassar2010approximately} and \cite{nassar2012rational}, respectively,
    Leaky: Leaky Integrator,
    Exact Bayes: \cite{adams2007bayesian}.
}\label{fig:gauss_afterswitch}
\end{figure}

The task is a generalized version of the experiment of \cite{nassar2010approximately,nassar2012rational}.
The goal of the agent is to estimate the mean $\theta_{t} = \mu_{t}$ of observed samples, which are drawn from a Gaussian distribution with known variance $\sigma^2$, i.e. $y_{t+1}|\mu_{t+1} \sim \mathcal{N}(\mu_{t+1}, \sigma^2)$.
The mean $\mu_{t+1}$ is itself drawn from a Gaussian distribution $\mu_{t+1} \sim \mathcal{N}(0, 1)$ whenever the environment changes.
In other words, the task is a special case of the generative model of \autoref{Eq:GenModel}, \autoref{Eq:GenModel2}, and \autoref{Eq:GenModel3}, with $\pi^{(0)}(\mu_t) = \mathcal{N}(\mu_t ; 0, 1)$ and
$P_{Y}(y_t | \mu_t) = \mathcal{N}(y_t ; \mu_{t}, \sigma^2)$.
An example of the task can be seen in \autoref{fig:gauss_afterswitch}A.

We simulated the task for all combinations of $\sigma \in \{ 0.1, 0.5, 1, 2, 5 \}$ and $p_c \in \{ 0.1, 0.05, 0.01, \allowbreak 0.005, \allowbreak 0.001, 0.0001 \}$.
For each combination of $\sigma$ and $p_c$, we first tuned the free parameter of each algorithm, i.e. $m$ for SMiLe and Variational SMiLe, the leak parameter
for the Leaky Integrator, and the $p_c$ of Nas10$^{*}$ and Nas12$^{*}$, by minimizing the MSE on three random initializations of the task.
For the Particle Filter (pf$N$), the Exact Bayes, the MP$N$, and the SOR$N$ we empirically checked that the true $p_c$ of the environment was indeed the value that gave the best performance, and we used this value for the simulations.
We evaluated the performance of the algorithms on ten different random task instances of $10^5$ steps each for $p_c \in \{ 0.1, 0.05, 0.01, 0.005 \}$ and $10^6$ steps each for $p_c \in \{ 0.001, 0.0001 \}$ (in order to sample more change points).
Note that the parameter $\sigma$ is not estimated and its actual value is used by all algorithms except the Leaky Integrator.

In \autoref{fig:gauss_afterswitch}B we show
the $\textbf{MSE} [ \approxs{\Theta}_t | R_t = n]$ in estimating the parameter after $n$ steps since the last change point, for each algorithm, computed over multiple changes, for two exemplar task settings.
The Particle Filter with 20 particles (pf20), the VarSMiLe and the Nas12$^{*}$ have an overall performance very close to that of the Exact Bayes algorithm (i.e.  $\textbf{MSE} [ \approxs{\Theta}_t^{\text{Opt}} | R_t = n ]$), with much lower memory requirements.
VarSMiLe sometimes slightly outperforms the other two early after an environmental change (\autoref{fig:gauss_afterswitch}B, right), but shows slightly higher error values at later phases.
The MP$N$ algorithm is the closest one to the optimal solution (i.e. Exact Bayes) for low $\sigma$ (\autoref{fig:gauss_afterswitch}B, left), but its performance is much worse for the case of high $\sigma$ and low $p_c$
(\autoref{fig:gauss_afterswitch}B, right).
For the Stratisfied Optimal Resampling (SOR$N$) we observe a counter-intuitive behavior in the regime of low $\sigma$; the inclusion of more particles leads to worse performance (\autoref{fig:gauss_afterswitch}B, left).
At higher $\sigma$ levels the performance of SOR20 is close to optimal and better than the MP20 in later time-steps.
This may be due to the fact that the MP$N$ discards particles in a deterministic and greedy way (i.e. the one with the lowest weight), whereas for the SOR$N$ there is a component of randomness in the process of particle elimination, which may be important for environments with higher stochasticity.

For the Leaky Integrator we observe a trade-off between good performance in the transient phase and the stationary phase; a fixed
leak value cannot fulfill both requirements.
The SMiLe rule, by construction, never narrows its belief $\approxs{\pi}(\theta)$ below some minimal value, which allows it to have a low error immediately after a change, but leads later to high errors.
Its performance deteriorates for higher $\sigma$ (\autoref{fig:gauss_afterswitch}B, right).
The Nas10$^{*}$ performs well for low, but not for higher values of $\sigma$.
Despite the fact that a Particle Filter with 1 particle (pf1) is in expectation similar to Nas10$^{*}$ and Nas12$^{*}$ (see Methods),
it performs worse than these two algorithms on trial-by-trial measures.
Still, it performs better than the MP1 and identically to the SOR1.

\begin{figure}
    \centering
\includegraphics[width=\textwidth]{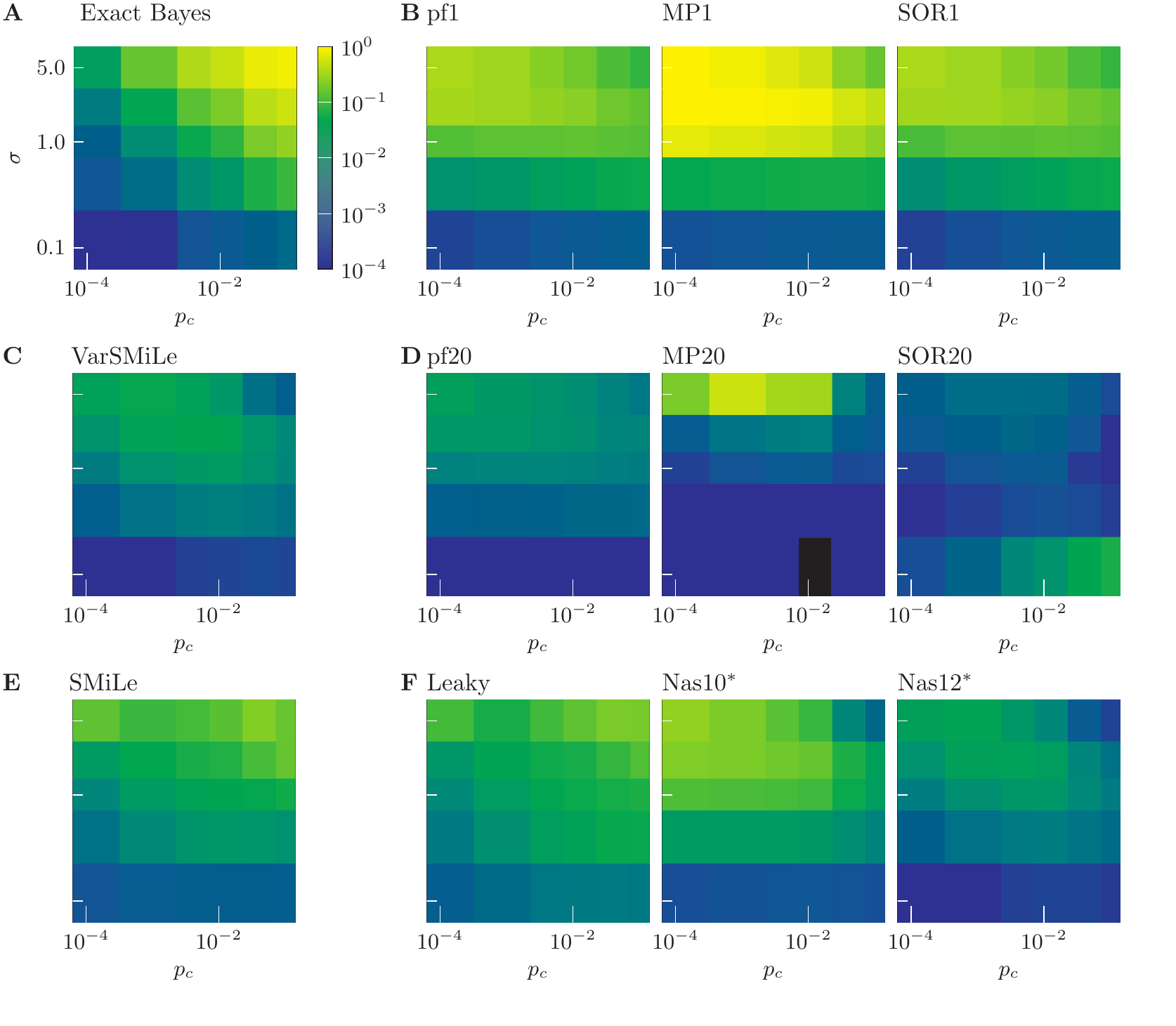}
    \caption{\textbf{Gaussian estimation task: Steady-state performance.}
    \textbf{A}. Mean squared error of the Exact Bayes algorithm (i.e. optimal solution) for each combination of $\sigma$ and $p_c$ averaged over time.
    \textbf{B -- F}. Difference between the mean squared error of each algorithm and the optimal solution (of panel A), i.e. the average of $\Delta \textbf{MSE} [ \approxs{\Theta}_t ]$ over time.
    The colorbar of panel A applies to these panels as well.
    Note that the black color for the MP20 indicates negative values, which are due to the finite sample size for the estimation of \textbf{MSE}.
    \textit{Abbreviations:} See the caption of \autoref{fig:gauss_afterswitch}.
    }\label{fig:gauss_heatmaps}
\end{figure}

In \autoref{fig:gauss_heatmaps}A, we have plotted the average of $\textbf{MSE} [ \approxs{\Theta}_t^{\text{Opt}}  ]$ of the Exact Bayes algorithm over the whole simulation time for each of the considered $\sigma$ and $p_c$ levels.
The difference between the other algorithms and this benchmark is called $\Delta \textbf{MSE} [ \approxs{\Theta}_t ]$ (see Methods) and is plotted in \autoref{fig:gauss_heatmaps}C--F.
All algorithms except for the SOR20 have lower average error values for low $\sigma$ and low $p_c$, than high $\sigma$ and high $p_c$.
The Particle Filter pf20 and the Message Passing MP20 have the smallest difference from the optimal solution.
The average error of MP20 is higher than that of pf20 for high $\sigma$ and low $p_c$, whereas pf20 is more robust across levels of environmental parameters.
The worst case performance for pf20 is $\Delta \textbf{MSE} [ \approxs{\Theta}_t ] = 0.033$ for $\sigma = 5$ and $p_c = 0.0001$, and for SOR20 it is $\Delta \textbf{MSE} [ \approxs{\Theta}_t ] = 0.061$ for $\sigma = 0.1$ and $p_c = 0.1$.
The difference between these two worst case scenarios is significant ($p\text{-value}= 2.79 \times 10^{-6}$, two-sample t-test, 10 random seeds for each algorithm).
Next in performance is the algorithm Nas12$^{*}$ and VarSMiLe.
VarSMiLe exhibits its largest deviation from the optimal solution for high $\sigma$ and low $p_c$, but is still more resilient compared to the MP$N$ algorithms for this type of environments.
Among the algorithms with only one unit of memory demands, i.e. pf1, MP1, SOR1, VarSMiLe, SMiLe, Leaky, Nas10$^{*}$ and Nas12$^{*}$, the winners are VarSMiLe and Nas12$^{*}$.
The SOR20 has low error overall, but unexpectedly high error for environmental settings that are presumably more relevant for biological agents (intervals of low stochasticity marked by abrupt changes).
The simple Leaky Integrator performs well at low $\sigma$ and $p_c$ but deviates more from the optimal solution as these parameters increase (\autoref{fig:gauss_heatmaps}F).
The SMiLe rule performs best at lower $\sigma$, i.e. in more deterministic environments.

\begin{figure}
    \centering
\makebox[\textwidth][c]{\includegraphics{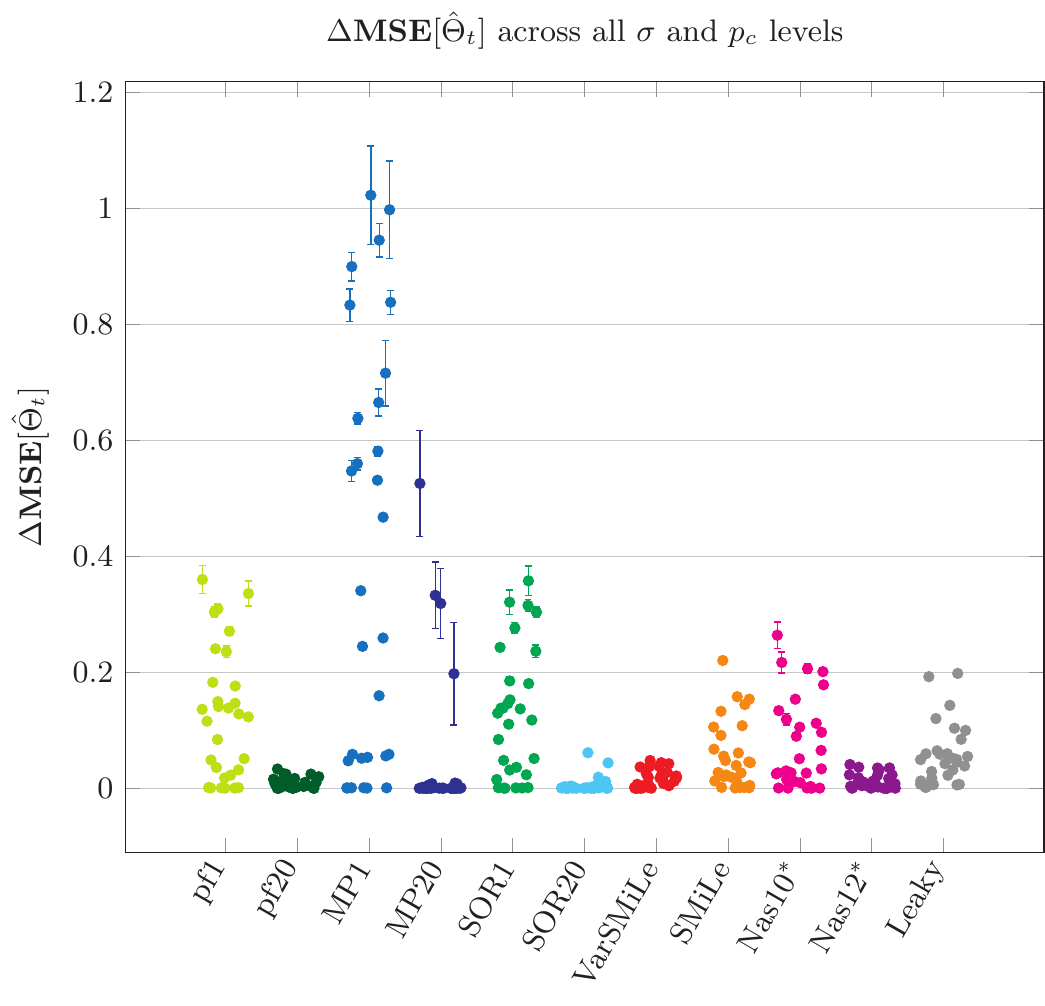}}
    \caption{\textbf{Gaussian estimation task: Steady-state performance summary.}
    Difference between the mean squared error of each algorithm and the optimal solution (Exact Bayes), i.e. the average of $\Delta \textbf{MSE} [ \approxs{\Theta}_t ]$ over time, for all combinations of $\sigma$ and $p_c$ together.
    For each algorithm we plot the 30 values (5 $\sigma$ times 6 $p_c$ values) of \autoref{fig:gauss_heatmaps} with respect to randomly jittered values in the $x$-axis.
    The color coding is the same as in \autoref{fig:gauss_afterswitch}.
    The errorbars mark the standard error of the mean across 10 random task instances.
    The difference between the worst case of SOR20 and pf20 is significant ($p\text{-value}= 2.79 \times 10^{-6}$, two-sample t-test, 10 random seeds for each algorithm).
    \textit{Abbreviations:} See the caption of \autoref{fig:gauss_afterswitch}.
}\label{fig:gauss_boxplot}
\end{figure}

A summary graph, where we collect the $\Delta \textbf{MSE} [ \approxs{\Theta}_t ]$ across all levels of $\sigma$ and $p_c$, is shown in \autoref{fig:gauss_boxplot}.
We can see that pf20, Nas12$^{*}$, and VarSMiLe give the lowest worst case (lowest maximum value) $\Delta \textbf{MSE} [ \approxs{\Theta}_t ]$ and are statistically better than the other 8 algorithms (the errorbars indicate the standard error of the mean across the ten random task instances).

\subsubsection{Categorical estimation task}
\label{sub:cat}
\begin{figure}
    \centering
    \includegraphics[width=\textwidth]{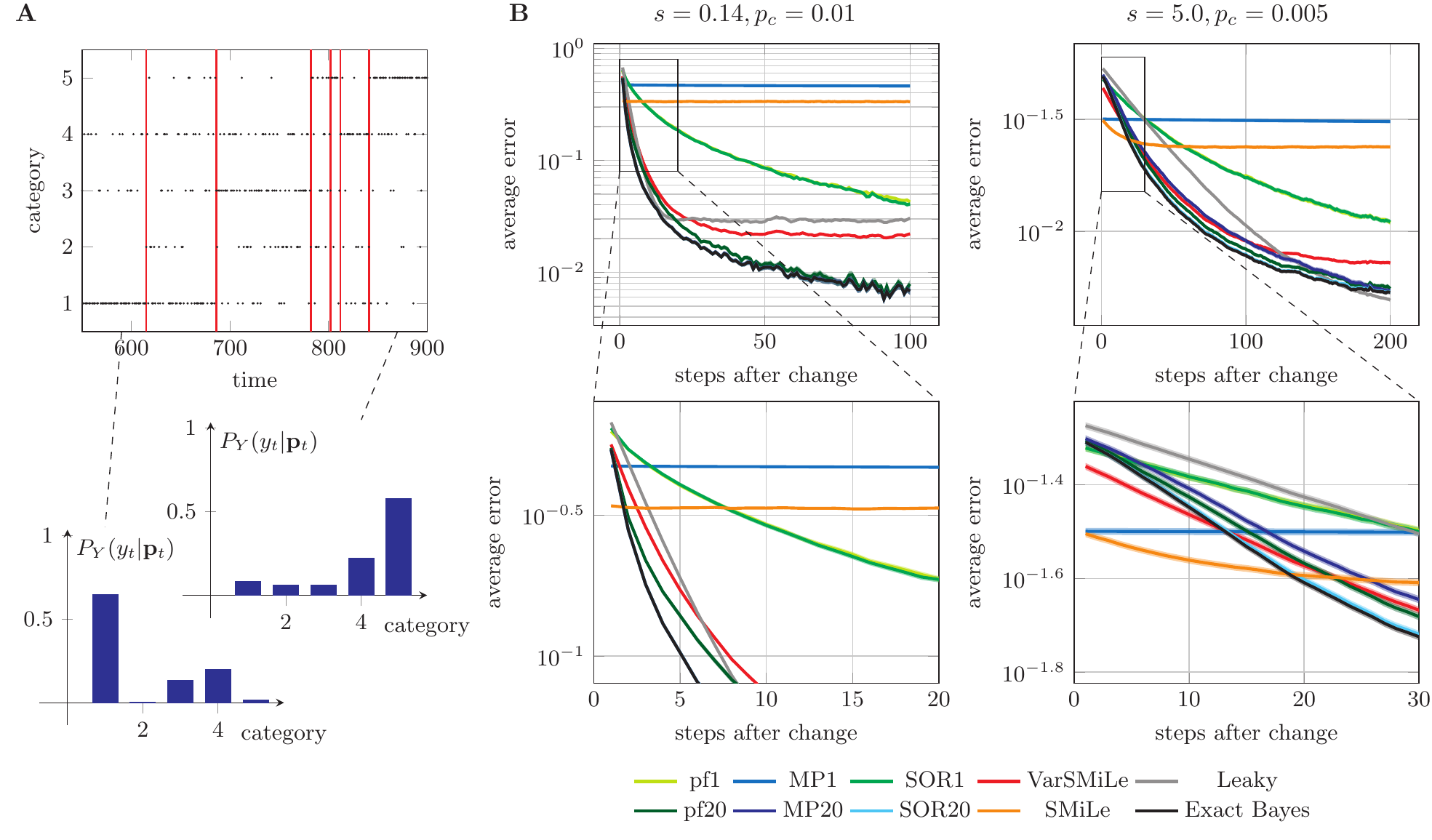}
    \caption{\textbf{Categorical estimation task: Transient performance after changes.}
    \textbf{A}. At each time step the agent sees one out of 5 possible categories (black dots) drawn from a categorical distribution with parameters $\boldsymbol{p_t}$.
    Occasional abrupt changes happen with probability $p_c$ and are marked with red lines.
    After each change a new $\boldsymbol{p_t}$ vector is drawn from a Dirichlet distribution with stochasticity parameter $s$.
    In this example: $s = 1$ and $p_c = 0.01$.
    \textbf{B}. Mean squared error for the estimation of $\boldsymbol{p_t}$ at each time step $n$ after an environmental change, i.e.
    the average of $\textbf{MSE} [ \approxs{\Theta}_t | R_t = n]$ over time; $s=0.14,\:p_c=0.01$ (left panel) and $s=5,\:p_c=0.005$ (right panel).
    The shaded area corresponds to the standard error of the mean.
    \textit{Abbreviations:} pf$N$: Particle Filtering with $N$ particles,
    MP$N$: Message Passing with $N$ particles,
    VarSMiLe: Variational SMiLe,
    SOR$N$: Stratisfied Optimal Resampling with $N$ particles \citep{fearnhead2007line},
    SMiLe: \cite{faraji2018balancing},
    Leaky: Leaky Integrator,
    Exact Bayes: \cite{adams2007bayesian}.
    }\label{fig:multi_afterswitch}
\end{figure}

The task is inspired by the experiments of \cite{behrens2007learning, mars2008trial, ostwald2012evidence, maheu2019brain}.
The goal of the agent is to estimate the occurrence probability of five possible states. Each observation $y_{t+1} \in \{ 1, ..., 5 \}$  is drawn from a categorical distribution with parameters $\theta_{t+1} = \boldsymbol{p}_{t+1}$, i.e. $y_{t+1}|\boldsymbol{p}_{t+1} \sim \text{Cat}(y_{t+1} ; \boldsymbol{p}_{t+1})$.
When there is a change $C_{t+1} = 1$ in the environment, the parameters $\boldsymbol{p}_{t+1}$ are drawn from a Dirichlet distribution $\text{Dir}(s \cdot \boldsymbol{1})$, where $s \in (0, \infty)$ is the stochasticity parameter.
In relation to the generative model of \autoref{Eq:GenModel}, \autoref{Eq:GenModel2}, and \autoref{Eq:GenModel3} we, thus, have $\pi^{(0)}(\boldsymbol{p}_{t}) = \text{Dir}(\boldsymbol{p}_{t} ; s \cdot \boldsymbol{1})$ and
$P_{Y}(y_t | \boldsymbol{p}_{t}) = \text{Cat}(y_{t} ; \boldsymbol{p}_{t})$.
An illustration of this task is depicted in \autoref{fig:multi_afterswitch}A.

We considered the combinations of stochasticity levels $s \in \{ 0.01, 0.1, 0.14, 0.25, \allowbreak 1, \allowbreak 2, 5 \}$ and change probability levels $p_c \in \{ 0.1, 0.05, 0.01, 0.005, 0.001, 0.0001 \}$.
The algorithms of \cite{nassar2012rational, nassar2010approximately} were specifically developed for a Gaussian estimation task and cannot be applied here.
All other algorithms were first optimized for each combination of environmental parameters before an experiment starts, and then evaluated on ten different random task instances, for $10^5$ steps each for $p_c \in \{ 0.1, 0.05, 0.01, 0.005 \}$ and $10^6$ steps each for $p_c \in \{ 0.001, 0.0001 \}$.
The parameter $s$ is not estimated and its actual value is used by all algorithms except the Leaky Integrator.

\begin{figure}
    \centering
    \includegraphics[width=\textwidth]{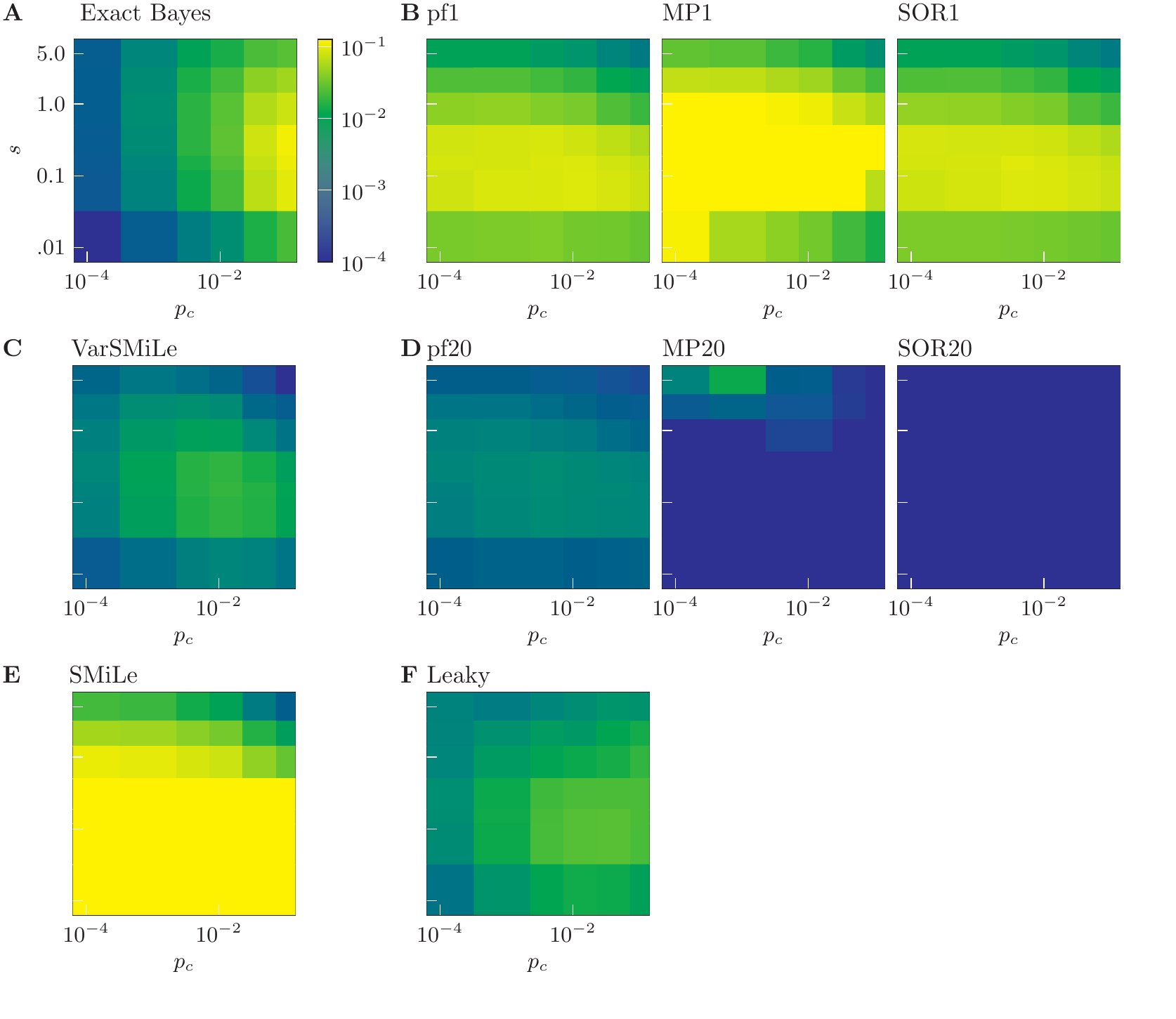}
    \caption{\textbf{Categorical estimation task: Steady-state performance.}
    \textbf{A}. Mean squared error of the Exact Bayes algorithm (i.e. optimal solution) for each combination of environmental parameters $s$ and $p_c$ averaged over time.
    \textbf{B -- F}. Difference between the mean squared error of each algorithm and the optimal solution (of panel A), i.e. the average of $\Delta \textbf{MSE} [ \approxs{\Theta}_t ]$ over time.
    The colorbar of panel A applies to these panels as well.
    \textit{Abbreviations:} See the caption of \autoref{fig:multi_afterswitch}.
    }\label{fig:multi_heatmaps}
\end{figure}

The Particle Filter pf20, the MP20 and the SOR20 have a performance closest to that of Exact Bayes, i.e. the optimal solution (\autoref{fig:multi_afterswitch}B).
VarSMiLe is the next in the ranking, with a behavior after a change similar to the Gaussian task.
pf20 performs better for $s>2$ and MP20 performs better for $s \leq 2$ (\autoref{fig:multi_heatmaps}).
For this task the biologically less plausible SOR20 is the winner in performance and it behaves most consistently across environmental parameters.
Its worst case performance is $\Delta \textbf{MSE} [ \approxs{\Theta}_t ] = 8.16 \times 10^{-5}$ for $s = 2$ and $p_c = 0.01$, and the worst case performance for pf20 is $\Delta \textbf{MSE} [ \approxs{\Theta}_t ] = 0.0048$ for $s = 0.25$ and $p_c = 0.005$ ($p\text{-value}= 1.148 \times 10^{-12}$, two-sample t-test, 10 random seeds for each algorithm).
For all the other algorithms, except for MP20, the highest deviations from the optimal solution are observed for medium stochasticity levels (\autoref{fig:multi_heatmaps}B--F).
When the environment is nearly deterministic (e.g. $s = 0.001$ so that the parameter vectors $\boldsymbol{p}_{t}$ have almost all mass concentrated in one component), or highly stochastic (e.g. $s > 1$ so that nearly uniform categorical distributions are likely to be sampled), these algorithms achieve higher performance, while the Particle Filter is the algorithm that is most resilient to extreme choices of the stochasticity parameter $s$.
For VarSMiLe in particular, the lowest mean error is achieved for high $s$ and high $p_c$ or low $s$ and low $p_c$.

\begin{figure}
    \centering
\makebox[\textwidth][c]{\includegraphics{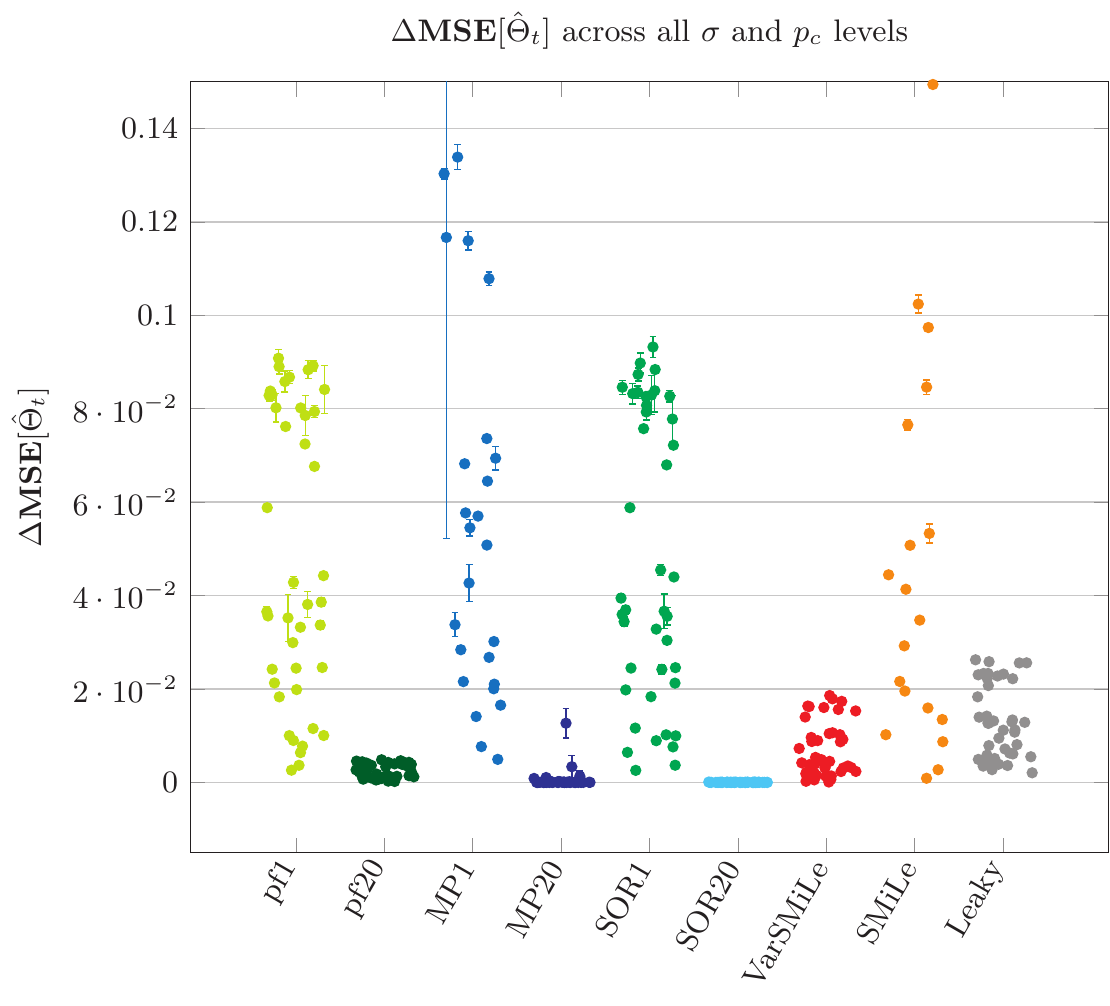}}
    \caption{\textbf{Categorical estimation task: Steady-state performance summary.}
    Difference between the mean squared error of each algorithm and the optimal solution (Exact Bayes), i.e. the average of $\Delta \textbf{MSE} [ \approxs{\Theta}_t ]$ over time, for all combinations of $s$ and $p_c$ together.
    For each algorithm we plot the 42 values (7 $s$ times 6 $p_c$ values) of \autoref{fig:multi_heatmaps} with respect to randomly jittered values in the $x$-axis.
    The color coding is the same as in \autoref{fig:multi_afterswitch}.
    The errorbars mark the standard error of the mean across 10 random task instances.
    The difference between the worst case of SOR20 and pf20 is significant ($p\text{-value}= 1.148 \times 10^{-12}$, two-sample t-test, 10 random seeds for each algorithm).
    \textit{Abbreviations:} See the caption of \autoref{fig:multi_afterswitch}.
Note that MP1 and SMiLe are out of bound with a maximum at 0.53.
    }\label{fig:multi_boxplot}
\end{figure}

A summary graph, with the $\Delta \textbf{MSE} [ \approxs{\Theta}_t ]$ across all levels of $s$ and $p_c$, can be seen in \autoref{fig:multi_boxplot}.
The algorithms with the lowest ``worst case'' are SOR20 and pf20.
The top-4 algorithms SOR20, pf20, MP20 and VarSMiLe are significantly better than the others (the errorbars indicate the standard error of the mean across the ten random task instances), whereas MP$1$ and SMiLe have the largest error with a maximum at 0.53.

\subsubsection{Summary of simulation results}

In summary, our simulation results of the two tasks collectively suggest that our Particle Filtering (pf$N$) and Message Passing (MP$N$) algorithms achieve a high level of performance, very close to the one of biologically less plausible algorithms with higher (Exact Bayes) and same (SOR$N$) memory demands.
Moreover, their behavior is more consistent across tasks.
Finally, among the algorithms with memory demands of one unit, VarSMiLe performs best.

\subsubsection{Robustness against suboptimal parameter choice}

In all algorithms we considered, the environment's hyper-parameters are assumed to be known.
We can distinguish between two types of hyper-parameters in our generative model:
1. the parameters of the likelihood function (e.g. $\sigma$ in the Gaussian task), and
2. the $p_c$ and the parameters of the conjugate prior (e.g. $s$ in the Categorical task).
Hyper-parameters of the first type can be added to the parameter vector $\theta$ and be inferred with the same algorithm.
However, learning the second type of hyper-parameters is not straightforward.
By assuming that these hyper-parameters are learned more slowly than $\theta$, one can fine-tune them after each $n$ (e.g. 10) change points, while change points can be detected by looking at the particles for the Particle Filter and at the peaks of surprise values for VarSMiLe.
Other approaches to hyper-parameter estimation can be found in \cite{george2017principled, liu2001combined, doucet2003parameter, wilson2010bayesian}.

When the hyper-parameters are fixed, a mismatch between the assumed values and the true values is a possible source of errors.
In this section, we investigate the robustness of the algorithms to a mismatch between the assumed and the actual probability of change points.
To do so, we first tuned each algorithm's parameter for an environment with a change probability $p_c$, and then tested the algorithms in environments with different change probabilities, while keeping the parameter fixed.
For each new environment with a different change probability, we calculated the difference between the MSE of these fixed parameters and the optimal MSE, i.e. the resulting MSE for the case that the Exact Bayes' parameter is tuned for the actual $p_c$.

More precisely, if we denote as $\textbf{MSE} [ \approxs{\Theta}_t ; p_c' , p_c ]$ the MSE of an algorithm with parameters tuned for an environment with $p_c'$, applied in an environment with $p_c$, we calculated the mean regret, defined as $\textbf{MSE} [ \approxs{\Theta}_t ; p_c' , p_c ] - \textbf{MSE} [ \approxs{\Theta}_t^{\text{Opt}} , p_c ]$ over time;
note that the second term is equal to $\textbf{MSE} [ \approxs{\Theta}_t ; p_c , p_c ]$ when the algorithm Exact Bayes is used for estimation.
The lower the values and the flatter the curve of the mean regret, the better the performance and the robustness of the algorithm in the face of lacking knowledge of the environment.
The slope of the curve indicates the degree of deviations of the performance as we move away from the optimally tuned parameter.
We ran three random (and same for all algorithms) tasks initializations for each $p_c$ level.

\begin{figure}
    \centering
\makebox[\textwidth][c]{\includegraphics{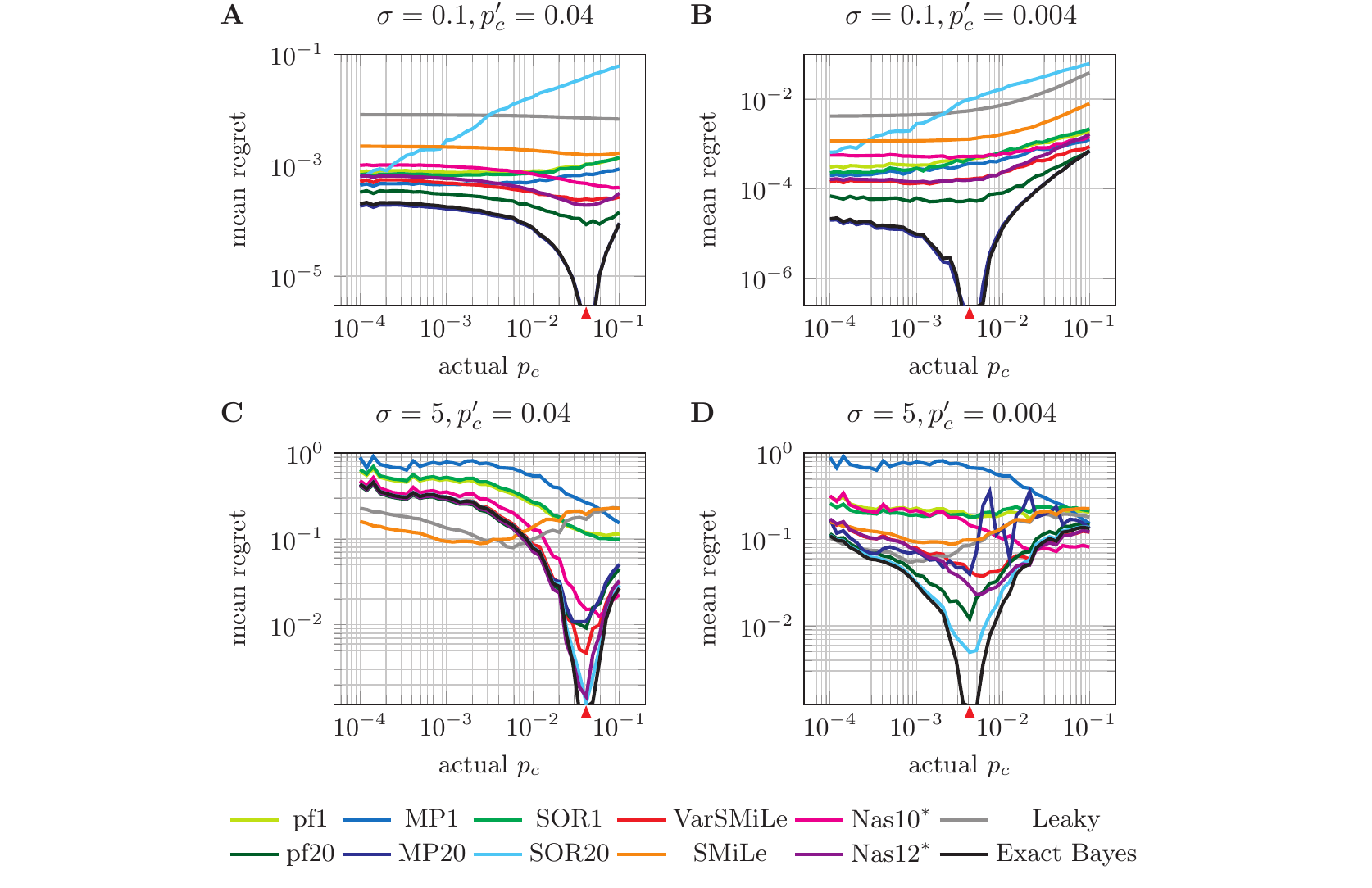}}
    \caption{\textbf{Robustness to mismatch between actual and assumed probability of changes for the Gaussian estimation task.}
    The mean regret is the mean squared error obtained with assumed change probability $p_c'$ minus the mean squared error obtained with the optimal parameter choice of Exact Bayes for the given actual $p_c$, i.e. the average of the quantity $\textbf{MSE} [ \approxs{\Theta}_t ; p_c' , p_c ] - \textbf{MSE} [ \approxs{\Theta}_t^{\text{Opt}} , p_c ]$ over time versus.
    A red triangle marks the $p_c'$ value each algorithm was tuned for.
    We plot the mean regret for the following parameter combinations:
    \textbf{A}. $\sigma=0.1$ and $p_c'=0.04$,
    \textbf{B}. $\sigma=0.1$ and $p_c'=0.004$,
    \textbf{C}. $\sigma=5$ and $p_c'=0.04$,
    \textbf{D}. $\sigma=5$ and $p_c'=0.004$.
    \textit{Abbreviations:} See the caption of \autoref{fig:gauss_afterswitch}.
    }\label{fig:robustnessgaussian}
\end{figure}

In \autoref{fig:robustnessgaussian} we plot the mean regret for each algorithm for the Gaussian task for four pairs of $s$ and $p_c'$ levels.
For $\sigma=0.1$ and $p_c' = 0.04$ (\autoref{fig:robustnessgaussian}A) the Exact Bayes and the MP20 show the highest robustness (smallest regret) and are closely followed by the pf20, VarSMiLe, and Nas12$^{*}$ (note the regret's small range of values).
The lower the actual $p_c$, the higher the regret, but still the changes are very small.
The curves for the SMiLe and the Leaky Integrator are also relatively flat, but the mean regret is much higher.
The SOR$20$ is the least robust algorithm.

Similar observations can be made for $\sigma=0.1$ and $p_c' = 0.004$ (\autoref{fig:robustnessgaussian}B).
In this case, the performance of all algorithms deteriorates strongly when the actual $p_c$ is higher than the assumed one.

However, for $\sigma=5$ (\autoref{fig:robustnessgaussian}C and \autoref{fig:robustnessgaussian}D), the ranking of algorithms changes.
The SOR$20$ is very robust for this level of stochasticity.
The pf$20$ and MP$20$  perform similarly for $p_c = 0.04$, but for lower $p_c'$ the pf$20$ is more robust and the MP$20$ exhibits high fluctuations in its performance.
The Nas12$^{*}$ is quite robust at this $\sigma$ level.
Overall for Exact Bayes, SOR$20$, pf$20$, VarSMiLe and Nas12$^{*}$, a mismatch of the assumed $p_c$ from the actual one does not deteriorate the performance dramatically for $\sigma=5$, $p_c' = 0.004$ (\autoref{fig:robustnessgaussian}D).
The SMiLe and the Leaky Integrator outperform the other algorithms for higher $p_c'$ if $p_c < p_c'$ (\autoref{fig:robustnessgaussian}C).
A potential reason is that the optimal behavior for the Leaky Integrator (according to the tuned parameters) is to constantly integrate new observations into its belief (i.e. to act like a Perfect Integrator) regardless of the $p_c'$ level.
This feature makes it blind to the $p_c$ and therefore very robust against the lack of knowledge of it (\autoref{fig:robustnessgaussian}C).

\begin{figure}
    \centering
\makebox[\textwidth][c]{\includegraphics{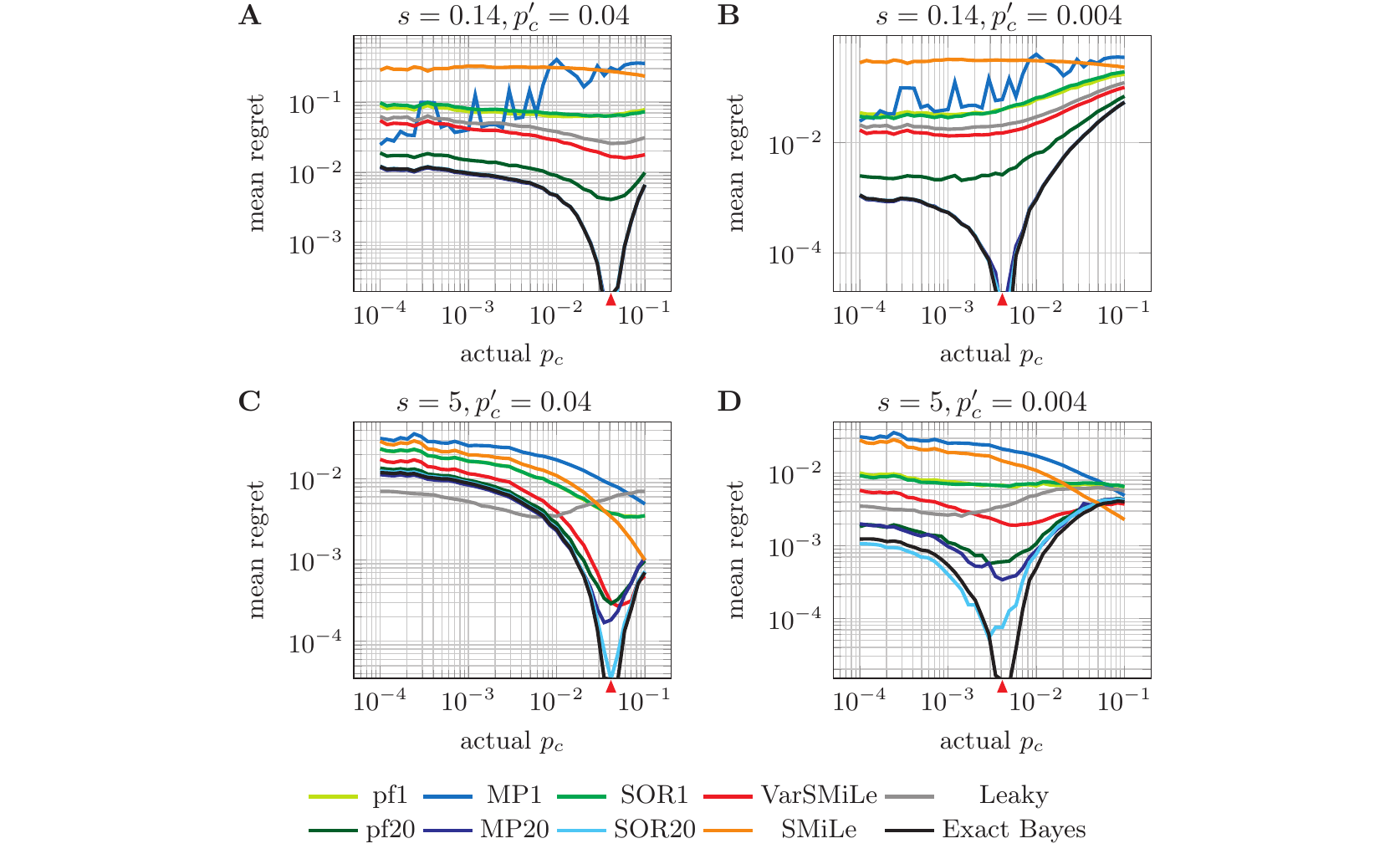}}
    \caption{\textbf{Robustness to mismatch between actual and assumed probability of changes for the Categorical estimation task.}
    The mean regret is the mean squared error obtained with assumed change probability $p_c'$ minus the mean squared error obtained with the optimal parameter choice of Exact Bayes for the given actual $p_c$, i.e. the average of the quantity $\textbf{MSE} [ \approxs{\Theta}_t ; p_c' , p_c ] - \textbf{MSE} [ \approxs{\Theta}_t^{\text{Opt}} , p_c ]$ over time.
    A red triangle marks the $p_c'$ value each algorithm was tuned for.
    We plot the mean regret for the following parameter combinations:
    \textbf{A}. $s=0.14$ and $p_c'=0.04$,
    \textbf{B}. $s=0.14$ and $p_c'=0.004$,
    \textbf{C}. $s=5$ and $p_c'=0.04$,
    \textbf{D}. $s=5$ and $p_c'=0.004$.
    \textit{Abbreviations:} See the caption of \autoref{fig:multi_afterswitch}
    }\label{fig:robustnessmulti}
\end{figure}

In summary, most of the time, the mean regret for Exact Bayes and MP20 is less than the mean regret for pf20 and VarSMiLe.
However, the variability in the mean regret for pf20 and VarSMiLe is smaller, and their curves are flatter across $p_c$ levels, which makes their performance more predictable.
The results for the Categorical estimation task are similar to those of the Gaussian task, with the difference that the SOR$20$ is very robust for this case (\autoref{fig:robustnessmulti}).


\subsection{Experimental prediction}

It has been experimentally shown that some important behavioral and physiological indicators statistically correlate with a measure of surprise or a prediction error.
Examples of such indicators are the pupil diameter \citep{preuschoff2011pupil, nassar2012rational, joshi2019pupil}, the amplitude of the P300, N400, and MMN components of EEG \citep{mars2008trial, ostwald2012evidence, lieder2013modelling, kopp2013electrophysiological, meyniel2016human, modirshanechi2019trial, musiolekmodeling},
the amplitude of MEG in specific time windows \citep{maheu2019brain}, BOLD responses in fMRI \citep{konovalov2018neurocomputational, loued2020anterior}, and reaction time \citep{huettel2002perceiving, meyniel2016human}.
The surprise measure is usually the negative log-probability of the observation, known as Shannon Surprise \citep{shannon1948mathematical}, and denoted here as $\textbf{S}_{\mathrm{Sh}}$.
However, as we show in this section, as long as there is an uninformative prior over observations, Shannon Surprise $\textbf{S}_{\mathrm{Sh}}$ is just an invertible function of our modulated adaptation rate $\gamma$ and hence an invertible function of the Bayes Factor Surprise $\textbf{S}_{\mathrm{BF}}$.
Thus, based on the results of previous works \citep{nassar2010approximately, nassar2012rational, meyniel2016human, ostwald2012evidence, modirshanechi2019trial}, that always used uninformative priors, one cannot determine whether the aforementioned physiological and behavioral indicators correlate with $\textbf{S}_{\mathrm{Sh}}$ or $\textbf{S}_{\mathrm{BF}}$.

In this section, we first investigate the theoretical differences between the Bayes Factor Surprise $\textbf{S}_{\mathrm{BF}}$ and Shannon Surprise $\textbf{S}_{\mathrm{Sh}}$.
Then, based on their observed differences, we formulate two experimentally testable predictions, with a detailed experimental protocol.
Our predictions make it possible to discriminate between the two measures of surprise, and to determine whether physiological or behavioral measurements are signatures of $\textbf{S}_{\mathrm{BF}}$ or of $\textbf{S}_{\mathrm{Sh}}$.

\subsubsection{Theoretical difference between $\textbf{S}_{\mathrm{BF}}$ and $\textbf{S}_{\mathrm{Sh}}$}

Shannon Surprise \citep{shannon1948mathematical} is defined as
\begin{equation}
\begin{aligned}
  \textbf{S}_{\mathrm{Sh}}(y_{t+1}; \pi^{(t)}) &= - \text{log} \Big( \textbf{P}(y_{t+1}|y_{1:t}) \Big)
\end{aligned}
\end{equation}
where for computing $\textbf{P}(y_{t+1}|y_{1:t})$, one should know the structure of the generative model.
For the generative model of \autoref{fig:gen_model}A, we find $\textbf{S}_{\mathrm{Sh}}(y_{t+1}; \pi^{(t)}) = - \text{log} \Big( (1-p_c) P(y_{t+1}; \pi^{(t)}) + p_c P(y_{t+1}; \pi^{(0)}) \Big)$.
While the Bayes Factor Surprise $\textbf{S}_{\mathrm{BF}}$ depends on a ratio between the probability of the new observation under the prior and the current beliefs, Shannon Surprise depends on a weighted sum of these probabilities.
Interestingly, it is possible to express (see Methods for derivation)
the adaptation rate $\gamma_{t+1}$ as a function of the ``difference in Shannon Surprise''
\begin{equation}
\begin{aligned}
  \label{gamma_Shannon}
  &\gamma_{t+1} = p_c \text{exp} \Big(  \Delta \textbf{S}_{\mathrm{Sh}}(y_{t+1}; \pi^{(t)}, \pi^{(0)})  \Big),\\
  & \text{    where } \Delta \textbf{S}_{\mathrm{Sh}}(y_{t+1}; \pi^{(t)}, \pi^{(0)}) = \textbf{S}_{\mathrm{Sh}}(y_{t+1}; \pi^{(t)}) - \textbf{S}_{\mathrm{Sh}}(y_{t+1}; \pi^{(0)}),
\end{aligned}
\end{equation}
where $\gamma_{t+1} = \gamma \big(\textbf{S}_{\mathrm{BF}}^{(t+1)}, m \big)$ depends on the Bayes Factor Surprise and the saturation parameter $m$ (cf. \autoref{Eq:Gamma_def}).
Equation \ref{gamma_Shannon} shows that the modulated adaptation rate is not just a function of Shannon Surprise upon observing $y_{t+1}$, but a function of the \textit{difference} between the Shannon Surprise of this observation under the current and under the prior beliefs.
In the next subsections, we exploit differences between $\textbf{S}_{\mathrm{BF}}$ and $\textbf{S}_{\mathrm{Sh}}$ to formulate our experimentally testable predictions.

\subsubsection{Experimental protocol}
Consider the variant of the Gaussian task of \cite{nassar2010approximately, nassar2012rational} which we used in our simulations, i.e. $P_Y(y|\theta) = \mathcal{N}(y; \theta, \sigma^2)$ and $\pi^{(0)}(\theta) =  \mathcal{N}(\theta; 0, 1)$.
Human subjects are asked to predict the next observation $y_{t+1}$ given what they have observed so far, i.e. $y_{1:t}$.
The experimental procedure is as follows:
\begin{enumerate}
  \item Fix the hyper parameters $\sigma^2$ and $p_c$.
  \item At each time $t$, show the observation $y_t$ (produced in the aforementioned way) to the subject, and measure a physiological or behavioral indicator $M_t$, e.g. pupil diameter \citep{nassar2010approximately, nassar2012rational}.
  \item At each time $t$, after observing $y_t$, ask the subject to predict the next observation $\approxs{y}_{t+1}$ and their confidence $C_t$ about their prediction.
\end{enumerate}
Note that the only difference between our task and the task of \cite{nassar2010approximately, nassar2012rational} is the choice of prior for $\theta$ (i.e. Gaussian instead of uniform).
The assumption is that, according to the previous studies, there is a \textit{positive} correlation between $M_t$ and a measure of surprise.

\subsubsection{Prediction 1}
\begin{figure}
    \centering
\includegraphics[width=\textwidth]{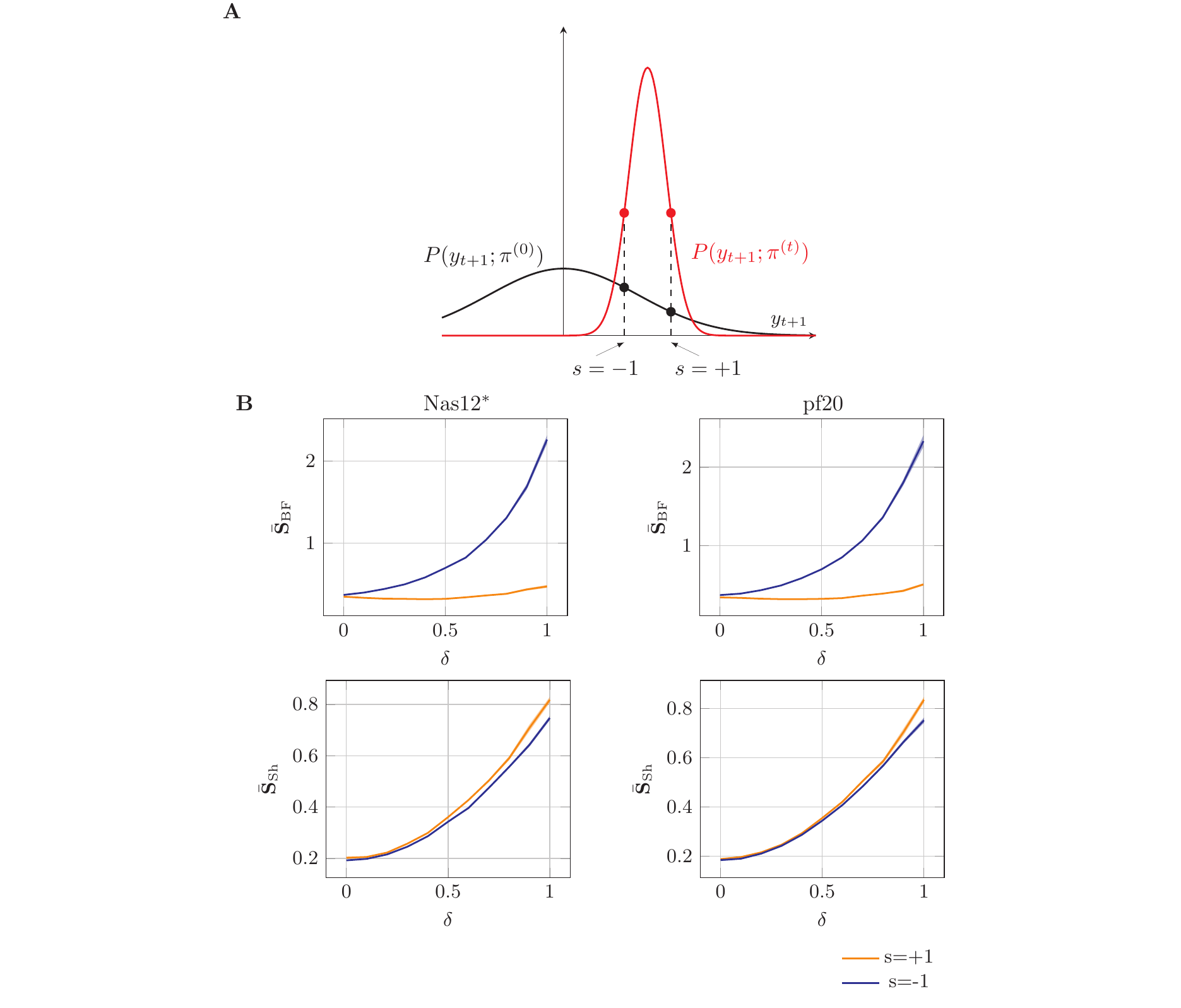}
    \caption{\textbf{Experimental prediction 1.}
    \textbf{A.}
    Schematic of the task for the case of a Gaussian belief.
    The distribution of $y_{t+1}$ under the prior belief $\pi^{(0)}$ and the current belief $\pi^{(t)}$ are shown by black and red curves, respectively.
    Two possible observations with equal absolute prediction error $\delta$ but opposite sign bias $s$ are indicated by dashed lines.
The two observations are equally probable under $\pi^{(t)}$, but not under $\pi^{(0)}$.
    $\textbf{S}_{\mathrm{BF}}$ is computed as the ratio between the red and black dots for a given observation, whereas $\textbf{S}_{\mathrm{Sh}}$ is a function of the weighted sum of the two.
    This phenomenon is the basis of our experimental prediction.
    \textbf{B.}
    The average surprise values $\Bar{\textbf{S}}_{\mathrm{Sh}}(\approxs{\theta}=1,\delta,s=\pm 1,\sigma_C=0.5)$ and $\Bar{\textbf{S}}_{\mathrm{BF}}(\approxs{\theta}=1,\delta,s=\pm 1,\sigma_C=0.5)$ over 20 subjects (each with 500 observations) are shown for two different learning algorithms (Nas12$^*$ and pf20).
    The mean $\Bar{\textbf{S}}_{\mathrm{BF}}$ is higher for negative sign bias (marked in blue) than for positive sign bias (marked in orange).
    The opposite is observed for the mean $\Bar{\textbf{S}}_{\mathrm{Sh}}$.
    This effect increases with increasing values of prediction error $\delta$.
    The shaded area corresponds to the standard error of the mean.
    The experimental task is the same as the Gaussian task we used in the previous section, with $\sigma = 0.5$ and $p_c = 0.1$ (see Methods for details).
    }\label{fig:experimentalpredictionfirst}
\end{figure}

Based on the results of \cite{nassar2010approximately, nassar2012rational}, in such a Gaussian task, the best fit for subjects' prediction $\approxs{y}_{t+1}$ is $\approxs{\theta}_t$, and the confidence $C_t$ is a monotonic function of $\approxs{\sigma}_t$.
In order to formalize our experimental prediction, we define, at time $t$, the prediction error as $\delta_t = y_{t} - \approxs{y}_{t}$ and the ``sign bias'' as $s_t = \text{sign}(\delta_t \approxs{y}_{t})$.
The variable $s_t$ is a crucial variable for our analysis.
It shows whether the prediction $\approxs{y}_{t}$ is an overestimation in absolute value ($s_t=+1$) or an underestimation in absolute value ($s_t=-1$).
\autoref{fig:experimentalpredictionfirst}A shows a schematic for the case that both the current and prior beliefs are Gaussian distributions.
The two observations indicated by dashed lines have same absolute error $|\delta_t|$, but differ in the sign bias $s$.

Given an absolute prediction value $\approxs{y}>0$, an absolute prediction error $\delta>0$, a confidence value $C>0$, and a sign bias $s \in \{ -1, 1 \}$, we can compute the average of $M_t$ over time for the time points with $|\approxs{y}_t| \approx \approxs{y}$, $|\delta_t| \approx \delta$, $C_t \approx C$, and $s_t = s$, which we denote as $\Bar{M}_1(\approxs{y},\delta,s,C)$ -- the index 1 stands for experimental prediction 1.
The approximation notation $\approx$ is used for continuous variables instead of equality, due to practical limitations, i.e. for obtaining adequate number of samples for averaging.
Note that for our theoretical proofs we use equality, but in our simulation we include the practical limitations of a real experiment, and hence, use an approximation.
The formal definitions can be found in Methods.
It is worth noting that the quantity $\Bar{M}_1(\approxs{y},\delta,s,C)$ is model independent; its calculation does not require any assumption on the learning algorithm the subject may employ.
Depending on whether the measurement $\Bar{M}_1(\approxs{y},\delta,s,C)$ reflects $\textbf{S}_{\mathrm{Sh}}$ or $\textbf{S}_{\mathrm{BF}}$, its relationship to the defined four variables (i.e. $\approxs{y}$, $\delta$, $s$, $C$) is qualitatively and quantitatively different.

In order to prove and illustrate our prediction, let us consider each subject as an agent enabled with one of the learning algorithms that we discussed.
Similar to above, given an absolute prediction $\approxs{\theta}>0$ (corresponding to the subjects' absolute prediction $\approxs{y}$), an absolute prediction error $\delta>0$, a standard deviation $\sigma_C$ (corresponding to the subjects' confidence value $C$), and a sign bias $s \in \{ -1, 1 \}$, we can compute the average Shannon Surprise $\textbf{S}_{\mathrm{Sh}}(y_{t}; \approxs{\pi}^{(t-1)})$ and the average Bayes Factor Surprise $\textbf{S}_{\mathrm{BF}}(y_{t}; \approxs{\pi}^{(t-1)})$
over time,
for the time points with $|\approxs{\theta}_{t-1}| \approx \approxs{\theta}$, $|\delta_t| \approx \delta $, $\approxs{\sigma}_t \approx \sigma_C$, and $s_t = s$, which we denote as $\Bar{\textbf{S}}_{\mathrm{Sh}}(\approxs{\theta},\delta,s,\sigma_C)$ and $\Bar{\textbf{S}}_{\mathrm{BF}}(\approxs{\theta}, \delta,s,\sigma_C)$ respectively.
We can show theoretically (see Methods) and in simulations (see \autoref{fig:experimentalpredictionfirst}B and Methods) that for any value of $\approxs{\theta}$, $\delta$, and $\sigma_C$, we have $\Bar{\textbf{S}}_{\mathrm{Sh}}(\approxs{\theta},\delta,s=+1,\sigma_C) > \Bar{\textbf{S}}_{\mathrm{Sh}}(\approxs{\theta},\delta,s=-1,\sigma_C)$ for the Shannon Surprise, and exactly the opposite relation, i.e. $\Bar{\textbf{S}}_{\mathrm{BF}}(\approxs{\theta},\delta,s=+1,\sigma_C) < \Bar{\textbf{S}}_{\mathrm{BF}}(\approxs{\theta},\delta,s=-1,\sigma_C)$
for the Bayes Factor Surprise.
Moreover, this effect increases with increasing $\delta$.

It should be noted that such an effect is due to the essential difference of $\textbf{S}_{\mathrm{Sh}}$ and $\textbf{S}_{\mathrm{BF}}$ in using the prior belief $\pi^{(0)}(\theta)$.
Our experimental prediction is theoretically provable for the cases that each subject's belief $\approxs{\pi}^{(t)}$ is a Gaussian distribution, which is the case if they employ VarSMiLe, Nas10$^{*}$, Nas12$^{*}$, pf1, MP1, or Leaky Integrator as their learning rule (see Methods).
For the cases that different learning rules (e.g. pf20) are used, where the posterior belief is a weighted sum of Gaussians, the theoretical analysis is more complicated, but our simulations show the same results (see \autoref{fig:experimentalpredictionfirst}B and Methods).
Therefore, independent of the learning rule, we have the same experimental prediction on the manifestation of different surprise measures on physiological signals, such as pupil dilation.
Our first experimental prediction can be summarized as a set of hypotheses shown in Table \ref{table:Hyp_3}.

\begin{table}[t]
\centering
\caption{\textbf{Experimental Hypotheses and Predictions 1.} $ \Delta \Bar{M}_1(\approxs{\theta},\delta,C)$ stands for $ \Bar{M}_1(\approxs{\theta},\delta,s=+1,C) - \Bar{M}_1(\approxs{\theta},\delta,s=-1,C)$}
\label{table:Hyp_3}
\begin{tabular}{ | p{6.2cm} | p{7.5cm} |}
 \hline
 \textbf{Hypothesis} & \textbf{Prediction}\\
 \hline
  The indicator reflects $\textbf{S}_{\mathrm{BF}}$  &
  $ \Delta \Bar{M}_1(\approxs{\theta},\delta,C) < 0$ and $\frac{\partial \Delta \Bar{M}_1(\approxs{\theta},\delta,C)}{\partial \delta} < 0$\\
 \hline
 The indicator reflects $\textbf{S}_{\mathrm{Sh}}$  &
  $ \Delta \Bar{M}_1(\approxs{\theta},\delta,C) > 0$ and $\frac{\partial \Delta \Bar{M}_1(\approxs{\theta},\delta,C)}{\partial \delta} > 0$\\
 \hline
 The prior is not used for inference & $ \Delta \Bar{M}_1(\approxs{\theta},\delta,C) = 0$ \\
 \hline
\end{tabular}
\end{table}

\subsubsection{Prediction 2}

Our second prediction follows the same experimental procedure as the one for the first prediction.
The main difference is that for the second prediction we need to fit a model to the experimental data.
Given one of the learning algorithms, the fitting procedure can be done by tuning the free parameters of the algorithm with the goal of minimizing the mean squared error between the model's prediction $\approxs{\theta}_t$ and a subject's prediction $\approxs{y}_{t+1}$ (similar to \cite{nassar2010approximately, nassar2012rational}) or with the goal of maximizing the likelihood of subject's prediction $\approxs{\pi}^{(t)}(\approxs{y}_{t+1})$.
Our prediction is independent of the learning algorithm, but in an actual experiment, we recommend to use model selection to find the model that fits the human data best.

\begin{figure}
    \centering
\makebox[\textwidth][c]{\includegraphics{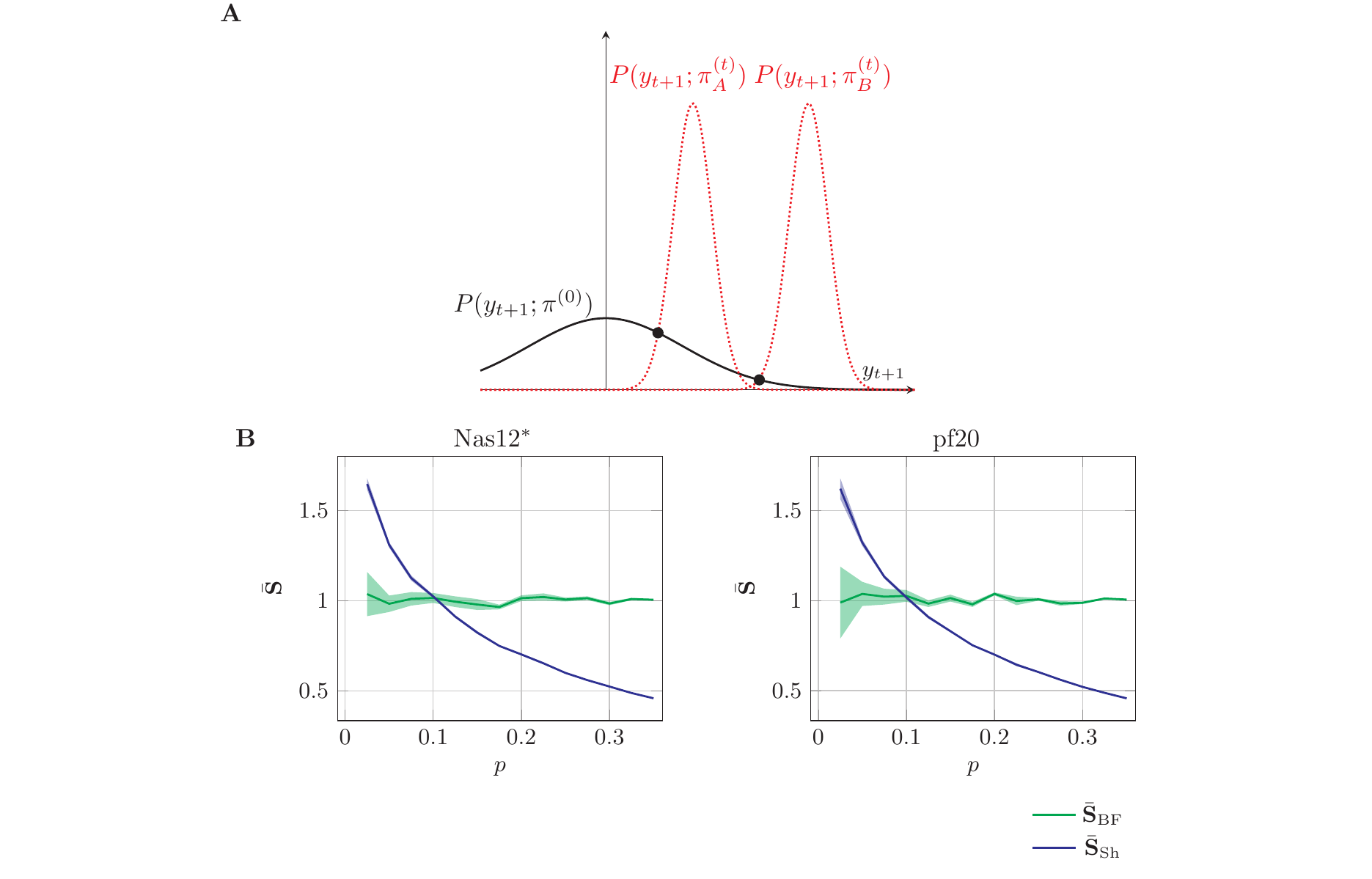}}
    \caption{\textbf{Experimental prediction 2.}
    \textbf{A.}
    Schematic of the task for the case of a Gaussian belief.
    The probability distribution of observations under the prior belief is shown by the solid black curve.
    Two different possible current beliefs (determined by the letters $A$ and $B$) are shown by dashed red curves.
    The intersections of the dashed red curves with the prior belief determine observations whose $\textbf{S}_{\mathrm{BF}}$ is same and equal to one, but their $\textbf{S}_{\mathrm{Sh}}$ is a function of their probabilities under the prior belief $p$.
    \textbf{B.}
    The average surprise values $\Bar{\textbf{S}}_{\mathrm{Sh}}(p)$ and $\Bar{\textbf{S}}_{\mathrm{BF}}(p)$ over 20 subjects (each with 500 observations) are shown for two different learning algorithms (Nas12$^*$ and pf20).
    The mean $\textbf{S}_{\mathrm{BF}}$ is constant (equal to 1) and independent of $p$, whereas the mean $\textbf{S}_{\mathrm{Sh}}$ is a decreasing function of $p$.
    The shaded area corresponds to the standard error of the mean.
    The experimental task is the same as the Gaussian task we used in the previous section.
    Observations $y_t$ are drawn from a Gaussian distribution with $\sigma = 0.5$, whose mean changes with change point probability $p_c = 0.1$ (see Methods for details).
}\label{fig:experimentalpredictionsecond}
\end{figure}

Having a fitted model, we can compute the probabilities $P(y_{t+1};\approxs{\pi}^{(t)})$ and $P(y_{t+1};\approxs{\pi}^{(0)})$.
For the case that these probabilities are equal, i.e. $P(y_{t+1};\approxs{\pi}^{(t)}) = P(y_{t+1};\approxs{\pi}^{(0)}) = p$, the Bayes Factor Surprise $\textbf{S}_{\mathrm{BF}}$ is equal to 1, independent of the value of $p$ (cf. \autoref{Eq:S_GM}).
However, the Shannon Surprise $\textbf{S}_{\mathrm{Sh}}$ is equal to $- \log p$, and varies with $p$.
\autoref{fig:experimentalpredictionsecond}A shows a schematic for the case that both current and prior beliefs are Gaussian distributions.
Two cases for which we have $P(y_{t+1};\approxs{\pi}^{(t)}) = P(y_{t+1};\approxs{\pi}^{(0)}) = p$, for two different $p$ values, are marked by black dots at the intersections of the curves.

Given a probability $p>0$, we can compute the average of $M_t$ over time for the time points with $P(y_{t+1};\approxs{\pi}^{(t)}) \approx p$ and $P(y_{t+1};\approxs{\pi}^{(0)}) \approx p$, which we denote as $\Bar{M}_2(p)$
-- the index 2 stands for experimental prediction 2.
Analogous to the first prediction, the approximation notation $\approx$ is used due to practical limitations.
Then, if $\Bar{M}_2(p)$ is independent of $p$, its behavior is consistent with $\textbf{S}_{\mathrm{BF}}$, whereas if it decreases by increasing $p$, it can be a signature of $\textbf{S}_{\mathrm{Sh}}$.
Our second experimental prediction can be summarized as two hypotheses shown in Table \ref{table:Hyp_4}.
Note that in contrast to our first prediction, with the assumption that the standard deviation of the prior belief is fitted using the behavioral data, we do not consider the hypothesis that the prior is not used for inference, because this is indistinguishable from a very large variance of the prior belief.

In order to illustrate the possible results and the feasibility of the experiment, we ran a simulation and computed $\Bar{\textbf{S}}_{\mathrm{BF}}(p)$ and $\Bar{\textbf{S}}_{\mathrm{Sh}}(p)$ for the time points with $P(y_{t+1};\approxs{\pi}^{(t)}) \approx p$ and $P(y_{t+1};\approxs{\pi}^{(0)}) \approx p$ (see Methods for details).
The results of the simulation are shown in \autoref{fig:experimentalpredictionsecond}B.

\begin{table}[t]
\caption{\textbf{Experimental Hypotheses and Predictions 2.}}
\label{table:Hyp_4}
\centering
\begin{tabular}{ | p{6cm} | p{3cm} | }
 \hline
 \textbf{Hypothesis} & \textbf{Prediction}\\
 \hline
  The indicator reflects $\textbf{S}_{\mathrm{BF}}$  &
 $\frac{\partial \Bar{M}_2(p)}{\partial p} = 0$\\
 \hline
 The indicator reflects $\textbf{S}_{\mathrm{Sh}}$  &
  $\frac{\partial \Bar{M}_2(p)}{\partial p} < 0$\\
\hline
\end{tabular}
\end{table}

\section{Discussion}
\label{discussion}

We have shown that performing exact Bayesian inference on a generative world model naturally leads to a definition of surprise and a surprise-modulated adaptation rate.
We have proposed three approximate algorithms (VarSMiLe, MP$N$, and pf$N$) for learning in non-stationary environments, which all exhibit the surprise-modulated adaptation rate of the exact Bayesian approach and are biologically plausible.
Empirically we observed that our algorithms achieve levels of performance comparable to approximate Bayesian methods with higher memory demands \citep{adams2007bayesian}, and are more resilient across different environments compared to methods with similar memory demands \citep{fearnhead2007line, nassar2010approximately,
nassar2012rational, faraji2018balancing}.

Learning in a volatile environment has been studied for a long time in the fields of Bayesian learning, neuroscience, and signal processing.
In the following, we discuss the biological relevance of our work, and we briefly review some of the previously developed algorithms,
with particular focus on the ones that have studied environments which can be modeled with a generative model similar to the one in \autoref{fig:gen_model}.
We then discuss further our results, and propose directions for future work on surprise-based learning.

\subsection{Biological interpretation}

Humans are able to quickly adapt to changes \citep{nassar2012rational, nassar2010approximately, behrens2007learning}, but human behaviour is also often observed to be suboptimal, compared to the normative approach of exact Bayesian inference \citep{mathys2011bayesian,
wilson2013mixture, nassar2010approximately, glaze2015normative, prat2020human}.
In general, biological agents have limited resources and possibly inaccurate assumptions about hyper-parameters, yielding sub-optimal behaviour, as we also see with our algorithms whose accuracies degrade with
a sub-optimal choice of hyper-parameters.
Performance also deteriorates with a decreasing number of particles in the sampling-based algorithms, which might be another possible explanation of suboptimal human behavior.
Previously, Particle Filtering has been shown to explain the behavior of human subjects in changing environments: \cite{daw2008pigeon} use a single particle, \citep{brown2009detecting} use a simple heuristic form of particle filtering based on direct simulation, \cite{findling2019imprecise} combine Particle Filtering with a noisy inference, and \cite{prat2020human} use it for a task with temporal structure.

At the level of neuronal implementation,
we do not propose a specific suggestion.
However, there are several hypotheses about neural implementations of related particle filters \citep{kutschireiter2017nonlinear, shi2009neural, huang2014neurons, legenstein2014ensembles},
on which, a neural model of pf$N$ and -- its greedy version -- MP$N$ could be based. In a similar spirit,
the updating scheme of Variational SMiLe may be implemented in biological neural networks (for distributions in the exponential family).

Our theoretical framework for modulation of learning by the Bayes Factor Surprise $\textbf{S}_{\mathrm{BF}}$ is related to
the body of literature on neo-Hebbian three-factor learning rules \citep{lisman2011neohebbian, fremaux2016neuromodulated, gerstner2018eligibility}, where a third factor indicating reward or surprise enables or modulates a synaptic change or a belief update \citep{angela2005uncertainty,
angela2012change}.
We have shown how Bayesian or approximate Bayesian inference naturally leads to such a third factor that modulates learning via the surprise modulated adaptation rate $\gamma (\textbf{S}_{\mathrm{BF}}, m)$.
This may offer novel interpretations of behavioural and neurophysiological data, and help in understanding how three-factor learning computations may be implemented in the brain.

\subsection{Related work}
\label{related_work}

\paragraph{Exact Bayesian inference}
As already described in the ``Message-Passing $N$'' section of the Results, for the generative model in \autoref{fig:gen_model}, it is possible to find an exact online Bayesian update of the belief using a message passing algorithm \citep{adams2007bayesian}.
The space and time complexity of the algorithm increases linearly with $t$, which makes it unsuitable for an online learning setting.
However, approximations like dropping messages below a certain threshold \citep{adams2007bayesian} or stratified resampling \citep{fearnhead2007line} allow to reduce the computational complexity.
The former has a variable number of particles in time, and the latter needs solving a complicated non-linear equation at each time step in order to reduce the number of particles to $N$ (called SOR$N$ in the Results section).

Our message passing algorithm with finite number of particles (messages) $N$ (MP$N$, Algo. 3) is closely related to these algorithms and can be seen as a biologically more plausible variant of the other two.
All three algorithms have the same update rules given by \autoref{Eq:MP_w_oldparticles} and \autoref{Eq:MP_w_newparticle}.
Hence the algorithms of both \cite{adams2007bayesian} and \cite{fearnhead2007line} have the same surprise modulation as our MP$N$.
The difference lies in their approaches to eliminate less ``important'' particles.

In the literature of switching state-space models \citep{barber2012bayesian}, the generative models of the kind in \autoref{fig:gen_model} are known as ``reset models'', and the message passing algorithm of \cite{adams2007bayesian} is known to be the standard algorithm for inference over these models \citep{barber2012bayesian}.
See \cite{barber2012bayesian, barber2006expectation, ghahramani2000variational} for other variations of switching state-space models and examples of approximate inference over them.

\paragraph{Leaky integration and variations of delta-rules}
In order to estimate some statistics, leaky integration of new observations is a particularly simple form of a trade-off between integrating and forgetting.
After a transient phase, the update of a leaky integrator takes the form of a delta-rule that can be seen as an approximation of exact Bayesian updates \citep{meyniel2016human,heilbron2019confidence,angela2009sequential,ryali2018demystifying}.
This update rule was found to be biologically plausible and consistent with human behavioral data \citep{meyniel2016human,angela2009sequential}.
However, \cite{behrens2007learning} and \cite{heilbron2019confidence} demonstrated that in some situations, the exact Bayesian model is significantly better than leaky integration in explaining human behavior.
The inflexibility of leaky integration with a single, constant leak parameter can be overcome by a weighted combination of multiple leaky integrators \citep{wilson2013mixture}, where the weights are updated in a similar fashion as in the exact online methods \citep{adams2007bayesian,fearnhead2007line}, or by considering an adaptive leak parameter \citep{nassar2012rational, nassar2010approximately}.
We have shown that the two algorithms of \cite{nassar2010approximately, nassar2012rational} can be generalized to Gaussian prior beliefs (Nas10$^{*}$ and  Nas12$^{*}$).
Our results show that these algorithms also inherit the surprise-modulation of the exact Bayesian inference.
Our surprise-dependent adaptation rate $\gamma$ can be interpreted as a surprise-modulated leak parameter.

\paragraph{Other approaches}
Learning in the presence of abrupt changes has also been considered without explicit assumptions about the underlying generative model.
One approach uses a surprise-modulated adaptation rate \citep{faraji2018balancing} similar to \autoref{Eq:Gamma_def}.
The Surprise-Minimization Learning (SMiLe) algorithm of \cite{faraji2018balancing} has an updating rule similar to the one of VarSMiLe (\autoref{Eq:Var_SMiLe_Min_1_2nd_Ver} and \autoref{Eq:Var_SMiLe_Min_1_2nd_Ver_Const}).
The adaptation rate modulation, however, is based on the Confidence Corrected Surprise \citep{faraji2018balancing} rather than the Bayes Factor Surprise, and the trade-off in its update rule is between resetting and staying with the latest belief rather than between resetting and integrating (see Methods).

Other approaches use different generative models, such as conditional sampling of the parameters also when there is a change \citep{angela2005uncertainty, glaze2015normative}, a deeper hierarchy without fixed change probability $p_c$ \citep{wilson2010bayesian}, or drift in the parameters \citep{mathys2011bayesian,gershman2014statistical}.
A recent work shows that inference on the generative model of \autoref{fig:gen_model}
can explain human behavior well even when the true generative model of the environment is different and more complicated \citep{findling2019imprecise}.
They develop a heuristic approach to add noise in the inference process of a Particle Filter.
Their algorithm can be interpreted as a surprise-modulated Particle Filter, where the added noise scales with a measure of surprise (conceptually equivalent to Bayesian surprise \citep{storck1995reinforcement, schmidhuber2010formal, itti2006bayesian}).
Moreover, another recent work \citep{prat2020human} shows that approximate sampling algorithms (like Particle Filtering) can explain human behavior better than their alternatives in tasks closely related to the generative model of \autoref{fig:gen_model}.
The signal processing literature provides further methods to address the problem of learning in non-stationary environments with abrupt changes; see \cite{aminikhanghahi2017survey} for a review, and \cite{lin2017sharp, cummings2018differentially, ozkan2013marginalized, masegosa2017bayesian} for a few recent examples.

\subsection{Surprise-modulation as a generic phenomenon}
Learning rate modulation similar to the one in \autoref{Eq:Gamma_def} has been previously proposed in the neuroscience literature with either heuristic arguments \citep{faraji2018balancing} or with Bayesian arguments for a particular experimental task, e.g. when samples are drawn from a Gaussian distribution \citep{ nassar2010approximately, nassar2012rational}.
The fact that the same form of modulation is at the heart of Bayesian inference for our relatively general generative model, that it is derived without any further assumptions, and is not a-priori defined is in our view an important contribution to the field of adaptive learning algorithms in computational neuroscience.

Furthermore, the results of our three approximate methods (Particle Filtering, Variational SMiLe, and Message Passing with fixed $N$ number of messages) as well as some previously developed ones \citep{adams2007bayesian, fearnhead2007line, nassar2010approximately, nassar2012rational} demonstrate that the surprise-based modulation of the learning rate is a generic phenomenon.
Therefore, regardless of whether the brain uses Bayesian inference or an approximate algorithm
\citep{mathys2011bayesian, friston2010free, nassar2010approximately, nassar2012rational, gershman2014statistical, bogacz2017tutorial, bogacz2019dopamine, gershman2019does, findling2019imprecise, prat2020human},
the notion of Bayes Factor Surprise and the way it modulates learning (i.e. \autoref{Eq:tradeoff} and \autoref{Eq:Gamma_def}) look generic.

The generality of the way surprise should modulate learning depends on an agent's inductive biases about its environment and is directly associated with the assumed generative model of the world.
The generative model we considered in this work involves abrupt
changes.
However, one can think of other realistic examples, where an improbable observation does not indicate a persistent change, but a singular event or an outlier, similar to \cite{d2016neural, nassar2019statistical}.
In such situations, the belief should not be changed and surprise should attenuate learning, rather than accelerate it.
Interestingly, we can show that exact and approximate Bayesian inference on such a generative model naturally lead to a surprise-modulated adaptation rate $\gamma (\textbf{S}_{\mathrm{BF}}, m)$, with the same definition of $\textbf{S}_{\mathrm{BF}}$, where the trade-off is not between integrating and resetting, but between integrating and ignoring the new observation
\footnote{\normalsize Liakoni, V., Lehmann, M. P., Modirshanechi, A., Brea, J., Herzog, M. H., Gerstner, W., \& Preuschoff, K. (in preparation). Dissociating brain regions encoding reward prediction error and surprise.}.
This extends previous work on such environments \citep{d2016neural, nassar2019statistical} to a general setting and highlights the general principle of surprise-based modulation, given the prior knowledge on the structure of the environment.

An aspect that the generative model we considered does not capture is the potential return to a previous state of the environment, rather than a change to a completely new situation.
If in our example of \autoref{fig:gen_model}B, the bridge with the shortest path is temporarily closed for repairs, your friend would again have to take the longer detour, therefore, her arrival times will return to their previous values, i.e. increase.
In such cases, an agent should infer whether the surprising observation stems from a new hidden state or from an old state stored in memory.
Relevant generative models have been studied in \cite{fox2011bayesian, gershman2014statistical, gershman2017computational} and are out of the scope of our present paper.

\subsection{Bayes Factor Surprise as a novel measure of surprise}
\label{surp_disc}

In view of a potential application in the neurosciences, a definition of surprise should exhibit two properties:
(i) surprise should reflect how unexpected an event is, and,
(ii) surprise should modulate learning.
Surprising events indicate that our belief is far from the real world and suggest to update our model of the world, or, for large surprise, simply forget it.
Forgetting is the same as returning to the prior belief.
However, an observation $y_{t+1}$ can be unexpected under both the prior $\pi^{(0)}$ and the current beliefs $\pi^{(t)}$.
In these situations, it is not obvious whether forgetting helps.
Therefore, the modulation between forgetting or not should be based on a comparison between the probability of an event under the current belief $P(y_{t+1}; \pi^{(t)})$ and its probability under the prior belief $P(y_{t+1}; \pi^{(0)})$.

The definition of the Bayes Factor Surprise $\textbf{S}_{\mathrm{BF}}$ as the ratio of $P(y_{t+1}; \pi^{(t)})$ and $P(y_{t+1}; \pi^{(0)})$ exploits this insight.
The Bayes Factor Surprise appears as a modulation factor in the recursive form of the exact Bayesian update rule for a hierarchical generative model of the environment.
When two events are equally probable under the prior belief, the one which is less expected under the current belief is more surprising - satisfying the first property.
At the same time, when two events are equally probable under the current belief, the one which is more expected under the prior belief is more surprising - signaling that forgetting may be beneficial.

$\textbf{S}_{\mathrm{BF}}$ can be written (using \autoref{Eq:p_y_t}) in a more explicit way as
\begin{equation} \label{Eq:S_GM_ind_GM}
    \textbf{S}_{\mathrm{BF}}(y_{t+1}; \pi^{(t)}) = \frac{P(y_{t+1}; \pi^{(0)})}{P(y_{t+1}; \pi^{(t)})} = \frac{ \mathbb{E}_{\pi^{(0)}} \big[ P_Y(y_{t+1} | \Theta )  \big] }{ \mathbb{E}_{\pi^{(t)}} \big[ P_Y(y_{t+1} | \Theta )  \big] }.
\end{equation}
Note that the definition by itself is independent of the specific form of the generative model.
In other words, even in the cases where data is generated with another generative model (e.g. the real world), $\textbf{S}_{\mathrm{BF}}$ could be a candidate surprise measure in order to interpret brain activity or pupil dilation.

We formally discussed the connections between the Bayes Factor Surprise and Shannon Surprise \citep{shannon1948mathematical}, and showed that
they are closely linked.
We showed that the modulated adaptation rate ($\gamma$) used in (approximate) Bayesian inference is a function of the difference between the Shannon Surprise under the current and the prior beliefs, but cannot be expressed solely by the Shannon Surprise under the current one.
Our formal comparisons between these two different measures of surprise lead to specific experimentally testable predictions.

The Bayesian Surprise $\textbf{S}_{\mathrm{Ba}}$ \citep{storck1995reinforcement, schmidhuber2010formal, itti2006bayesian} and the Confidence Corrected Surprise $\textbf{S}_{\mathrm{CC}}$
\citep{faraji2018balancing} are two other measures of surprise in neuroscience.
The learning modulation derived in our generative model cannot be expressed as a function of $\textbf{S}_{\mathrm{Ba}}$ and $\textbf{S}_{\mathrm{CC}}$.
However, one can hypothesize that $\textbf{S}_{\mathrm{Ba}}$ is computed after the update of the belief to measure the information gain of the observed event, and is therefore not a good candidate for online learning modulation.
The Confidence Corrected surprise $\textbf{S}_{\mathrm{CC}}$ takes into account the shape of the belief, and therefore includes the effects of confidence, but it does not consider any information about the prior belief.
Hence, a result of $ \Bar{M}_1(\approxs{\theta},\delta,s=+1,C) = \Bar{M}_1(\approxs{\theta},\delta,s=-1,C)$ in our first experimental prediction would be consistent with the corresponding behavioral or physiological indicator reflecting the $\textbf{S}_{\mathrm{CC}}$.

\subsection{Difference in Shannon Surprise, an alternative perspective}

Following our formal comparison in the ``Experimental prediction'' section, $\textbf{S}_{\mathrm{BF}}$ can be expressed as a deterministic function of the difference in Shannon Surprise as
\begin{equation}
\begin{aligned}
  \label{Sgm_Shannon}
  \textbf{S}_{\mathrm{BF}} &= \frac{(1-p_c) e^ {\Delta \textbf{S}_{\mathrm{Sh}} } }{1 - p_c e^ {\Delta \textbf{S}_{\mathrm{Sh}} }}.
\end{aligned}
\end{equation}
All of our theoretical results can be rewritten by replacing $\textbf{S}_{\mathrm{BF}}$ with this function of $\Delta \textbf{S}_{\mathrm{Sh}}$.
Moreover, because there is a 1-to-1 mapping between $\textbf{S}_{\mathrm{BF}}$ and $\Delta \textbf{S}_{\mathrm{Sh}}$, from a systemic point of view, it is not possible to specify whether the brain computes the former or the latter by analysis of behavioral data and biological signals.
This suggests an alternative interpretation of surprise-modulated learning as an approximation of Bayesian inference:
What the brain computes and perceives as surprise or prediction error may be Shannon Surprise, but the modulating factor in a three-factor synaptic plasticity rule \citep{lisman2011neohebbian, fremaux2016neuromodulated, gerstner2018eligibility} may be implemented by comparing the Shannon Surprise values under the current and the prior beliefs.

\subsection{Future directions}
A natural continuation of our study is to test our experimental predictions in human behavior and physiological signals, in order to investigate which measures of surprise are used by the brain.
Along a similar direction, our approximate learning algorithms can be evaluated on human behavioral data from experiments that use a similar generative model \citep{nassar2012rational, nassar2010approximately,  wilson2013mixture, behrens2007learning, angela2005uncertainty, glaze2015normative, heilbron2019confidence}
in order to assess if our proposed algorithms achieve similar or better performance in explaining data.

Finally, our methods can potentially be applied to model-based reinforcement learning in non-stationary environments.
In recent years, there has been a growing interest in adaptive or continually learning agents in changing environments in the form of Continual learning and Meta-learning \citep{lomonaco2019continual, traore2019discorl}.
Many Continual learning model-based approaches make use of some procedure to detect changes \citep{lomonaco2019continual, nagabandi2018learning}.
Integrating $\textbf{S}_{\mathrm{BF}}$ and a learning rate $\gamma(\textbf{S}_{\mathrm{BF}})$ into a reinforcement learning agent would be an interesting future direction.

\section{Methods}

\subsection{Proof of the proposition}
By definition
\begin{equation}
  \label{Eq:Belief_Def_1_App}
  \pi^{(t+1)}(\theta)\equiv \textbf{P}(\Theta_{t+1}=\theta|y_{1:t+1}).
\end{equation}
We exploit the Markov property of the generative model in \autoref{Eq:GenModel}, \autoref{Eq:GenModel2}, and \autoref{Eq:GenModel3}, condition on the fixed past $y_{1:t}$ and rewrite
\begin{equation}
  \label{Eq:Belief_Def_2_App}
\pi^{(t+1)}(\theta) = \frac{ P_{Y}(y_{t+1} | \theta) \textbf{P}(\Theta_{t+1}=\theta|y_{1:t}) }{\textbf{P}( y_{t+1}|y_{1:t})}.
\end{equation}
By marginalization over the hidden state $C_{t+1}$, the second factor in the numerator of \autoref{Eq:Belief_Def_2_App} can be written as
\begin{equation}
\begin{aligned}
\textbf{P}(\Theta_{t+1}=\theta|y_{1:t})
&= (1-p_c) \pi^{(t)}(\theta) + p_c \pi^{(0)}(\theta).
\end{aligned}
\end{equation}
The denominator in \autoref{Eq:Belief_Def_2_App} can be written as
\begin{equation}
\begin{aligned}
\textbf{P}(y_{t+1}|&y_{1:t}) = \int P_Y(y_{t+1}|\theta) \textbf{P}(\Theta_{t+1}=\theta|y_{1:t}) d \theta\\
&= (1-p_c) \int P_Y(y_{t+1}|\theta) \pi^{(t)}(\theta) d \theta + p_c \int P_Y(y_{t+1}|\theta) \pi^{(0)}(\theta) d \theta\\
&= (1-p_c) P(y_{t+1}; \pi^{(t)}) + p_c P(y_{t+1}; \pi^{(0)}).
\end{aligned}
\end{equation}
where we used the definition in \autoref{Eq:p_y_t}.
Using these two expanded forms, \autoref{Eq:Belief_Def_2_App} can be rewritten
\begin{equation}
\begin{aligned}
\label{Eq:pi_post_1}
\pi^{(t+1)}(\theta) &= \frac{ P_Y(y_{t+1}|\theta) \Big( (1-p_c) \pi^{(t)}(\theta) + p_c \pi^{(0)}(\theta) \Big) }{ (1-p_c) P(y_{t+1}; \pi^{(t)}) + p_c P(y_{t+1}; \pi^{(0)}) }.
\end{aligned}
\end{equation}
We define $P(\theta|y_{t+1})$ as the posterior given a change in the environment as
\begin{equation}
\begin{aligned}
P(\theta|y_{t+1}) =
\frac{P_Y(y_{t+1}|\theta) \pi^{(0)}(\theta)} {P(y_{t+1}; \pi^{(0)})}.
\end{aligned}
\end{equation}
Then, we can write \autoref{Eq:pi_post_1} as
\begin{equation}
\begin{aligned}
\label{Eq:pi_post_2}
\pi^{(t+1)}(\theta) &= \frac{ (1-p_c) P(y_{t+1}; \pi^{(t)}) \pi_B^{(t+1)}(\theta) + p_c P(y_{t+1}; \pi^{(0)}) P(\theta|y_{t+1}) }{ (1-p_c) P(y_{t+1}; \pi^{(t)}) + p_c P(y_{t+1}; \pi^{(0)}) }\\
&= \frac{ \pi_B^{(t+1)}(\theta) + \frac{p_c}{1-p_c} \frac{P(y_{t+1}; \pi^{(0)})}{P(y_{t+1}; \pi^{(t)})}  P(\theta|y_{t+1}) }{ 1 + \frac{p_c}{1-p_c} \frac{P(y_{t+1}; \pi^{(0)})}{P(y_{t+1}; \pi^{(t)})} }\\
&= (1 - \gamma_{t+1}) \pi^{(t+1)}_B(\theta) + \gamma_{t+1} P(\theta|y_{t+1}),
\end{aligned}
\end{equation}
where $\pi^{(t+1)}_B(\theta)$ is defined in \autoref{Eq:pi_B}, and
\begin{equation}
\begin{aligned}
\gamma_{t+1} &= \gamma \Big(\textbf{S}_{\mathrm{BF}}(y_{t+1}; \pi^{(t)}), \frac{p_c}{1-p_c} \Big)\\
\end{aligned}
\end{equation}
with $\textbf{S}_{\mathrm{BF}}$ defined in \autoref{Eq:S_GM}, and $\gamma(\text{S}, m)$ defined in \autoref{Eq:Gamma_def}.
Thus our calculation yields a specific choice of surprise ($\text{S} = \textbf{S}_{\mathrm{BF}}$) and a specific value for the saturation parameter $m = \frac{p_c}{1-p_c}$.

\subsection{Derivation of the optimization-based formulation of VarSMiLe  (Algo. 1)}

To derive the optimization-based update rule for the Variational SMiLe rule and the relation of the bound $B_{t+1}$ with surprise, we used the same approach used in \cite{faraji2018balancing}.

\paragraph{Derivation of the update rule}
Consider the general form of the following variational optimization problem:
\begin{equation}
\begin{aligned}
\label{Eq:Genral_Opt}
q^*(\theta) &= \text{argmin }\textbf{D}_{KL} \big[ q(\theta) || p_1(\theta) \big]\\
& q(\theta) \text{  s.t.  }  \textbf{D}_{KL} \big[ q(\theta) || p_2(\theta) \big] < B \text{ and } \mathbb{E}_q[1]=1,
\end{aligned}
\end{equation}
where $B \in \big[ 0, \textbf{D}_{KL}[ p_1(\theta) || p_2(\theta)] \big]$.
On the extremes of $B$, we will have trivial solutions
\begin{equation}
\begin{aligned}
q^*(\theta)= \left\{
  \begin{array}{lr}
    p_2(\theta) & \text{if} \quad B=0\\
    p_1(\theta) & \text{if} \quad B = \textbf{D}_{KL}[ p_1(\theta) || p_2(\theta)].
  \end{array}
\right.\\
\end{aligned}
\end{equation}

Note that the Kullback–Leibler divergence is a convex function with respect to its first argument, i.e. $q$ in our setting.
Therefore, both the objective function and the constraints of the optimization problem in \autoref{Eq:Genral_Opt} are convex.
For convenience, we assume that the parameter space for $\theta$ is discrete, but the final results can be generalized also to the continuous case with some considerations - see \cite{beal2003variational} and \cite{faraji2018balancing}.
For the discrete setting, the optimization problem in \autoref{Eq:Genral_Opt} can be rewritten as
\begin{equation}
\begin{aligned}
\label{Eq:Genral_Opt_Disc}
q^*(\theta) &= \text{argmin } \sum_{\theta} q(\theta) \big( \text{log}(q(\theta)) - \text{log}(p_1(\theta)) \\ & q(\theta) \text{  s.t.  }  \sum_{\theta} q(\theta) \big( \text{log}(q(\theta)) - \text{log}(p_2(\theta)) < B \text{ and } \sum_{\theta} q(\theta) = 1.
\end{aligned}
\end{equation}

For solving the mentioned problem, one should find a $q$ which satisfies the Karush–Kuhn–Tucker (KKT) conditions \citep{boyd2004convex} for
\begin{equation}
\mathcal{L} =   \sum_{\theta} q(\theta) \text{log}\Big( \frac{q(\theta)}{p_1(\theta)} \Big) + \lambda \sum_{\theta} q(\theta) \text{log}\Big( \frac{q(\theta)}{p_2(\theta)} \Big) - \lambda B + \alpha -  \alpha \sum_{\theta} q(\theta),
\end{equation}
\begin{equation}
\begin{aligned}
\frac{\partial \mathcal{L}}{\partial q(\theta)}  &= \text{log}\Big( \frac{q(\theta)}{p_1(\theta)} \Big) + 1 + \lambda \text{log}\Big( \frac{q(\theta)}{p_2(\theta)} \Big) + \lambda - \alpha \\ &= (1+\lambda)\text{log}( q(\theta)) - \text{log}( p_1(\theta)) - \lambda \text{log}( p_2(\theta)) + 1 + \lambda - \alpha,
\end{aligned}
\end{equation}
where $\lambda$ and $\alpha$ are the parameters of the dual problem.
Defining $\gamma = \frac{\lambda}{1+\lambda}$, and considering the partial derivative to be zero, we have
\begin{equation}
\text{log}( q^*(\theta)) = (1-\gamma) \text{log}( p_1(\theta)) + \gamma \text{log}( p_2(\theta)) + \text{Const}(\alpha, \gamma),
\end{equation}
where $\alpha$ is always specified in a way to have $\text{Const}(\alpha, \gamma)$ as the normalization factor
\begin{equation}
\begin{aligned}
\text{Const} & (\alpha, \gamma) = -\text{log}(Z(\gamma))\\
& \text{where } Z(\gamma) = \sum_{\theta} p_1^{1-\gamma}(\theta) p_2^{\gamma}(\theta).
\end{aligned}
\end{equation}
According to the KKT conditions, $\lambda \geq 0$, and as a result $\gamma \in [0,1]$.
Therefore, considering $p_1(\theta)=\approxs{\pi}_B^{(t+1)}(\theta)$ and $p_2(\theta)=P(\theta|y_{t+1})$, the solution to the optimization problem of \autoref{Eq:Var_SMiLe_Min_1_2nd_Ver} and \autoref{Eq:Var_SMiLe_Min_1_2nd_Ver_Const} is \autoref{Eq:Var_SMiLe_Log_Rec_2nd_Ver}.

\paragraph{Proof of the claim that $B$ is a decreasing function of Surprise}
According to the KKT conditions
\begin{equation}
\lambda \Big( \textbf{D}_{KL}[ q^{*}(\theta) || p_2(\theta)] - B \Big) = 0.
\end{equation}
For the case that $\lambda \neq 0$ (i.e. $\gamma \neq 0$), we have $B$ as a function of $\gamma$
\begin{equation}
\begin{aligned}
B(\gamma) &= \textbf{D}_{KL}[ q^{*}(\theta) || p_2(\theta)]\\
&= (1-\gamma) \mathbb{E}_{q^{*}} \Big[ \text{log} \big( \frac{p_1(\theta)}{p_2(\theta)} \big) \Big] - \text{log} \big( Z(\gamma) \big).
\end{aligned}
\end{equation}
Now, we show that the derivate of $B(\gamma)$ with respect to $\gamma$ is always non-positive.
To do so, we first compute the derivative of $Z(\gamma)$ as
\begin{equation}
\begin{aligned}
\frac{\partial \text{log}(Z(\gamma))}{\partial \gamma}  &= \frac{1}{Z(\gamma)} \frac{\partial}{\partial \gamma} \sum_{\theta} p_1^{1-\gamma}(\theta) p_2^{\gamma}(\theta)\\
&= \frac{1}{Z(\gamma)} \sum_{\theta} p_1^{1-\gamma}(\theta) p_2^{\gamma}(\theta) \text{log} \big( \frac{p_2(\theta)}{p_1(\theta)} \big)\\
&= \mathbb{E}_{q^{*}} \Big[ \text{log} \big( \frac{p_2(\theta)}{p_1(\theta)} \big)\Big],
\end{aligned}
\end{equation}
and the derivate of $\mathbb{E}_{q^{*}} [ h(\theta)]$ for an arbitrary $h(\theta)$ as
\begin{equation}
\begin{aligned}
\frac{\partial \mathbb{E}_{q^{*}} [ h(\theta)]}{\partial \gamma}  &= \frac{\partial }{\partial \gamma} \sum_{\theta} q^{*}(\theta) h(\theta)\\
&= \sum_{\theta} q^{*}(\theta) h(\theta) \frac{\partial }{\partial \gamma} \text{log}(q^{*}(\theta))\\
&= \sum_{\theta} q^{*}(\theta) h(\theta) \frac{\partial }{\partial \gamma} \Big( (1-\gamma) \text{log}( p_1(\theta)) + \gamma \text{log}( p_2(\theta)) -\text{log}(Z(\gamma))\Big)\\
&= \mathbb{E}_{q^{*}} \Big[ h(\theta) \text{log} \big( \frac{p_2(\theta)}{p_1(\theta)} \big)\Big] - \mathbb{E}_{q^{*}} \big[ h(\theta) \big] \mathbb{E}_{q^{*}} \Big[ \text{log} \big( \frac{p_2(\theta)}{p_1(\theta)} \big)\Big].
\end{aligned}
\end{equation}

Using the last three equations, we have
\begin{equation}
\begin{aligned}
\frac{\partial B(\gamma)}{\partial \gamma}  &= -(1-\gamma) \Big( \mathbb{E}_{q^{*}} \Big[ \Big( \text{log} \big( \frac{p_2(\theta)}{p_1(\theta)} \big) \Big)^2 \Big] - \mathbb{E}_{q^{*}} \Big[ \text{log} \big( \frac{p_2(\theta)}{p_1(\theta)} \big)\Big]^2 \Big)\\
&= -(1-\gamma) \textbf{Var}_{q^{*}} \Big[ \text{log} \big( \frac{p_2(\theta)}{p_1(\theta)} \big) \Big] \leq 0,
\end{aligned}
\end{equation}
which means that $B$ is a decreasing function of $\gamma$.
Because $\gamma$ is an increasing function of surprise, $B$ is also a decreasing function of surprise.

\subsection{Derivations of Message Passing $N$ (Algo. 2)}
For the sake of clarity and coherence, we repeat here some steps performed in the Results section.

Following the idea of \cite{adams2007bayesian} let us first define the random variable $R_t = \min \{ n \in \mathbb{N} : C_{t-n+1} = 1\}$.
This is the time window from the last change point.
Then the exact Bayesian form for $\pi^{(t)}(\theta)$ can be written as
\begin{equation}
\begin{aligned}
  \pi^{(t)}(\theta) &= \textbf{P}(\Theta_{t+1} = \theta |y_{1:t})\\
  &= \sum_{r_t=1}^{t} \textbf{P}(r_t |y_{1:t}) \textbf{P}(\Theta_{t+1} = \theta | r_t, y_{1:t}).
\end{aligned}
\end{equation}
To have a formulation similar to the one of Particle Filtering we rewrite the belief as
\begin{equation}
\begin{aligned}
  \label{Eq:MPN_particle_form}
  \pi^{(t)}(\theta) &= \sum_{k=0}^{t-1} w^{(k)}_t \textbf{P}(\Theta_{t+1} = \theta | R_t = t-k, y_{1:t}),= \sum_{k=0}^{t-1} w^{(k)}_t \pi^{(t)}_k(\theta),
\end{aligned}
\end{equation}
where $\pi^{(t)}_k(\theta) = \textbf{P}(\Theta_{t} = \theta | R_t = t-k, y_{1:t})$ is the term corresponding to $R_t = t-k$, and $w^{(k)}_t = \textbf{P}(R_t = t-k |y_{1:t})$ is its corresponding at time $t$.

To update the belief after observing $y_{t+1}$, one can use the exact Bayesian recursive formula \autoref{Eq:Bayesian_Rec_Formula}, for which one needs to compute $\pi_B^{(t+1)}(\theta)$ as
\begin{equation}
\begin{aligned}
  \pi_B^{(t+1)}(\theta) &= \frac{\pi^{(t)}(\theta)  P_Y(y_{t+1}|\theta)} {P(y_{t+1}; \pi^{(t)})}\\
  &= \frac{P_Y(y_{t+1}|\theta)}{P(y_{t+1}; \pi^{(t)})} \sum_{k=0}^{t-1} w^{(k)}_t \textbf{P}(\Theta_{t+1} = \theta | R_t = t-k, y_{1:t}).\\
\end{aligned}
\end{equation}
Using Bayes' rule and the conditional independence of observations, we have
\begin{equation}
\begin{aligned}
  \pi_B^{(t+1)}(\theta) &= \frac{P_Y(y_{t+1}|\theta)}{P(y_{t+1}; \pi^{(t)})} \sum_{k=0}^{t-1} w^{(k)}_t \frac{ \textbf{P}(y_{k+1:t} | \Theta_{t+1} = \theta, R_t = t-k) \pi^{(0)}(\theta) } { \textbf{P}( y_{k+1:t} | R_t = t-k)  } \\
  &= \frac{1}{P(y_{t+1}; \pi^{(t)})} \sum_{k=0}^{t-1} w^{(k)}_t \frac{ \prod_{i=k+1}^{t+1} P_Y( y_i | \theta) \pi^{(0)}(\theta) } { \textbf{P}( y_{k+1:t} | R_t = t-k)  }\, ,
  \end{aligned}
\end{equation}
and once again, by using the Bayes' rule and the conditional independence of observations, we find
\begin{equation}
\begin{aligned}
  \pi_B^{(t+1)}(\theta) &= \frac{1}{P(y_{t+1}; \pi^{(t)})} \sum_{k=0}^{t-1} w^{(k)}_t \frac{ \textbf{P}( y_{k+1:t+1} | R_{t+1} = t-k+1) } { \textbf{P}( y_{k+1:t} | R_t = t-k)  } \times\\
  & \quad \quad \textbf{P}(\Theta_{t+1} = \theta | R_{t+1} = t-k+1, y_{k+1:t+1})\\
  &= \frac{1}{P(y_{t+1}; \pi^{(t)})} \sum_{k=0}^{t-1} w^{(k)}_t \textbf{P}( y_{t+1} | R_{t+1} = t-k+1, y_{1:t} ) \times\\
  & \quad \quad \textbf{P}(\Theta_{t+1} = \theta | R_{t+1} = t-k+1, y_{1:t+1}).
\end{aligned}
\end{equation}
This gives us
\begin{equation}
\begin{aligned}
  \pi_B^{(t+1)}(\theta) = \sum_{k=0}^{t-1} & w^{(k)}_t \frac{ P(y_{t+1}; \pi_k^{(t)}) } { P(y_{t+1}; \pi^{(t)}) } \times\\
  & \times \textbf{P}(\Theta_{t+1} = \theta | R_{t+1} = t-k+1, Y_{1:t+1}=y_{1:t+1}),
\end{aligned}
\end{equation}
and finally
\begin{equation}
\begin{aligned}
  w^{(k)}_{B,t+1}  = \frac{ P(y_{t+1}; \pi_k^{(t)}) } { P(y_{t+1}; \pi^{(t)}) } w^{(k)}_t.
\end{aligned}
\end{equation}
Using the recursive formula, the update rule for the weights for $0 \leq k \leq t-1$ is
\begin{equation}
\begin{aligned}
  \label{Eq:SMP_w_oldparticles}
  & w^{(k)}_{t+1} = (1-\gamma_{t+1}) w^{(k)}_{B,t+1}  = (1-\gamma_{t+1}) \frac{ P(y_{t+1}; \pi_k^{(t)}) } { P(y_{t+1}; \pi^{(t)}) } w^{(k)}_t,
\end{aligned}
\end{equation}
and for the newly added particle $t$
\begin{equation}
\begin{aligned}
  \label{Eq:SMP_w_newparticle}
  & w^{(t)}_{t+1} = \gamma_{t+1},
\end{aligned}
\end{equation}
where $\gamma_{t+1} = \gamma \big(\textbf{S}_{\mathrm{BF}}(y_{t+1}; \pi^{(t)}) , m = \frac{p_c}{1-p_c} \big)$ of \autoref{Eq:Gamma_def}.

The MP$N$ algorithm uses \autoref{Eq:MPN_particle_form}, \autoref{Eq:SMP_w_oldparticles}, and \autoref{Eq:SMP_w_newparticle} for computing the belief for $t \leq N$ - which is same as the exact Bayesian inference.
For $t > N$, it first updates the weights in the same fashion as \autoref{Eq:SMP_w_oldparticles} and \autoref{Eq:SMP_w_newparticle}, keeps the greatest $N$ weights, and sets the rest weights equal to 0.
After normalizing the new weights, it uses \autoref{Eq:MPN_particle_form} (but only over the particles with non-zero weights) to compute the belief $\approxs{\pi}^{(t)}$.
For the particular case of exponential family, see Algorithm \ref{Alg:MPN} for the pseudocode.

\subsection{Derivation of the weight update for Particle Filtering  (Algo. 3)}
We derive here the weight update for the particle filter.
The difference in our formalism from a standard derivation \citep{sarkka2013bayesian} is the absence of the Markov property of conditional observations (i.e. $\textbf{P}(y_{t+1} | c_{1:t+1}, y_{1:t}) \neq \textbf{P}(y_{t+1} | c_{t+1})$).
Our goal is to perform the approximation
\begin{equation}
\label{Eq:pDh_pf_approx_method}
P(c_{1:t+1} | y_{1:t+1}) \approx \sum_{i=1}^{N} w_{t+1}^{(i)} \delta (c_{1:t+1} - c_{1:t+1}^{(i)} ) \, .
\end{equation}
Given a proposal sampling distribution $Q$, for the weight of particle $i$ at time $t+1$ we have
\begin{equation}
\begin{aligned}
w_{t+1}^{(i)} & \propto \frac{\textbf{P}(c_{1:t+1}^{(i)} | y_{1:t+1})}{Q(c_{1:t+1}^{(i)} | y_{1:t+1})}
\propto \frac{\textbf{P}(c_{1:t+1}^{(i)}, y_{t+1} | y_{1:t})}{Q(c_{1:t+1}^{(i)} | y_{1:t+1})} \\
w_{t+1}^{(i)} & \propto \frac{\textbf{P}(y_{t+1} , c_{t+1}^{(i)}| c_{1:t}^{(i)}, y_{1:t}) \textbf{P}(c_{1:t}^{(i)} | y_{1:t}) } {Q(c_{t+1}^{(i)} | c_{1:t}^{(i)}, y_{1:t+1}) Q(c_{1:t}^{(i)} | y_{1:t})}\, ,
\end{aligned}
\end{equation}
where the only assumption for the proposal distribution $Q$ is that the previous hidden states $c_{1:t}^{(i)}$ are independent of the next observation $y_{t+1}$, which allows to keep the previous samples $c_{1:t}^{(i)}$ when going from $c_{1:t}^{(i)}$ to $c_{1:t+1}^{(i)}$ and to write the update of the weights in a recursive way \citep{sarkka2013bayesian}.

Notice that $ w_{t}^{(i)} \propto \frac{\textbf{P}(c_{1:t}^{(i)} | y_{1:t})} {Q(c_{1:t}^{(i)} | y_{1:t})} $ are the weights calculated at the previous time step.
Therefore
\begin{equation}
\begin{aligned}
\label{w_pf_1}
w_{t+1}^{(i)} & \propto \frac{\textbf{P}(y_{t+1}, c_{t+1}^{(i)} | c_{1:t}^{(i)}, y_{1:t})} {Q(c_{t+1}^{(i)} | c_{1:t}^{(i)}, y_{1:t+1})} w_{t}^{(i)}\, .
\end{aligned}
\end{equation}
For the choice of $Q$, we use the optimal proposal function in terms of variance of the weights \citep{doucet2000sequential}
\begin{equation}
\begin{aligned}
  \label{Q_pf_1}
  Q(c_{t+1}^{(i)} | c_{1:t}^{(i)}, y_{1:t+1}) = \textbf{P}(c_{t+1}^{(i)} | c_{1:t}^{(i)}, y_{1:t+1})\, .
\end{aligned}
\end{equation}
Using Bayes' rule and \autoref{w_pf_1} and \autoref{Q_pf_1}, after a few steps of algebra, we have
\begin{equation}
\begin{aligned}
w_{t+1}^{(i)} &
\propto \frac{\textbf{P}(y_{t+1}, c_{t+1}^{(i)} | c_{1:t}^{(i)}, y_{1:t})} {\textbf{P}(c_{t+1}^{(i)} | c_{1:t}^{(i)}, y_{1:t+1})} w_{t}^{(i)}
= \textbf{P}(y_{t+1}| c_{1:t}^{(i)}, y_{1:t}) w_{t}^{(i)}\\
& \propto \Big( (1-p_c)\textbf{P}(y_{t+1} | c_{1:t}^{(i)}, y_{1:t}, c^{(i)}_{t+1}=0 ) + p_c \textbf{P}(y_{t+1} | c_{1:t}^{(i)}, y_{1:t}, c^{(i)}_{t+1}=1) \Big) w_{t}^{(i)}\, .
\end{aligned}
\end{equation}
Using the definition in \autoref{Eq:p_y_t}, we have $\textbf{P}(y_{t+1} | c_{1:t}^{(i)}, y_{1:t}, c^{(i)}_{t+1}=0 ) = P(y_{t+1}; \approxs{\pi}_i^{(t)})$ and $\textbf{P}(y_{t+1} | c_{1:t}^{(i)}, y_{1:t}, c^{(i)}_{t+1}=1 ) = P(y_{t+1}; \pi^{(0)})$.
Therefore, we have
\begin{equation}
\begin{aligned}
  \label{Eq:w_update_normalized}
w_{t+1}^{(i)}
& = \Big[(1-p_c) P(y_{t+1}; \approxs{\pi}_i^{(t)}) + p_c P(y_{t+1}; \pi^{(0)}) \Big] w_{t}^{(i)} / Z\, ,
\end{aligned}
\end{equation}
where $Z$ is the normalization factor
\begin{equation}
  \label{Eq:w_norm_factor}
Z = (1-p_c) P(y_{t+1}; \approxs{\pi}^{(t)}) + p_c P(y_{t+1}; \pi^{(0)})\, ,
\end{equation}
where we have
\begin{equation}
  P(y_{t+1}; \approxs{\pi}^{(t)}) = \sum_{i=1}^N w_{t}^{(i)} P(y_{t+1}; \approxs{\pi}_i^{(t)})\, .
\end{equation}
We now compute the weights corresponding to $\pi_{B}^{(t+1)}$ as defined in \autoref{Eq:pi_B}
\begin{equation}
  \label{Eq:wB_sup}
w_{B, t+1}^{(i)} = \frac{P(y_{t+1}; \approxs{\pi}_i^{(t)})}{P(y_{t+1}; \approxs{\pi}^{(t)})} w_{t}^{(i)}\, .
\end{equation}
Combining \autoref{Eq:w_update_normalized}, \autoref{Eq:w_norm_factor} and \autoref{Eq:wB_sup} we can then re-write the weight update rule as
\begin{equation}
\begin{aligned}
  \label{Eq:w_update_final}
w_{t+1}^{(i)} &= (1-\gamma_{t+1})w_{B, t+1}^{(i)} + \gamma_{t+1}w_{t}^{(i)}\, ,
\end{aligned}
\end{equation}
where $\gamma_{t+1} = \gamma \Big(\textbf{S}_{\mathrm{BF}}(y_{t+1}; \approxs{\pi}^{(t)}), \frac{p_c}{1-p_c} \Big)$ of \autoref{Eq:Gamma_def}.

At every time step $t+1$ we sample each particle's hidden state $c_{t+1}$ from the proposal distribution.
Using \autoref{Q_pf_1}, we have
\begin{equation}
\begin{aligned}
  \label{Eq:proposal_final_supp}
Q(c_{t+1}^{(i)} = 1| c_{1:t}^{(i)}, y_{1:t+1})
&= \frac{p_c P(y_{t+1}; \pi^{(0)})}{(1-p_c) P(y_{t+1}; \approxs{\pi}_i^{(t)}) + p_c P(y_{t+1}; \pi^{(0)})}\\
&= \gamma \Big( \textbf{S}_{\mathrm{BF}}(y_{t+1}; \approxs{\pi}_i^{(t)}), \frac{p_c}{1-p_c} \Big)\, .
\end{aligned}
\end{equation}

We implemented the Sequential Importance Resampling algorithm \citep{gordon1993novel,doucet2000sequential}, where
the particles are resampled when their effective number falls below a threshold.
The effective number of the particles is defined as \citep{doucet2000sequential,sarkka2013bayesian}
\begin{equation}
N_{\text{eff}} \approx \frac{1}{\sum_{i=1}^{N}(w_t^{(i)})^2}\, .
\end{equation}
When $N_{\text{eff}}$ is below a critical threshold, the particles are resampled with replacement from the categorical distribution defined by their weights, and all their weights are set to $w_t^{(i)} = 1/N$.
We did not optimize the parameter $N_{\text{eff}}$, and following \cite{doucet2009tutorial}, we performed resampling when $N_{\text{eff}} \leq N/2$.

\subsection{Surprise-modulation as a framework for other algorithms}
\label{ext_int_related}

\subsubsection{SMiLe Rule}
\label{sec:smile}

The Confidence Corrected Surprise \citep{faraji2018balancing} is
\begin{equation}
    \textbf{S}_{CC}(y_{t+1}; \approxs{\pi}^{(t)}) = \textbf{D}_{KL} \big[ \approxs{\pi}^{(t)}(\theta) || \tilde{P}(\theta|y_{t+1}) \big],
\end{equation}
where $\tilde{P}(\theta|y_{t+1})$ is the scaled likelihood defined as
\begin{equation}
  \label{Eq:scaled_like}
    \tilde{P}(\theta|y_{t+1}) = \frac{P_Y(y_{t+1}|\theta)}{\int P_Y(y_{t+1}|\theta') d \theta'}.
\end{equation}

Note that this becomes equal to $P(\theta|y_{t+1})$ if the prior belief $\pi^{(0)}$ is a uniform distribution; cf. \autoref{Eq:cond_prob_ThetaY}.

With the aim of minimizing the Confidence Corrected Surprise by updating the belief during time, \cite{faraji2018balancing} suggested an update rule solving the optimization problem
\begin{equation}
\begin{aligned}
\label{Eq:Mod_SMiLe_Min_1}
\approxs{\pi}^{(t+1)}(\theta) &= \arg \min_q \textbf{D}_{KL} \big[ q(\theta) || \tilde{P}(\theta|y_{t+1}) \big]
\\ & \text{s.t.  }  \textbf{D}_{KL} \big[ q(\theta) ||  \approxs{\pi}^{(t)}(\theta) \big] \leq B_{t+1},
\end{aligned}
\end{equation}
where $B_{t+1} \in \big[ 0, \textbf{D}_{KL}[ P(\theta|y_{t+1}) || \approxs{\pi}^{(t)}(\theta)] \big]$ is an arbitary bound.
The authors showed that the solution to this optimization problem is
\begin{equation}
\label{Eq:Mod_SMiLe_Log_Rec}
\text{log} \big( \approxs{\pi}^{(t+1)}(\theta) \big) = (1-\gamma_{t+1}) \text{ log} \big( \approxs{\pi}^{(t)}(\theta) \big) + \gamma_{t+1} \text{ log} \big( \tilde{P}(\theta|y_{t+1}) \big) + \text{Const.},
\end{equation}
where $\gamma_{t+1} \in [0,1]$ is specified so that it satisfies the constraint in \autoref{Eq:Mod_SMiLe_Min_1}.

Although \autoref{Eq:Mod_SMiLe_Log_Rec} looks very similar to \autoref{Eq:Var_SMiLe_Log_Rec_2nd_Ver}, it signifies a trade-off between the latest belief $\approxs{\pi}^{(t)}$ and the belief updated by only the most recent observation $\tilde{P}(\theta|y_{t+1})$, i.e. a trade-off between adherence to the current belief and reset.
While SMiLe adheres to the current belief $\approxs{\pi}^{(t)}$, Variational SMiLe integrates the new observation with the current belief to get $\approxs{\pi}_B^{(t)}$, which leads to a trade-off similar to the one of the exact Bayesian inference (\autoref{Eq:tradeoff} and \autoref{Eq:Bayesian_Rec_Formula}).

To modulate the learning rate by surprise, \cite{faraji2018balancing} considered the boundary $B_{t+1}$ as a function of the Confidence Corrected Surprise, i.e.
\begin{equation}
  \label{Eq:SMiLe_modulation}
\begin{aligned}
  B_{t+1} &= B_{\text{max}} \gamma \Big(\textbf{S}_{CC}(y_{t+1}) , m \Big)\\
  & \text{where } B_{\text{max}} = \textbf{D}_{KL}[ P(\theta|y_{t+1}) || \approxs{\pi}^{(t)}(\theta)]\, ,
\end{aligned}
\end{equation}
where $m$ is a free parameter.
Then, $\gamma_{t+1}$ is found by satisfying the constraint of the optimization problem in \autoref{Eq:Mod_SMiLe_Min_1} using \autoref{Eq:Mod_SMiLe_Log_Rec} and \autoref{Eq:SMiLe_modulation}.

\subsubsection{Nassar's algorithm}

For the particular case that observations are drawn from a Gaussian distribution with known variance and unknown mean, i.e. $y_{t+1}|\mu_{t+1} \sim \mathcal{N}(\mu_{t+1}, \sigma^2)$ and $\theta_t = \mu_t$, \cite{nassar2012rational, nassar2010approximately} considered the problem of estimating the expected $\mu_t$ and its variance rather than a probability distribution (i.e. belief) over it, implicitly assuming that the belief is always a Gaussian distribution.
The algorithms of \cite{nassar2012rational, nassar2010approximately} were developed for the case that, whenever the environment changes, the mean $\mu_{t+1}$ is drawn from a uniform prior with a range of values much larger than the width of the Gaussian likelihood function.
The authors showed that in this case, the expected $\mu_{t+1}$ (i.e. $\approxs{\mu}_{t+1}$) estimated by the agent upon observing a new sample $y_{t+1}$ is
\begin{equation}
  \label{Eq:nassar_original}
\approxs{\mu}_{t+1} = \approxs{\mu}_t + \alpha_{t+1} ( y_{t+1} - \approxs{\mu}_t),
\end{equation}
with $\alpha_{t+1}$ the adaptive learning rate given by
\begin{equation}
  \label{Eq:nassar_original_learningrate}
\alpha_{t+1} = \frac{1 + \Omega_{t+1}\approxs{r_t}}{1 + \approxs{r_t}},
\end{equation}
where $\approxs{r_t}$ is the estimated time since the last change point (i.e. the estimated $R_t = \min \{ n \in \mathbb{N} : C_{t-n+1} = 1$) and $\Omega_{t+1} = \textbf{P}(c_{t+1}=1|y_{1:t+1})$ the probability of a change given the observation.
Note that this quantity, i.e. the posterior change point probability, is the same as our adaptation rate $\gamma_{t+1}$ of \autoref{Eq:Gamma_def}.

In the next subsection, we extend their approach to a more general case where the prior is a Gaussian distribution with arbitrary variance, i.e. $\mu_{t+1} \sim \mathcal{N}(\mu_0, \sigma_0^2)$.
We then discuss the relation of this method to Particle Filtering.
A performance comparison between our extended algorithms Nas10$^{*}$ and Nas12$^{*}$ and their original versions Nas10 and Nas12 is depicted in Supplementary \autoref{fig:gauss_afterswitch_nassar} and Supplementary \autoref{fig:gauss_heatmaps_nassar}.

\subsubsection{Nas10$^{*}$ and Nas12$^{*}$ algorithms}
\label{sec:nassar_star}
Let us consider
that $y_{1:t}$ are observed, the time since the last change point $r_t$ is known, and the agent's current estimation of $\mu_t$ is $\approxs{\mu}_{t}$.
It can be shown (see Appendix for the derivation) that the expected $\mu_{t+1}$ (i.e. $\approxs{\mu}_{t+1}$) upon observing the new sample $y_{t+1}$ is
\begin{equation}
  \label{Eq:Nassar_post_m_4b}
  \approxs{\mu}_{t+1} = (1 - \gamma_{t+1})\Big( \approxs{\mu}_t + \frac{1}{\rho + r_t + 1} (y_{t+1} - \approxs{\mu}_t) \Big) + \gamma_{t+1} \Big( \mu_0 + \frac{1}{\rho + 1} (y_{t+1} - \mu_0) \Big)\, ,
\end{equation}
where $\rho = \frac{\sigma^2}{\sigma^2_0} $, $\mu_0$ is the mean of the prior distribution and $\gamma_{t+1}$ is the adaptation rate of \autoref{Eq:Gamma_def}.

We can see that the updated mean is a weighted average, with surprise-modulated weights, between integrating the new observation with the current mean $\approxs{\mu}_t$ and integrating it with the prior mean $\mu_0$, in the same spirit as the other algorithms we considered here.
\autoref{Eq:Nassar_post_m_4b} can also be seen as a surprise-modulated weighted sum of two delta rules: one including a prediction error between the new observation and the current mean $(y_{t+1} - \approxs{\mu}_t)$ and one including a prediction error between the observed sample and the prior mean $(y_{t+1} - \mu_0)$.

In order to obtain a form similar to the one of \cite{nassar2012rational, nassar2010approximately}, we can rewrite the above formula as
\begin{equation}
  \label{Eq:Nassar_post_m_final_2}
  \approxs{\mu}_{t+1} = \frac{\rho}{\rho + 1} \Big(\approxs{\mu}_t + \gamma_{t+1}(\mu_0 - \approxs{\mu}_t) \Big) + \frac{1}{\rho + 1} \Big( \approxs{\mu}_t + \alpha_{t+1} (y_{t+1} - \approxs{\mu}_t) \Big)\, ,
\end{equation}
where we have defined $\alpha_{t+1} = \frac{\rho + \gamma_{t+1} r_t +1}{\rho + r_t + 1}$.
Hence the update rule takes the form of a weighted average, with fixed weights, between two delta rules: one including a prediction error between the prior mean and the current mean $(\mu_0 - \approxs{\mu}_t)$ and one including a prediction error between the observed sample and the current mean $(y_{t+1} - \approxs{\mu}_t)$, both with surprise-modulated learning rates.

In \cite{nassar2012rational, nassar2010approximately} the true new mean after a change point is drawn from a uniform distribution with a range of values much larger than the width of the Gaussian likelihood.
Their derivations implicitly approximate the uniform distribution with a Gaussian distribution with $\sigma_0 \gg \sigma$.
Note that if $\sigma_0 \gg \sigma$ then $\rho \rightarrow 0$, so that the first term of \autoref{Eq:Nassar_post_m_final_2} disappears, and $\alpha_{t+1} = \frac{1 + \gamma_{t+1} r_t}{1 + r_t}$.
This results in the delta-rule of the original algorithm in \autoref{Eq:nassar_original} and \autoref{Eq:nassar_original_learningrate}, with $\gamma_{t+1}=\Omega_{t+1}$.

All of the calculations so far were done by assuming that $r_t$ is known.
However, for the case of a non-stationary regime with a history of change points, the time interval $r_t$ is not known. \cite{nassar2012rational, nassar2010approximately} used the expected time interval $\approxs{r}_t$ as an estimate.
We make a distinction here between \cite{nassar2012rational} and \cite{nassar2010approximately}:

In \cite{nassar2010approximately} $\approxs{r}_t$ is calculated recursively on each trial in the same spirit as \autoref{Eq:Bayesian_Rec_Formula}: $\approxs{r}_{t+1} = (1 - \gamma_{t+1})(\approxs{r}_t + 1) +  \gamma_{t+1}$,
i.e., at each time step, there is a probability $(1 - \gamma_{t+1})$ that $\approxs{r}_t$ increments by 1 and a probability $\gamma_{t+1}$ that it is reset to 1. So $\approxs{r}_{t+1}$ is the weighted sum of these two outcomes.
Hence, \autoref{Eq:Nassar_post_m_final_2} combined with the expected time interval $\approxs{r}_t$ constitutes a generalization of the update rule of \cite{nassar2010approximately} for the case of Gaussian prior $\mathcal{N}(\mu_0, \sigma_0^2)$.
We call this algorithm Nas10$^{*}$ (see Appendix for pseudocode).

In \cite{nassar2012rational}, the variance $\approxs{\sigma}^2_{t+1} = \text{Var}[\mu_{t+1} | y_{1:t+1}]$ is estimated given $\approxs{\mu}_t$, $\approxs{r}_t$, and $\approxs{\sigma}^2_{t}$.
Based on this variance, $\approxs{r}_{t+1} = \frac{\sigma^2}{\approxs{\sigma}^2_{t+1}} - \frac{\sigma^2}{\sigma^2_{0}}$ is computed.
The derivation of the recursive computation of $\approxs{\sigma}^2_{t+1}$ for the case of Gaussian priors can be found in the Appendix.
We call the combination of \autoref{Eq:Nassar_post_m_final_2} with this way of computing the expected time interval $\approxs{r}_t$ Nas12$^{*}$ (see Appendix for pseudocode).
These two versions of calculating $\approxs{r}_t$ in \cite{nassar2010approximately} and \cite{nassar2012rational} give different results, and we compare our algorithms with both Nas10$^{*}$ and Nas12$^{*}$ in our simulations.
Note that, as discussed in the section ``Online Bayesian inference modulated by surprise'' of the Results, the posterior belief at time $t+1$ does not generally belong to the same family of distributions as the belief of time $t$. However, we therefore approximate for both algorithms the posterior belief $P(\theta | y_{1:t+1})$ by a Gaussian.

\subsubsection{Nassar's algorithm and Particle Filtering with one particle}
\label{sec:nassar_pf1}
In the case of Particle Filtering (cf. \autoref{Eq:proposal_final}) with only one particle, at each time step we sample the particle's hidden state with change probability $Q(c_{t+1}^{(1)} = 1| c_{1:t}^{(1)}, y_{1:t+1}) = \gamma_{t+1}$,
generating a posterior belief that takes two possible values with probability (according to the proposal distribution)
\begin{equation}
  \begin{aligned}
  \label{Eq:pf1_postbelief}
  Q \Big( \approxs{\pi}^{(t+1)}(\theta) = \approxs{\pi}^{(t+1)}_B(\theta) | \ y_{t+1} \Big) &= 1 - \gamma_{t+1},\\
  Q \Big( \approxs{\pi}^{(t+1)}(\theta) = P(\theta|y_{t+1}) | \ y_{t+1} \Big) &= \gamma_{t+1}.
\end{aligned}
\end{equation}
So, \textit{in expectation}, the updated belief will be
\begin{equation}
  \begin{aligned}
  \label{Eq:pf1_expectedpostbelief}
  \mathbb{E}_{Q}[\approxs{\pi}^{(t+1)}(\theta)] = (1 - \gamma_{t+1})\approxs{\pi}^{(t+1)}_B(\theta) + \gamma_{t+1} P(\theta|y_{t+1}).
\end{aligned}
\end{equation}
If we apply \autoref{Eq:pf1_expectedpostbelief} to $\approxs{\mu}_{t+1}$, we find that $\mathbb{E}_{Q}[\approxs{\mu}^{(t+1)}] $, is identical to the generalization of \cite{nassar2010approximately} (see \autoref{Eq:Nassar_post_m_4b}).

Moreover, in Particle Filtering with a single particle, we sample the particle's hidden state, which is equivalent to sampling the interval $\approxs{R}_{t+1}$.
Because $\approxs{R}_{t+1}$ takes the value $\approxs{r}_t + 1$ with $(1 - \gamma_{t+1})$ and the value of $1$ (=reset) with probability $\gamma_{t+1}$,
the \textit{expected value} of $\approxs{R}_{t+1}$ is
\begin{equation}
  \label{Eq:pf1_expectedrt+1}
  \mathbb{E}_{Q}[\approxs{R}_{t+1}] = (1 - \gamma_{t+1})(\approxs{r}_t + 1) + \gamma_{t+1}.
\end{equation}
In other words, in \cite{nassar2010approximately}, the belief is updated based on the \textit{expected} $\approxs{r}_t$, whereas in Particle Filtering with one particle, the belief is updated using the \textit{sampled} $\approxs{r}_t$.

In summary, the two methods will give different estimates on a trial-per-trial basis, but the same result in expectation.
The pseudocode for Particle Filtering with one particle for the particular case of the Gaussian estimation task can be found in the Appendix.

\subsection{Application to the exponential family}
\label{expfam}

For our all three algorithms Variational SMiLe, Message Passing with fixed number $N$ of particles, and Particle Filtering, we derive compact update rules for $\approxs{\pi}^{(t+1)}(\theta)$ when the likelihood function $P_Y(y|\theta)$ is in the exponential family and $\pi^{(0)}(\theta)$ is its conjugate prior.
In that case, the likelihood function has the form
\begin{equation}
    \label{Eq:exp_fam_likelihood}
    P_Y(y|\theta) = h(y)\text{exp}\big(\theta^T \phi(y) - A(\theta)\big),
\end{equation}
where $\theta$ is the vector of natural parameters, $h(y)$ is a positive function, $\phi(y)$ is the vector of sufficient statistics, and $A(\theta)$ is the normalization factor.
Then, the conjugate prior $\pi^{(0)}$ has the form
\begin{equation}
\label{Eq:exp_fam_conj_prior}
  \begin{aligned}
    \pi^{(0)}(\theta) &= \textbf{P}_{\pi}\big(\Theta = \theta; \chi^{(0)}, \nu^{(0)}\big)\\
    &= \tilde{h}(\theta) f\big(\chi^{(0)}, \nu^{(0)}\big) \text{exp}\big(\theta^T \chi^{(0)} - \nu^{(0)} A(\theta)\big)
  \end{aligned}
\end{equation}
where $\chi^{(0)}$ and $\nu^{(0)}$ are the distribution parameters, $\tilde{h}(\theta)$ is a positive function, and $f\big(\chi^{(0)}, \nu^{(0)}\big)$ is the normalization factor.
For this setting and while $\pi^{(t)}=\textbf{P}_{\pi}\big(\Theta = \theta; \chi^{(t)}, \nu^{(t)}\big)$, the ``Bayes Factor Surprise'' has the compact form
\begin{equation}
\label{Eq:S_GM_exp}
\textbf{S}_{\mathrm{BF}} \Big( y_{t+1};\textbf{P}_{\pi}\big(\Theta = \theta; \chi^{(t)}, \nu^{(t)}\big) \Big) =
\frac{f\big(\chi^{(t)}+\phi(y_{t+1}), \nu^{(t)}+1\big)}{f\big(\chi^{(0)}+\phi(y_{t+1}), \nu^{(0)}+1\big)}
\frac{f\big(\chi^{(0)}, \nu^{(0)}\big)}{f\big(\chi^{(t)}, \nu^{(t)}\big)}\, .
\end{equation}

The pseudocode for Variational SMiLe, MP$N$, and Particle Filtering can be seen in Algorithms \ref{Alg:Var_SMiLe}, \ref{Alg:MPN}, and \ref{Alg:PF}, respectively.

\subsection{Simulation task}

In this subsection, we first argue why the mean squared error is a proper measure for comparing different algorithms with each other, and then we explain the version of Leaky integrator which we used for simulations.

\subsubsection{Mean squared error as an optimality measure}

Consider the case that at each time point $t$, the goal of an agent is to have an estimation of the parameter $\Theta_t$ as a function of the observations $Y_{1:t}$, i.e. $\approxs{\Theta}_t = f(Y_{1:t})$.
The estimator which minimizes the mean squared error $\textbf{MSE} [ \approxs{\Theta}_t ] = \mathbb{E}_{\textbf{P}(Y_{1:t},\Theta_t)} \big[ ( \approxs{\Theta}_t - \Theta_t)^2 \big]$ is
\begin{equation}
  \approxs{\Theta}_t^{\text{Opt}} = \mathbb{E}_{\textbf{P}(\Theta_{t}|Y_{1:t} )} \big[ \Theta_t \big] = \mathbb{E}_{\pi^{(t)}} \big[ \Theta_t \big],
\end{equation}
which is the expected value of $\Theta_t$ conditioned on the observations $Y_{1:t}$, or in other words under the Bayes-optimal current belief (see \cite{papoulis1989probability} for a proof).
The MSE for any other estimator $\approxs{\Theta}_t$ can be written as (see below for the proof)
\begin{equation}
\begin{aligned}
  \label{eq:MSE_dMSE}
  &\textbf{MSE} [ \approxs{\Theta}_t ] = \textbf{MSE} [ \approxs{\Theta}_t^{\text{Opt}} ] + \Delta \textbf{MSE} [ \approxs{\Theta}_t ],\\
  & \text{ where } \Delta \textbf{MSE} [ \approxs{\Theta}_t ] = \mathbb{E}_{\textbf{P}(Y_{1:t})} \Big[ (\approxs{\Theta}_t - \approxs{\Theta}^{\text{Opt}}_t)^2  \Big] \geq 0.
\end{aligned}
\end{equation}
This means that the MSE for any arbitrary estimator $\approxs{\Theta}_t$ includes two terms: the optimal MSE and the mismatch of the actual estimator from the optimal estimator $\approxs{\Theta}_t^{\text{Opt}}$.
As a result, if the estimator we are interested in is the expected value of $\Theta_t$ under the approximate belief $\approxs{\pi}^{(t)}$ computed by each of our algorithms (i.e. $\approxs{\Theta}_t' = \mathbb{E}_{\approxs{\pi}^{(t)}} \big[ \Theta_t \big]$), the second term in \autoref{eq:MSE_dMSE}, i.e. the deviation from optimality, is a measure of how good the approximation is.

\textbf{Proof for the algorithms without sampling:}

Consider $\approxs{\Theta}^{\text{Opt}}_t = f_{\text{Opt}}(Y_{1:t})$. Then, for any other arbitrary estimator $\approxs{\Theta}_t = f(Y_{1:t})$ (except for the ones with sampling), we have
\begin{equation}
\begin{aligned}
    \textbf{MSE} [ \approxs{\Theta}_t ] &= \mathbb{E}_{\textbf{P}(Y_{1:t},\Theta_t)} \big[ ( \approxs{\Theta}_t - \Theta_t)^2 \big]\\
    &= \mathbb{E}_{\textbf{P}(Y_{1:t},\Theta_t)} \big[ ( f(Y_{1:t}) - \Theta_t)^2 \big]\\
    &= \mathbb{E}_{\textbf{P}(Y_{1:t},\Theta_t)} \Big[ \big( (f(Y_{1:t}) - f_{\text{Opt}}(Y_{1:t})) + (f_{\text{Opt}}(Y_{1:t}) - \Theta_t) \big)^2 \Big].
\end{aligned}
\end{equation}
The quadratic term in the last line can be expanded and written as
\begin{equation}
\begin{aligned}
    \textbf{MSE} [ \approxs{\Theta}_t ] =  &\mathbb{E}_{\textbf{P}(Y_{1:t},\Theta_t)} \Big[ (f(Y_{1:t}) - f_{\text{Opt}}(Y_{1:t}))^2  \Big] +\\
    & \mathbb{E}_{\textbf{P}(Y_{1:t},\Theta_t)} \Big[ (f_{\text{Opt}}(Y_{1:t}) - \Theta_t)^2  \Big]+\\
    & 2\mathbb{E}_{\textbf{P}(Y_{1:t},\Theta_t)} \Big[ (f(Y_{1:t}) - f_{\text{Opt}}(Y_{1:t}))(f_{\text{Opt}}(Y_{1:t}) - \Theta_t)  \Big].
\end{aligned}
\end{equation}
The random variables in the expected value of the first line are not dependent on $\Theta_t$, so it can be computed over $Y_{1:t}$. The expected value of the second line is equal to $\textbf{MSE} [ \approxs{\Theta}^{\text{Opt}}_t ]$.
It can also be shown that the expected value of the third line is equal to 0, i.e.
\begin{equation}
\begin{aligned}
    \text{3rd line } &= 2 \mathbb{E}_{\textbf{P}(Y_{1:t})} \Big[ \mathbb{E}_{\textbf{P}(\Theta_t|Y_{1:t})} \Big[(f(Y_{1:t}) - f_{\text{Opt}}(Y_{1:t}))(f_{\text{Opt}}(Y_{1:t}) - \Theta_t)  \Big] \Big]\\
    &= 2 \mathbb{E}_{\textbf{P}(Y_{1:t})} \Big[ (f(Y_{1:t}) - f_{\text{Opt}}(Y_{1:t}))(f_{\text{Opt}}(Y_{1:t}) - \mathbb{E}_{\textbf{P}(\Theta_t|Y_{1:t})} [\Theta_t]) \Big]
= 0,
\end{aligned}
\end{equation}
where in the last line we used the definition of the optimal estimator.
All together, we have
\begin{equation}
\begin{aligned}
    \textbf{MSE} [ \approxs{\Theta}_t ] = \textbf{MSE} [ \approxs{\Theta}^{\text{Opt}}_t ] + \mathbb{E}_{\textbf{P}(Y_{1:t})} \Big[ (\approxs{\Theta}_t - \approxs{\Theta}^{\text{Opt}}_t)^2  \Big].
\end{aligned}
\end{equation}

\textbf{Proof for the algorithms with sampling:}

For particle filtering (and any kind of estimator with sampling), the estimator is not a deterministic function of observations $Y_{1:t}$.
Rather, the estimator is a function of observations as well as a set of random variables (samples) which are drawn from a distribution which is also a function of observations $Y_{1:t}$.
In our case, the samples are the sequence of hidden states $C_{1:t}$.
The estimator can be written as
\begin{equation}
    \approxs{\Theta}^{\text{PF}}_t = f(Y_{1:t}, C_{1:t}^{(1:N)}),
\end{equation}
where $C_{1:t}^{(1:N)}$ are $N$ iid samples drawn from the proposal distribution $Q(C_{1:t}|Y_{1:t})$.
MSE for this estimator should also be averaged over the samples, which leads to
\begin{equation}
\begin{aligned}
    \textbf{MSE} [ \approxs{\Theta}^{\text{PF}}_t ] &= \mathbb{E}_{\textbf{P}(Y_{1:t},\Theta_t)Q(C_{1:t}^{(1:N)}|Y_{1:t})} \big[ ( \approxs{\Theta}^{\text{PF}}_t - \Theta_t)^2 \big]\\
    &= \mathbb{E}_{\textbf{P}(Y_{1:t},\Theta_t)Q(C_{1:t}^{(1:N)}|Y_{1:t})} \big[ ( f(Y_{1:t}, C_{1:t}^{(1:N)}) - \Theta_t)^2 \big].
\end{aligned}
\end{equation}
Similar to what we did before, the MSE for particle filtering can be written as
\begin{equation}
\begin{aligned}
    \textbf{MSE} [ \approxs{\Theta}^{\text{PF}}_t ] &= \textbf{MSE} [ \approxs{\Theta}^{\text{Opt}}_t ] + \mathbb{E}_{\textbf{P}(Y_{1:t})Q(C_{1:t}^{(1:N)}|Y_{1:t})} \Big[ (\approxs{\Theta}^{\text{PF}}_t - \approxs{\Theta}^{\text{Opt}}_t)^2  \Big]\\
    &= \textbf{MSE} [ \approxs{\Theta}^{\text{Opt}}_t ] + \mathbb{E}_{\textbf{P}(Y_{1:t})} \Big[  \mathbb{E}_{Q(C_{1:t}^{(1:N)}|Y_{1:t})} \big[ (\approxs{\Theta}^{\text{PF}}_t - \approxs{\Theta}^{\text{Opt}}_t)^2  \big] \Big]
\end{aligned}
\end{equation}
which can be written in terms of bias and variance over samples as
\begin{equation}
\begin{aligned}
    \textbf{MSE} [ &\approxs{\Theta}^{\text{PF}}_t ] = \textbf{MSE} [ \approxs{\Theta}^{\text{Opt}}_t ]\\ &+ \mathbb{E}_{\textbf{P}(Y_{1:t})} \Big[ \text{Var}_{Q(C_{1:t}^{(1:N)}|Y_{1:t})}(\approxs{\Theta}^{\text{PF}}_t) +  \text{Bias}_{Q(C_{1:t}^{(1:N)}|Y_{1:t})}(\approxs{\Theta}^{\text{PF}}_t,\approxs{\Theta}^{\text{Opt}}_t)^2   \Big].
\end{aligned}
\end{equation}

\subsubsection{Leaky integration}

Gaussian task:
The goal is to have an estimation of the mean of the Gaussian distribution at each time $t$, denoted by $\approxs{\theta}_t$.
Given a leak parameter $\omega \in (0,1]$, the leaky integrator estimation is
\begin{equation}
\begin{aligned}
    \approxs{\theta}_t = \frac{\sum_{k=1}^{t} \omega^{t-k} y_k }{ \sum_{k=1}^{t} \omega^{t-k} }.
\end{aligned}
\end{equation}

Categorical task:
The goal is to have an estimation of the parameters of the categorical distribution at each time $t$, denoted by $\approxs{\theta}_t = [\approxs{\theta}_{i,t}]_{i=1}^{N}$ for the case that there are $N$ categories.
Given a leak parameter $\omega \in (0,1]$, the leaky integrator estimation is
\begin{equation}
\begin{aligned}
    \approxs{\theta}_{i,t} = \frac{\sum_{k=1}^{t} \omega^{t-k} \delta(y_k - i) }{ \sum_{k=1}^{t} \omega^{t-k} },
\end{aligned}
\end{equation}
where $\delta$ is the Kronecker delta function.

\subsection{Derivation of the Formula Relating Shannon Surprise to the Modulated Learning Rate}

Given the defined generative model, the Shannon surprise upon observing $y_{t+1}$ can be written as
\begin{equation} \label{Eq:S_Shannon}
\begin{aligned}
  \textbf{S}_{\mathrm{Sh}}(y_{t+1}; \pi^{(t)}) &= \text{log} \Big( \frac{1}{\textbf{P}(y_{t+1}|y_{1:t})} \Big)\\
  &= \text{log} \Big( \frac{1}{(1-p_c) P(y_{t+1}; \pi^{(t)}) + p_c P(y_{t+1}; \pi^{(0)})} \Big)\\
  &= \text{log} \Big( \frac{1}{P(y_{t+1}; \pi^{(0)})} \Big) + \text{log} \Big( \frac{1}{ p_c } \frac{1}{1 + \frac{1}{m} \frac{1}{\textbf{S}_{\mathrm{BF}}(y_{t+1}; \pi^{(t)})}} \Big)\\
  &= \textbf{S}_{\mathrm{Sh}}(y_{t+1}; \pi^{(0)}) + \text{log} \Big( \frac{\gamma_{t+1}}{ p_c }  \Big),
\end{aligned}
\end{equation}
where $\gamma_{t+1} = \gamma \Big(\textbf{S}_{\mathrm{BF}}(y_{t+1}; \pi^{(t)}) , m = \frac{p_c}{1-p_c} \Big) $ of \autoref{Eq:Gamma_def}.
As a result, the modulated adaptation rate can be written as in \autoref{gamma_Shannon} and the Bayes Factor Surprise as in \autoref{Sgm_Shannon}.

\subsection{Experimental predictions}

\subsubsection{Setting}
Consider a Gaussian task where $Y_t$ can take values in $\mathbb{R}$.
The likelihood function $P_Y(y|\theta)$ is defined as
\begin{equation}
\begin{aligned}
P_Y(y|\theta) = \mathcal{N}(y; \theta, \sigma^2),
\end{aligned}
\end{equation}
where $\sigma \in \mathbb{R}^{+}$ is the standard deviation, and $\theta \in \mathbb{R}$ is the mean of the distribution, i.e. the parameter of the likelihood.
Whenever there is a change in the environment (with probability $p_c \in (0,1)$), the value $\theta$ is drawn from the prior distribution $\pi^{(0)}(\theta) =  \mathcal{N}(\theta; 0, 1)$.

\subsubsection{Theoretical proofs for prediction 1}
For our theoretical derivations for our first prediction, we consider the specific but relatively mild assumption that the subjects' belief $\pi^{(t)}$ at each time is a Gaussian distribution
\begin{equation}
\begin{aligned}
\pi^{(t)}(\theta) =  \mathcal{N}(\theta; \approxs{\theta}_t, \approxs{\sigma}_t^2),
\end{aligned}
\end{equation}
where $\approxs{\theta}_t$ and $\approxs{\sigma}_t$ are determined by the learning algorithm and the sequence of observations $y_{1:t}$.
This is the case when the subjects use either VarSMiLe, Nas10$^{*}$, Nas12$^{*}$, pf1, MP1, or Leaky Integration as their learning rule.
With such assumptions, the inferred probability distribution $P(y;\pi^{(t)})$ can be written as
\begin{equation}
\begin{aligned}
P(y;\pi^{(t)}) = \mathcal{N}(y ; \approxs{\theta}_t, \sigma^2 + \approxs{\sigma}_t^2).
\end{aligned}
\end{equation}

As mentioned in the Results section, we define, at time $t$, the prediction error as $\delta_{t+1} = y_{t+1} - \approxs{\theta}_{t}$ and the ``sign bias'' as $s_{t+1} = \text{sign}(\delta_{t+1} \approxs{\theta}_{t})$.
Then, given an absolute prediction $\approxs{\theta}>0$, an absolute prediction error $\delta>0$, a standard deviation $\sigma_C$, and a sign bias $s \in \{ -1, 1 \}$, the average Bayes Factor Surprise is computed as
\begin{equation}
\begin{aligned}
\Bar{\textbf{S}}_{\mathrm{BF}}&(\approxs{\theta}, \delta,s,\sigma_C) = \frac{1}{|\mathcal{T}|} \sum_{t \in \mathcal{T}} \textbf{S}_{\mathrm{BF}}(y_{t}; \approxs{\pi}^{(t-1)}),\\
& \text{where } \mathcal{T} = \{ t: |\approxs{\theta}_{t-1}| = \approxs{\theta}, |\delta_t| = \delta, \approxs{\sigma}_t = \sigma_C, s_t = s \}.
\end{aligned}
\end{equation}
It can easily be shown that the value $\textbf{S}_{\mathrm{BF}}(y_{t}; \approxs{\pi}^{(t-1)})$ is same for all $t \in \mathcal{T}$, and hence the average surprise is same as the surprise for each time point.
For example, the average surprise for $s=+1$ is equal to
\begin{equation}
\begin{aligned}
\Bar{\textbf{S}}_{\mathrm{BF}}(\approxs{\theta},\delta,s=+1,\sigma_C) =
\frac{\mathcal{N}(\approxs{\theta}+ \delta ; 0, \sigma^2 + 1)}
{\mathcal{N}(\delta ; 0, \sigma^2 + \sigma_C^2)}.
\end{aligned}
\end{equation}
Similar formulas can be computed for $\Bar{\textbf{S}}_{\mathrm{BF}}(\approxs{\theta}, \delta,s=-1,\sigma_C)$.
Then, the difference $\Delta \Bar{\textbf{S}}_{\mathrm{BF}}(\approxs{\theta},\delta,\sigma_C) = \Bar{\textbf{S}}_{\mathrm{BF}}(\approxs{\theta},\delta,s=+1,\sigma_C) - \Bar{\textbf{S}}_{\mathrm{BF}}(\approxs{\theta},\delta,s=-1,\sigma_C)$ can be computed as
\begin{equation}
\begin{aligned}
\Delta \Bar{\textbf{S}}_{\mathrm{BF}}(\approxs{\theta},\delta,\sigma_C) = \frac{\mathcal{N}(\approxs{\theta}+ \delta ; 0, \sigma^2 + 1) - \mathcal{N}(\approxs{\theta} - \delta ; 0, \sigma^2 + 1)}{\mathcal{N}(\delta ; 0, \sigma^2 + \sigma_C^2)}.
\end{aligned}
\end{equation}
It can be shown that
\begin{equation}
\begin{aligned}
\Delta \Bar{\textbf{S}}_{\mathrm{BF}}(\approxs{\theta},\delta,\sigma_C) < 0 \text{ and } \frac{\partial}{\partial \delta} \Delta \Bar{\textbf{S}}_{\mathrm{BF}}(\approxs{\theta},\delta,\sigma_C) < 0,
\end{aligned}
\end{equation}
for all $\approxs{\theta}>0$, $\delta>0$, and $\sigma_C>0$.
The first inequality is trivial, and the proof for the second inequality is given below.

The average Shannon Surprise can be computed in a similar way.
For example, for $s=+1$, we have
\begin{equation}
\begin{aligned}
\Bar{\textbf{S}}_{Sh}(\approxs{\theta},\delta,s=+1,\sigma_C) &= - \log \Big(
p_c \mathcal{N}(\approxs{\theta} + \delta ; 0, \sigma^2 + 1)
+ (1-p_c) \mathcal{N}(\delta ; 0, \sigma^2 + \sigma_C^2) \Big),
\end{aligned}
\end{equation}
and then the difference $\Delta \Bar{\textbf{S}}_{Sh}(\approxs{\theta},\delta,\sigma_C) = \Bar{\textbf{S}}_{Sh}(\approxs{\theta},\delta,s=+1,\sigma_C) - \Bar{\textbf{S}}_{Sh}(\approxs{\theta},\delta,s=-1,\sigma_C)$ can be computed as
\begin{equation}
\begin{aligned}
\label{Eq:Shannon_diff_experiment}
\Delta \Bar{\textbf{S}}_{Sh}(\approxs{\theta},\delta,\sigma_C) = \log \Big(  \frac{1 + m \Bar{\textbf{S}}_{\mathrm{BF}}(\approxs{\theta},\delta,s=-1,\sigma_C)}{1 + m \Bar{\textbf{S}}_{\mathrm{BF}}(\approxs{\theta},\delta,s=+1,\sigma_C)} \Big),
\end{aligned}
\end{equation}
where $m=\frac{p_c}{1-p_c}$.
Then, using the results for the Bayes Factor Surprise, we have
\begin{equation}
\begin{aligned}
\Delta \Bar{\textbf{S}}_{Sh}(\approxs{\theta},\delta,\sigma_C) > 0 \text{ and } \frac{\partial}{\partial \delta} \Delta \Bar{\textbf{S}}_{Sh}(\approxs{\theta},\delta,\sigma_C) > 0,
\end{aligned}
\end{equation}
for all $\approxs{\theta}>0$, $\delta>0$, and $\sigma_C>0$.
See below for the proof of the second inequality.

\paragraph{Proof of the 2nd inequality for $\textbf{S}_{\mathrm{BF}}$:}
Let us define the variables
\begin{equation}
\begin{aligned}
\sigma_d^2 &= \sigma^2 + \sigma_C^2
\quad \sigma_n^2 = \sigma^2 + 1,\\
\end{aligned}
\end{equation}
as well as the functions
\begin{equation}
\begin{aligned}
f_1(\delta) &= \Bar{\textbf{S}}_{\mathrm{BF}}(\approxs{\theta},\delta,s=+1,\sigma_C) = \frac{\sigma_d}{\sigma_n} \text{exp}\big( \frac{\delta^2}{2\sigma_d^2} - \frac{(\delta + \approxs{\theta})^2}{2\sigma_n^2} \big)\\
f_2(\delta) &= \Bar{\textbf{S}}_{\mathrm{BF}}(\approxs{\theta},\delta,s=-1,\sigma_C) = \frac{\sigma_d}{\sigma_n} \text{exp}\big( \frac{\delta^2}{2\sigma_d^2} - \frac{(\delta - \approxs{\theta})^2}{2\sigma_n^2} \big)\\
f(\delta) &= \Delta \Bar{\textbf{S}}_{\mathrm{BF}}(\approxs{\theta},\delta,\sigma_C) = f_1(\delta) - f_2(\delta).
\end{aligned}
\end{equation}
The following inequalities hold true
\begin{equation}
\begin{aligned}
f(\delta) <0  & \Rightarrow f_1(\delta) < f_2(\delta)\\
\sigma_C^2 < \sigma^2_0 = 1  & \Rightarrow \sigma_d^2 < \sigma_n^2.
\end{aligned}
\end{equation}
Then, the derivative of $f(\delta)$ can be compute as
\begin{equation}
\begin{aligned}
\frac{d}{d\delta}f(\delta) &= f_1(\delta) \big( \frac{\delta}{\sigma_d^2} - \frac{\delta + \approxs{\theta}}{\sigma_n^2} \big) - f_2(\delta) \big( \frac{\delta}{\sigma_d^2} - \frac{\delta - \approxs{\theta}}{\sigma_n^2} \big) \\
&= - \frac{\approxs{\theta}}{\sigma_n^2} \big( f_1(\delta) + f_2(\delta) \big) \Big( 1 - \frac{f_1(\delta) - f_2(\delta)}{f_1(\delta) + f_2(\delta)} \frac{\delta}{\approxs{\theta}} \big( \frac{\sigma_n^2}{\sigma_d^2} -1 \big) \Big)\\
&< 0.
\end{aligned}
\end{equation}
Therefore we have $\frac{\partial}{\partial \delta} \Delta \Bar{\textbf{S}}_{\mathrm{BF}}(\approxs{\theta},\delta,\sigma_C) = \frac{d}{d\delta}f(\delta) <0$.

\paragraph{Proof of the 2nd inequality for $\textbf{S}_{\mathrm{Sh}}$:}
Using the functions we defined for the previous proof, and after computing the partial derivative of \autoref{Eq:Shannon_diff_experiment}, we have
\begin{equation}\
\begin{aligned}
\frac{\partial}{\partial \delta} \Delta\Bar{\textbf{S}}_{\mathrm{Sh}}(\approxs{\theta},\delta,\sigma_C) &= \frac{\partial}{\partial \delta} \log \Big(  \frac{1 + m \Bar{\textbf{S}}_{\mathrm{BF}}(\approxs{\theta},\delta,s=-1,\sigma_C)}{1 + m \Bar{\textbf{S}}_{\mathrm{BF}}(\approxs{\theta},\delta,s=+1,\sigma_C)} \Big)\\
&= \frac{d}{d \delta} \log \Big(  \frac{1 + m f_2 (\delta)}{1 + m f_1 (\delta)} \Big).
\end{aligned}
\end{equation}
The derivative of the last term can be written in terms of the derivates of $f_1$ and $f_2$, indicated by $f_1'$ and $f_2'$, respectively,
\begin{equation}\
\begin{aligned}
\frac{\partial}{\partial \delta} \Delta\Bar{\textbf{S}}_{\mathrm{Sh}}(\approxs{\theta},\delta,\sigma_C)
&= \frac{m f_2' (\delta)}{1 + m f_2 (\delta)}-\frac{m f_1' (\delta)}{1 + m f_1 (\delta)}\\
&=
\frac{ - m f' (\delta)}{\big( 1 + m f_1 (\delta) \big) \big( 1 + m f_2 (\delta) \big)} +
\frac{ m^2 (f_1 f_2'-f_1' f_2) (\delta)}{\big( 1 + m f_1 (\delta) \big) \big( 1 + m f_2 (\delta) \big)}.
\end{aligned}
\end{equation}
The 1st term is always positive based on the proof for $\textbf{S}_{\mathrm{BF}}$.
The 2nd term is also always positive, because
\begin{equation}\
\begin{aligned}
(f_1 f_2'-f_1' f_2) (\delta) &= f_1(\delta)f_2(\delta) \Big( \big( \frac{\delta}{\sigma_d^2} - \frac{\delta - \approxs{\theta}}{\sigma_n^2} \big) - \big( \frac{\delta}{\sigma_d^2} - \frac{\delta + \approxs{\theta}}{\sigma_n^2} \big) \Big) \\
&= f_1(\delta)f_2(\delta) \frac{2 \approxs{\theta}}{\sigma_n^2} > 0.
\end{aligned}
\end{equation}
As a result, we have $\frac{\partial}{\partial \delta} \Delta\Bar{\textbf{S}}_{\mathrm{Sh}}(\approxs{\theta},\delta,\sigma_C)>0$.

\subsubsection{Simulation procedure for prediction 1}

In order to relax the main assumption of our theoretical proofs (i.e. the belief is always a Gaussian distribution), to include the practical difficulties of a real experiment (e.g. to use $|\delta_t| \approx \delta$ instead of $|\delta_t| = \delta$), and to have an estimation of the effect size, we also performed simulations for our first experimental prediction.

For each simulated subject, the procedure of our simulation was as follows:
\begin{enumerate}
  \item We fixed the hyper parameters $\sigma^2$ and $p_c$ for producing samples.
  \item We selected a learning algorithm (e.g. pf20) and fixed its corresponding tuned parameters (based on our simulations in the Results section).
  \item We applied the learning algorithm over a sequence of observations $y_{1:T}$.
  Note that in a real experiment, this step can be done through a few episodes, which makes it possible to have a long sequence of observations, i.e. large $T$.
  \item At each time $t$, we saved the values $y_t$, $\approxs{\theta}_t$, $\approxs{\sigma}_t$, $\delta_t$, $s_t$, $\textbf{S}_{\mathrm{Sh}}(y_{t}; \approxs{\pi}^{(t-1)})$, and $\textbf{S}_{\mathrm{BF}}(y_{t}; \approxs{\pi}^{(t-1)})$.
\end{enumerate}

Then, given an absolute prediction $\approxs{\theta}>0$, an absolute prediction error $\delta>0$, a standard deviation $\sigma_C>0$, and a sign bias $s \in \{ -1, 1 \}$, we defined the set of time points
\begin{equation}\
\begin{aligned}
\mathcal{T} = \{ 1 < t \leq T: | |\approxs{\theta}_{t-1}| - \approxs{\theta}| < \Delta \theta, | |\delta_t| - \delta| < \Delta \delta, |\approxs{\sigma}_t - \sigma_C| < \Delta \sigma_C, s_t = s \},
\end{aligned}
\end{equation}
where
$| |\approxs{\theta}_{t-1}| - \approxs{\theta}| < \Delta \theta$, $| |\delta_t| - \delta| < \Delta \delta$, and $|\approxs{\sigma}_t - \sigma_C| < \Delta \sigma_C$ are equivalent to
$|\approxs{\theta}_{t-1}| \approx \approxs{\theta}$, $|\delta_t| \approx \delta$, and $\approxs{\sigma}_t \approx \sigma_C$, respectively.
$\Delta \theta$, $\Delta \delta$, and $\Delta \sigma_C$ are positive real values that should be determined based on practical limitations (mainly the length of the observation sequence $T$).
We then computed the average surprise values as
\begin{equation}\
\begin{aligned}
\Bar{\textbf{S}}_{\mathrm{BF}}(\approxs{\theta}, \delta,s,\sigma_C) &= \frac{1}{|\mathcal{T}|} \sum_{t \in \mathcal{T}} \textbf{S}_{\mathrm{BF}}(y_{t}; \approxs{\pi}^{(t-1)})\\
\Bar{\textbf{S}}_{Sh}(\approxs{\theta}, \delta,s,\sigma_C) &= \frac{1}{|\mathcal{T}|} \sum_{t \in \mathcal{T}} \textbf{S}_{\mathrm{Sh}}(y_{t}; \approxs{\pi}^{(t-1)}).
\end{aligned}
\end{equation}
We repeated this procedure for $N$ different simulated subjects (with different random seeds).
The average of $\Bar{\textbf{S}}_{\mathrm{BF}}$ and $\Bar{\textbf{S}}_{Sh}$ over $N=20$ subjects, for two learning algorithms (i.e. Nas12$^{*}$ and pf20), and for $T=500$, $\approxs{\theta} = 1$, $\sigma_C = 0.5$, $\Delta \theta = 0.25$, $\Delta \delta = 0.1$, and $\Delta \sigma_C = 1$ is shown in \autoref{fig:experimentalpredictionfirst}B.
The results are the same as what was predicted by our theoretical analysis.

\subsubsection{Simulation procedure for prediction 2}

For our second prediction, the theoretical proof is trivial.
However, in order to have a setting similar to a real experiment (e.g. to use $P(y_{t+1};\approxs{\pi}^{(t)}) \approx p$ instead of $P(y_{t+1};\approxs{\pi}^{(t)}) = p$), and to have an estimation of the effect size, we used simulations also for our second experimental predictions.

We followed the same procedure as the one for the simulation of the first prediction.
For each simulated subject, and at each time $t$, we saved the quantities $P(y_{t+1};\approxs{\pi}^{(t)})$, $P(y_{t+1};\approxs{\pi}^{(0)})$, $\textbf{S}_{\mathrm{Sh}}(y_{t}; \approxs{\pi}^{(t-1)})$, and $\textbf{S}_{\mathrm{BF}}(y_{t}; \approxs{\pi}^{(t-1)})$.
Then, for a given a probability value $p>0$, we defined the set of time points
\begin{equation}\
\begin{aligned}
\mathcal{T} = \{ 0 \leq t \leq T: | P(y_{t+1};\approxs{\pi}^{(t)}) - p| < \Delta p, | P(y_{t+1};\approxs{\pi}^{(0)}) - p| < \Delta p \},
\end{aligned}
\end{equation}
where
$| P(y_{t+1};\approxs{\pi}^{(t)}) - p| < \Delta p$ and $| P(y_{t+1};\approxs{\pi}^{(0)}) - p| < \Delta p$ are equivalent to
$P(y_{t+1};\approxs{\pi}^{(t)}) \approx p$ and $P(y_{t+1};\approxs{\pi}^{(0)}) \approx p$, respectively.
$\Delta p$ is a positive real value that should be determined based on practical limitations (mainly the length of the observation sequence $T$).
We then computed the average surprise $\Bar{\textbf{S}}_{\mathrm{BF}}(p)$ and $\Bar{\textbf{S}}_{\mathrm{Sh}}(p)$ over $\mathcal{T}$ for each value of $p$.
We repeated this procedure for $N$ different simulated subjects (with different random seeds).
The average of $\Bar{\textbf{S}}_{\mathrm{BF}}$ and $\Bar{\textbf{S}}_{Sh}$ over $N=20$ subjects, for two learning algorithms (i.e. Nas12$^{*}$ and pf20), and for $T=500$ and $\Delta p = 0.0125$ is shown in \autoref{fig:experimentalpredictionsecond}B.

\section*{Acknowledgments}
We thank both reviewers for their constructive and helpful comments.
This research was supported by Swiss National Science Foundation No. 200020\_184615) and by the European Union Horizon 2020 Framework Program under grant agreement No.785907 (Human Brain Project, SGA2).

\section*{Appendix}

\subsection*{Modified algorithm of \cite{nassar2012rational, nassar2010approximately}: Adaptation for Gaussian prior}
\label{adaptednassar}
\subsubsection*{Recursive update of the estimated mean for Gaussian prior}
Let us first consider the case of a stationary regime (i.e. no change points) where observed samples are drawn from a Gaussian distribution with known variance, i.e. $y_{t+1}|\theta \sim \mathcal{N}(\theta, \sigma^2)$, and the parameter $\theta$ is also drawn from a Gaussian distribution $\theta \sim \mathcal{N}(\mu_0, \sigma_0^2)$.
After having observed samples $y_1, ..., y_{t+1}$, it can be shown that, using Bayes' rule, the posterior distribution $P(\theta | y_{1:t+1}) = \pi^{(t+1)}_B(\theta)$ is
\begin{Sequation}
  \label{Eq:Nassar_Gaussian_mean_update}
P(\theta | y_{1:t+1}) = \mathcal{N}\Big( \theta; \mu_{B, t+1} = \frac{1}{\frac{1}{\sigma^2_0} + \frac{t+1}{\sigma^2}} \Big(\frac{\mu_0}{\sigma^2_0} + \frac{\sum_{i=1}^{t+1}y_i}{\sigma^2}\Big), \sigma^2_{B, t+1} = \frac{1}{\frac{1}{\sigma^2_0} + \frac{t+1}{\sigma^2}} \Big) .
\end{Sequation}
An estimate of $\theta$ is its expected value $\mathbb{E}(\theta | y_{1:t+1}) = \mu_{B, t+1}$.

In a non-stationary regime where, after having observed $y_1, ..., y_t$ from the same hidden state, there is the possibility for a change point upon observing $y_{t+1}$, the posterior distribution is
\begin{Sequation}
  \label{Eq:Nassar_Bayesian_posterior_Rec_Formula_0}
P(\theta | y_{1:t+1}) = (1 - \gamma_{t+1}) P(\theta | y_{1:t+1}, c_{t+1} = 0) + \gamma_{t+1} P(\theta|y_{t+1}, c_{t+1} = 1)\, .
\end{Sequation}
To facilitate notation in this subsection we denote $c_{t+1} = 0$ as ``stay'' and $c_{t+1} = 1$ as ``change'' so that
\begin{Sequation}
  \label{Eq:Nassar_Bayesian_posterior_Rec_Formula_1}
P(\theta | y_{1:t+1}) = (1 - \gamma_{t+1}) P(\theta | y_{1:t+1}, \text{stay}) + \gamma_{t+1} P(\theta|y_{t+1}, \text{change})\,
\end{Sequation}
Note that the above is equivalent to Bayesian recursive formula (\autoref{Eq:Bayesian_Rec_Formula}) of the main text,
where $\gamma_{t+1}$ is the adaptation rate we saw in \autoref{Eq:Gamma_def} of the main text, and is essentially the probability to change given the new observation, i.e. $\textbf{P}(c_{t+1}=1 | y_{1:t+1})$.
In \cite{nassar2010approximately} this quantity is denoted as $\Omega_{t+1}$.
Taking \autoref{Eq:Nassar_Gaussian_mean_update} into account we have
\begin{Sequation}
  \begin{aligned}
  \label{Eq:SNassar_Exp_values}
  & \mathbb{E}(\theta | y_{1:t+1}, \text{stay}) = \mu_{B, t+1} = \frac{1}{\frac{1}{\sigma^2_0} + \frac{r_t+1}{\sigma^2}} \Big(\frac{\mu_0}{\sigma^2_0} + \frac{\sum_{i=t+1 - r_t}^{t+1}y_i}{\sigma^2}\Big)\, , \\
  & \mathbb{E}(\theta | y_{1:t+1}, \text{change}) = \frac{1}{\frac{1}{\sigma^2_0} + \frac{1}{\sigma^2}} \Big(\frac{\mu_0}{\sigma^2_0} + \frac{y_{t+1}}{\sigma^2}\Big)\, ,
\end{aligned}
\end{Sequation}
where $r_t$ is the time interval of observations coming from the same hidden state, calculated at time $t$.
Taking the expectation of \autoref{Eq:Nassar_Bayesian_posterior_Rec_Formula_1} the estimated mean upon observing the new sample $y_{t+1}$ is
\begin{Sequation}
  \label{Eq:SNassar_post_m_1}
  \approxs{\mu}_{t+1} = (1 - \gamma) \frac{1}{\frac{1}{\sigma^2_0} + \frac{r_t+1}{\sigma^2}} \Big(\frac{\mu_0}{\sigma^2_0} + \frac{\sum_{i=t+1 - r_t}^{t+1}y_i}{\sigma^2}\Big) + \gamma
  \frac{1}{\frac{1}{\sigma^2_0} + \frac{1}{\sigma^2}} \Big(\frac{\mu_0}{\sigma^2_0} + \frac{y_{t+1}}{\sigma^2}\Big)\, ,
\end{Sequation}
where we dropped the subscript $t+1$ in $\gamma$ to simplify notations.
We have
\begin{Sequation}
  \label{Eq:SNassar_post_m_2}
  \approxs{\mu}_{t+1} = (1 - \gamma) \frac{1}{\frac{1}{\sigma^2_0} + \frac{r_t+1}{\sigma^2}} \Big(\frac{\mu_0}{\sigma^2_0} + \frac{\sum_{i=t+1 - r_t}^{t}y_i}{\sigma^2} + \frac{y_{t+1}}{\sigma^2} \Big) + \gamma
  \frac{1}{\frac{1}{\sigma^2_0} + \frac{1}{\sigma^2}} \Big(\frac{\mu_0}{\sigma^2_0} + \frac{y_{t+1}}{\sigma^2}\Big) \, .
\end{Sequation}
Because $\approxs{\mu}_{t} = \frac{1}{\frac{1}{\sigma^2_0} + \frac{r_t}{\sigma^2}} \Big(\frac{\mu_0}{\sigma^2_0} + \frac{\sum_{i=t+1 - r_t}^{t}y_i}{\sigma^2}\Big)$, after a few lines of algebra we have
\begin{Sequation}
  \begin{aligned}
  \label{Eq:SNassar_post_m_3}
  \approxs{\mu}_{t+1} &= (1 - \gamma)\approxs{\mu}_t + \gamma \mu_0 + (1 - \gamma) \frac{1}{\frac{\sigma^2}{\sigma^2_0} + r_t + 1} (y_{t+1} - \approxs{\mu}_t) + \gamma \frac{1}{\frac{\sigma^2}{\sigma^2_0} + 1} (y_{t+1} - \mu_0).
\end{aligned}
\end{Sequation}
We now define $\rho = \frac{\sigma^2}{\sigma^2_0} $ and find
\begin{Sequation}
  \label{Eq:SNassar_post_m_4}
  \approxs{\mu}_{t+1} = (1 - \gamma)\approxs{\mu}_t + \gamma \mu_0 + (1 - \gamma) \frac{1}{\rho + r_t + 1} (y_{t+1} - \approxs{\mu}_t) + \gamma \frac{1}{\rho + 1} (y_{t+1} - \mu_0)\, .
\end{Sequation}

A rearrangement of the terms and inclusion of the dependency of $\gamma$ on time yields
\begin{Sequation}
  \label{Eq:SNassar_post_m_4b}
  \approxs{\mu}_{t+1} = (1 - \gamma_{t+1})\Big( \approxs{\mu}_t + \frac{1}{\rho + r_t + 1} (y_{t+1} - \approxs{\mu}_t) \Big) + \gamma_{t+1} \Big( \mu_0 + \frac{1}{\rho + 1} (y_{t+1} - \mu_0) \Big)\, .
\end{Sequation}
In order to obtain a form similar to the one of \cite{nassar2012rational, nassar2010approximately} we continue and we spell out the terms that include the quantities $\approxs{\mu}_t, \mu_0$ and $y_{t+1}$
\begin{Sequation}
  \begin{aligned}
  \label{Eq:SNassar_post_m_5}
  \approxs{\mu}_{t+1} &= (1 - \gamma)\approxs{\mu}_t - (1 - \gamma) \frac{1}{\rho + r_t + 1}\approxs{\mu}_t\\
  &+ \gamma \mu_0 - \gamma \frac{1}{\rho + 1} \mu_0\\
  &+ (1 - \gamma) \frac{1}{\rho + r_t + 1} y_{t+1} + \gamma \frac{1}{\rho + 1} y_{t+1}
\end{aligned}
\end{Sequation}
Using that $\frac{1}{\rho + r_t + 1} = \frac{1}{\rho + 1} - \frac{r_t}{(\rho + 1)(\rho + r_t + 1)}$ we have
\begin{Sequation}
  \begin{aligned}
  \label{Eq:SNassar_post_m_6}
  \approxs{\mu}_{t+1} &= (1 - \gamma)\approxs{\mu}_t - (1 - \gamma) \frac{1}{\rho + 1}\approxs{\mu}_t  + (1 - \gamma) \frac{r_t}{(\rho + 1)(\rho + r_t + 1)}\approxs{\mu}_t\\
  &+ \gamma \mu_0 - \gamma \frac{1}{\rho + 1} \mu_0\\
  &+ (1 - \gamma) \frac{1}{\rho + 1} y_{t+1} - (1 - \gamma) \frac{r_t}{(\rho + 1)(\rho + r_t + 1)} y_{t+1} + \gamma \frac{1}{\rho + 1} y_{t+1}.
\end{aligned}
\end{Sequation}

After a further step of algebra we arrive at
\begin{Sequation}
  \begin{aligned}
  \label{Eq:SNassar_post_m_7}
  \approxs{\mu}_{t+1} &= \frac{\rho}{\rho + 1} \Big( (1 - \gamma)\approxs{\mu}_t + \gamma \mu_0 \Big) + \frac{1}{\rho + 1} \Big( (1 - \gamma) \frac{r_t}{\rho + r_t + 1} (\approxs{\mu}_t - y_{t+1}) + y_{t+1} \Big)\, .
\end{aligned}
\end{Sequation}
If we define $1 - \alpha = (1 - \gamma) \frac{r_t}{\rho + r_t + 1} \Rightarrow \alpha = 1 - (1 - \gamma) \frac{r_t}{\rho + r_t + 1} \Rightarrow \alpha = \frac{\rho + \gamma r_t +1}{\rho + r_t + 1}$ and rearrange the terms, we have
\begin{Sequation}
  \begin{aligned}
  \label{Eq:SNassar_post_m_final_1}
  \approxs{\mu}_{t+1} &= \frac{\rho}{\rho + 1} \Big( (1 - \gamma)\approxs{\mu}_t + \gamma \mu_0 \Big) + \frac{1}{\rho + 1} \Big( (1 - \alpha)\approxs{\mu}_t + \alpha y_{t+1} \Big)\\
  \approxs{\mu}_{t+1} &= \frac{\rho}{\rho + 1} \Big(\approxs{\mu}_t + \gamma(\mu_0 - \approxs{\mu}_t) \Big) + \frac{1}{\rho + 1} \Big( \approxs{\mu}_t + \alpha (y_{t+1} - \approxs{\mu}_t) \Big)\, .
\end{aligned}
\end{Sequation}
Adding back the dependency of $\gamma$ and $\alpha$ on time we finally have
\begin{Sequation}
  \label{Eq:SNassar_post_m_final_2}
  \approxs{\mu}_{t+1} = \frac{\rho}{\rho + 1} \Big(\approxs{\mu}_t + \gamma_{t+1}(\mu_0 - \approxs{\mu}_t) \Big) + \frac{1}{\rho + 1} \Big( \approxs{\mu}_t + \alpha_{t+1} (y_{t+1} - \approxs{\mu}_t) \Big)\, .
\end{Sequation}

\subsubsection*{Recursive update of the the Estimated Variance for Gaussian Prior}
In \cite{nassar2012rational} the authors calculate first the variance $\approxs{\sigma}^2_{t+1} = \text{Var}(\theta | y_{1:t+1})$ and based on this compute then $\approxs{r}_{t+1}$.
We derive here these calculations for the case of Gaussian prior.
We remind once again that
\begin{Sequation}
P(\theta | y_{1:t+1}) = (1 - \gamma_{t+1}) P(\theta | y_{1:t+1}, \text{stay}) + \gamma_{t+1} P(\theta|y_{t+1}, \text{change})\,
\end{Sequation}
Then for the variance $\approxs{\sigma}^2_{t+1} = \text{Var}(\theta | y_{1:t+1})$ we have
\begin{Sequation}
  \begin{aligned}
    \label{SNassar_variance_1}
    \approxs{\sigma}^2_{t+1} &= (1- \gamma)\sigma^2_{stay} + \gamma \sigma^2_{change} + (1- \gamma)\gamma (\mu_{stay} - \mu_{change})^2\\
    &= (1- \gamma)\sigma^2_{B, t+1} + \gamma \sigma^2_{change} + (1- \gamma)\gamma (\mu_{B, t+1} - \mu_{change})^2
  \end{aligned}
\end{Sequation}
where $\sigma^2_{B, t+1} = \frac{1}{\frac{1}{\sigma^2_0} + \frac{r_t+1}{\sigma^2}}$ and $\sigma^2_{change} = \frac{1}{\frac{1}{\sigma^2_0} + \frac{1}{\sigma^2}}$.

We have defined earlier $\rho = \frac{\sigma^2}{\sigma^2_0} $ so that
\begin{Sequation}
  \begin{aligned}
    \label{SNassar_variance_A}
    A &= (1- \gamma)\sigma^2_{B, t+1} + \gamma \sigma^2_{change}
= (1- \gamma)\frac{\sigma^2}{\rho + r_t + 1} + \gamma \frac{\sigma^2}{\rho + 1}
  \end{aligned}
\end{Sequation}
Using, as before, that $\frac{1}{\rho + r_t + 1} = \frac{1}{\rho + 1} - \frac{r_t}{(\rho + 1)(\rho + r_t + 1)}$ we have
\begin{Sequation}
  \begin{aligned}
    A
&= \frac{\sigma^2}{\rho + 1} \Big( 1 - (1 - \gamma)\frac{1}{\rho + r_t + 1}  \Big).
  \end{aligned}
\end{Sequation}
We have defined earlier the learning rate $\alpha = 1 - (1 - \gamma) \frac{r_t}{\rho + r_t + 1}$, so we can write
\begin{Sequation}
  \begin{aligned}
    A &= \frac{\sigma^2}{\rho + 1} \alpha
  \end{aligned}
\end{Sequation}
Note that $\mu_{t} = \frac{1}{\frac{1}{\sigma^2_0} + \frac{r_t}{\sigma^2}} \Big(\frac{\mu_0}{\sigma^2_0} + \frac{\sum_{i=t+1 - r_t}^{t}y_i}{\sigma^2}\Big)$ so for the calculation of the last term we have
\begin{Sequation}
  \begin{aligned}
  \label{Eq:SNassar_variance_B1}
  B &= \mu_{B, t+1} - \mu_{change}\\
  &= \frac{1}{\frac{1}{\sigma^2_0} + \frac{r_t+1}{\sigma^2}} \Big(\frac{\mu_0}{\sigma^2_0} + \frac{\sum_{i=t+1 - r_t}^{t+1}y_i}{\sigma^2}\Big) -
  \frac{1}{\frac{1}{\sigma^2_0} + \frac{1}{\sigma^2}} \Big(\frac{\mu_0}{\sigma^2_0} + \frac{y_{t+1}}{\sigma^2}\Big).
\end{aligned}
\end{Sequation}
We now rearrange terms
\begin{Sequation}
  \begin{aligned}
  \label{Eq:SNassar_variance_B2}
  B
&= \mu_t + (\frac{1}{\rho + 1} - \frac{r_t}{(\rho + 1)(\rho + r_t + 1)}) (y_{t+1} - \mu_t) - \mu_0 - \frac{1}{\rho + 1} (y_{t+1} - \mu_0),\\
\end{aligned}
\end{Sequation}
and finally we have
\begin{Sequation}
  \begin{aligned}
    \label{Eq:SNassar_variance_final}
\approxs{\sigma}^2_{t+1} = & \frac{\sigma^2}{\rho + 1} \alpha + ( 1- \gamma)\gamma B^2.
  \end{aligned}
\end{Sequation}

\subsection*{Implementation of Nas10$^{*}$, Nas12$^{*}$ and Particle Filtering with 1 particle for the Gaussian estimation task}

\label{nassar_pseudocodes}
We provide here the pseudocode for the algorithms Nas10$^{*}$, Nas12$^{*}$ and Particle Filtering with 1 particle for the Gaussian estimation task.
Observations are drawn from a Gaussian distribution with known variance and unknown mean, i.e. $y_{t+1}|\mu_{t+1} \sim \mathcal{N}(\mu_{t+1}, \sigma^2)$ and $\theta_t = \mu_t$.
When there is a change, the parameter $\mu$ is also drawn from a Gaussian distribution $\mu \sim \mathcal{N}(\mu_0, \sigma_0^2)$.
All three algorithms estimate the expected $\mu_{t+1}$ (i.e. $\approxs{\mu}_{t+1}$) upon observing a new sample $y_{t+1}$.

After re-writing \autoref{Eq:S_GM_exp} we have
\begin{Sequation}
\label{Eq:S_GM_gaussian}
\textbf{S}_{\mathrm{BF}} \Big( y_{t+1}; \approxs{\mu}_{t}, \approxs{\sigma}_{t} \Big) =
  \frac{\sigma^2 + \approxs{\sigma}^2_{t}}{\sigma^2 + \sigma^2_{0}}
    \exp{\Big[- \frac{\mu^2_0}{2\sigma^2_0}
    - \frac{\approxs{\mu}^2_{t}}{2\approxs{\sigma}^2_{t}}
    + \frac{\frac{\approxs{\mu}_{t}}{\approxs{\sigma}^2_{t}} + \frac{y_{t+1}}{\sigma^2}}{ 2(\frac{\sigma^2}{\approxs{\sigma}^2_t} + 1)}
    + \frac{\frac{\mu_0}{\sigma^2_0} + \frac{y_{t+1}}{\sigma^2}}{2(\frac{\sigma^2}{\sigma^2_0} + 1)}
    \Big]}
\end{Sequation}

Note that the pseudocode for pf1 provided here, is a translation of Algorithm \ref{Alg:PF} to the case of a single particle
 (where there are no weights to calculate) and for the Gaussian distribution as a particular instance of the exponential family.

\begin{Salgorithm} \caption{Pseudocode for Nas10$^{*}$ for the Gaussian estimation task}
  \label{Alg:Nas10}
  \begin{algorithmic}[1]
\State Specify $m=p_c/(1-p_c)$, $\mu_{0}$, $\sigma_{0}$, $\sigma$ and $\rho = \sigma^2 / \sigma_{0}^2$.
  \State Initialize $\approxs{\mu}_{0}$, $\approxs{\sigma}_{0}$, $\approxs{r}_0$ and $t \gets 0$.
  \While {the sequence is not finished}
  \State Observe $y_{t+1}$
  \MyStatex{Surprise}
  \State Compute $\textbf{S}_{\mathrm{BF}}(y_{t+1}; \approxs{\mu}_{t}, \approxs{\sigma}_{t})$ using \autoref{Eq:S_GM_gaussian}
  \MyStatex{Modulation factor}
  \State Compute $\gamma_{t+1} = \gamma \big( \textbf{S}_{\mathrm{BF}}(y_{t+1}; \approxs{\mu}_{t}, \approxs{\sigma}_{t}) , m \big)$ as in \autoref{Eq:Gamma_def}
  \MyStatex{Expected mean}
  \State Compute $\approxs{\mu}_{t+1}$ using \autoref{Eq:Nassar_post_m_4b}
  \MyStatex{Expected time interval}
  \State Compute $\approxs{r}_{t+1} = (1 - \gamma_{t+1})(\approxs{r}_t + 1) +  \gamma_{t+1}$
  \MyStatex{Expected variance}
  \State Compute $\approxs{\sigma}_{t+1} = \big[ \frac{1}{\sigma_0^2} + \frac{\approxs{r}_{t+1}}{\sigma^2} \big]^{-1}$
  \MyStatex{Iterate}
  \State $t \gets t+1$
  \EndWhile
\end{algorithmic}
\end{Salgorithm}

\begin{Salgorithm} \caption{Pseudocode for Nas12$^{*}$ for the Gaussian estimation task}
  \label{Alg:Nas12}
  \begin{algorithmic}[1]
\State Specify $m=p_c/(1-p_c)$, $\mu_{0}$, $\sigma_{0}$, $\sigma$ and $\rho = \sigma^2 / \sigma_{0}^2$.
  \State Initialize $\approxs{\mu}_{0}$, $\approxs{\sigma}_{0}$, $\approxs{r}_0$ and $t \gets 0$.
  \While {the sequence is not finished}
  \State Observe $y_{t+1}$
  \MyStatex{Surprise}
  \State Compute $\textbf{S}_{\mathrm{BF}}(y_{t+1}; \approxs{\mu}_{t}, \approxs{\sigma}_{t})$ using \autoref{Eq:S_GM_gaussian}
  \MyStatex{Modulation factor}
  \State Compute $\gamma_{t+1} = \gamma \big( \textbf{S}_{\mathrm{BF}}(y_{t+1}; \approxs{\mu}_{t}, \approxs{\sigma}_{t}) , m \big)$ as in \autoref{Eq:Gamma_def}
  \MyStatex{Expected mean}
  \State Compute $\approxs{\mu}_{t+1}$ using \autoref{Eq:Nassar_post_m_4b}
  \MyStatex{Expected variance}
  \State Compute the expected variance $\approxs{\sigma}_{t+1}$ using \autoref{Eq:SNassar_variance_final}
  \MyStatex{Expected time interval}
  \State Compute the expected time interval $\approxs{r}_{t+1} = \frac{\sigma^2}{\approxs{\sigma}^2_{t+1}} - \frac{\sigma^2}{\sigma^2_{0}}$
  \MyStatex{Iterate}
  \State $t \gets t+1$
  \EndWhile
\end{algorithmic}
\end{Salgorithm}

\begin{Salgorithm} \caption{Pseudocode for Particle Filtering with 1 particle for the Gaussian estimation task}
  \label{Alg:PF_Gaussian}
  \begin{algorithmic}[1]
  \State Specify $m=p_c/(1-p_c)$, $\mu_{0}$, $\sigma_{0}$, $\sigma$ and $\rho = \sigma^2 / \sigma_{0}^2$.
  \State Initialize $\approxs{\mu}_{0}$, $\approxs{\sigma}_{0}$, $\approxs{r}_0$ and $t \gets 0$.
  \While {the sequence is not finished}
  \State Observe $y_{t+1}$
  \MyStatex{Surprise}
  \State Compute $\textbf{S}_{\mathrm{BF}}(y_{t+1}; \approxs{\mu}_{t}, \approxs{\sigma}_{t})$ using \autoref{Eq:S_GM_gaussian}
  \MyStatex{Modulation factor}
  \State Compute $\gamma_{t+1} = \gamma \big( \textbf{S}_{\mathrm{BF}}(y_{t+1}; \approxs{\mu}_{t}, \approxs{\sigma}_{t}) , m \big)$ as in \autoref{Eq:Gamma_def}
  \MyStatex{Hidden state of particle}
  \State Sample $ c_{t+1}^{(1)} \sim \text{Bernoulli}\big(\gamma_{t+1}\big)$
  \MyStatex{Expected mean}
  \If {$c_{t+1}^{(1)} = 0$}
  \State $\approxs{\mu}_{t+1} \gets \approxs{\mu}_{t} + \frac{1}{\rho + \approxs{r}_t + 1} (y_{t+1} - \approxs{\mu}_{t})$ and $\approxs{r}_{t+1} \gets \approxs{r}_{t} + 1$
  \Else
  \State $\approxs{\mu}_{t+1} \gets \mu_{0} + \frac{1}{\rho + 1} (y_{t+1} - \mu_{0})$ and $\approxs{r}_{t+1} \gets  1$
  \EndIf
  \MyStatex{Expected variance}
  \State Compute the expected variance $\approxs{\sigma}_{t+1} = \frac{\sigma^2}{\approxs{r}_{t+1} + \rho}$
  \MyStatex{Iterate}
  \State $t \gets t+1$
  \EndWhile
\end{algorithmic}
\end{Salgorithm}

\begin{Sfigure}
    \centering
\makebox[\textwidth][r]{\includegraphics{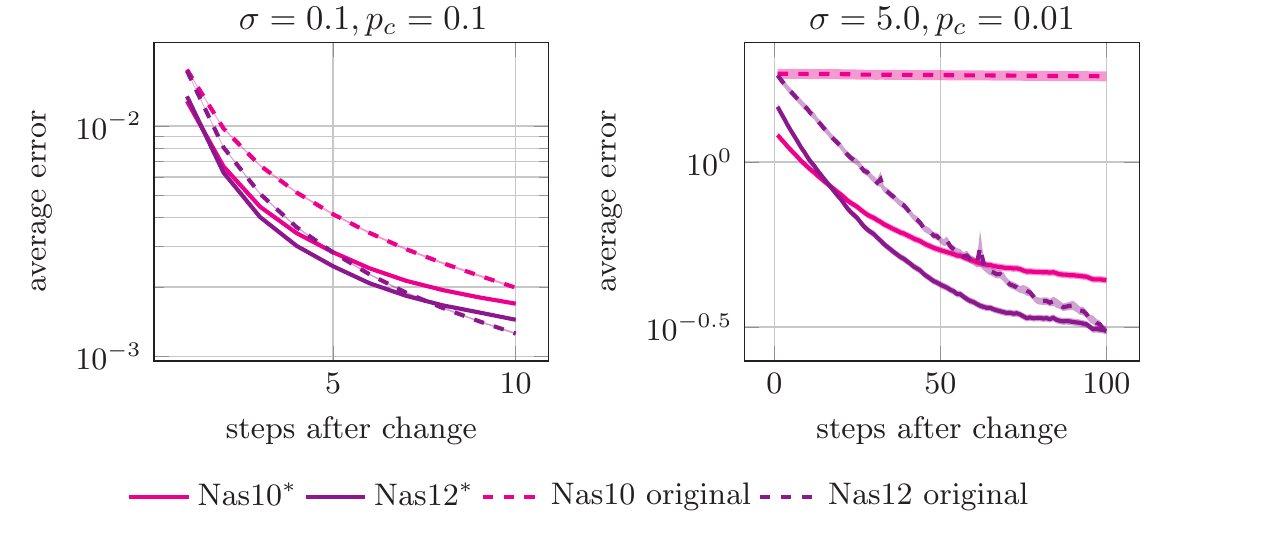}}
    \caption{\textbf{Gaussian estimation task: Transient performance after changes for original algorithms of \cite{nassar2010approximately} and \cite{nassar2012rational}.}
    Mean squared error for the estimation of $\mu_t$ at each time step $n$ after an environmental change, i.e.
    the average of $\textbf{MSE} [ \approxs{\Theta}_t | R_t = n]$ over time; $\sigma=0.1,\:p_c=0.1$ (left panel) and $\sigma=5,\:p_c=0.01$ (right panel).
    The shaded area corresponds to the standard error of the mean.
\textit{Abbreviations:}
  Nas10$^{*}$, Nas12$^{*}$: Variants of \cite{nassar2010approximately} and \cite{nassar2012rational} respectively,
  Nas10 Original, Nas12 Original: Original algorithms of \cite{nassar2010approximately} and \cite{nassar2012rational} respectively.
    }\label{fig:gauss_afterswitch_nassar}
\end{Sfigure}
\begin{Sfigure}
    \centering
\includegraphics[width=\textwidth]{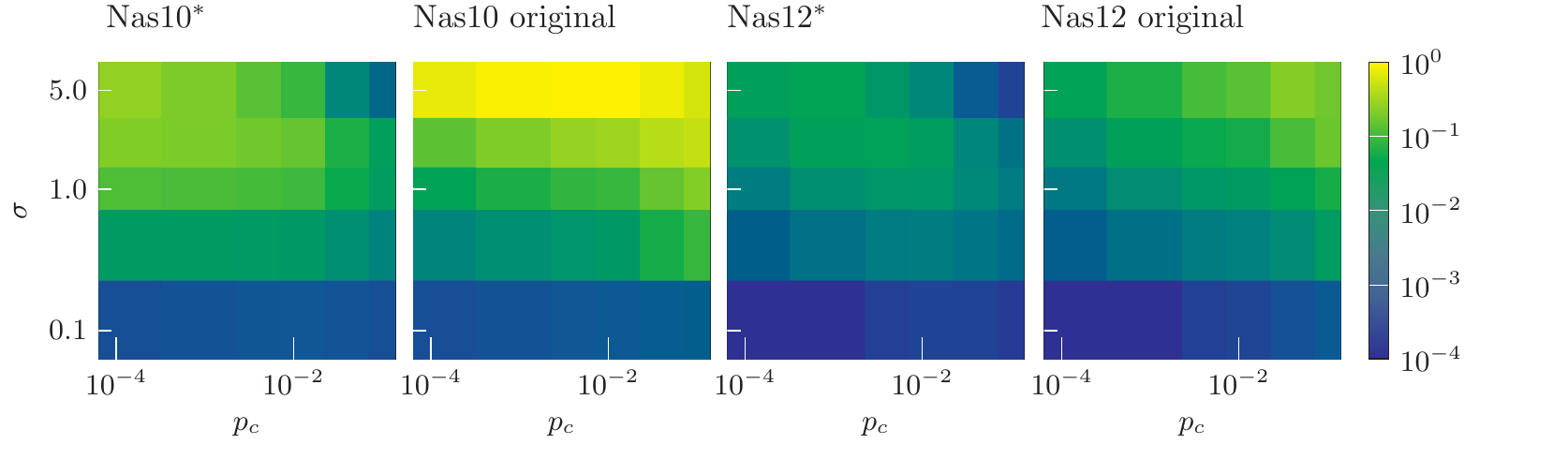}
    \caption{\textbf{Gaussian estimation task: Steady-state performance for original algorithms of \cite{nassar2010approximately} and \cite{nassar2012rational}.}
    Difference between the mean squared error of each algorithm and the optimal solution (Exact Bayes), i.e. the average $\Delta \textbf{MSE} [ \approxs{\Theta}_t ]$ over time for each combination of environmental parameters $\sigma$ and $p_c$.
\textit{Abbreviations:}
    Nas10$^{*}$, Nas12$^{*}$: Variants of \cite{nassar2010approximately} and \cite{nassar2012rational} respectively,
    Nas10 Original, Nas12 Original: Original algorithms of \cite{nassar2010approximately} and \cite{nassar2012rational} respectively.
    }\label{fig:gauss_heatmaps_nassar}
\end{Sfigure}

\newpage

\small
\bibliographystyle{apacite}
\def\url#1{}\def\doi#1{}\def\issn#1{}\def\isbn#1{} \newcommand{\noop}[1]{}

\end{document}